\documentclass[dvipsnames]{article}

\usepackage{styles/arxiv_ml_institute}

\usepackage[utf8]{inputenc} % allow utf-8 input
\usepackage[T1]{fontenc}    % use 8-bit T1 fonts
\usepackage{hyperref}       % hyperlinks
\usepackage{url}            % simple URL typesetting
\usepackage{booktabs}       % professional-quality tables
\usepackage{amsfonts}       % blackboard math symbols
\usepackage{nicefrac}       % compact symbols for 1/2, etc.
\usepackage{microtype}      % microtypography
\usepackage{xcolor}       % colors
\usepackage{svg}
\usepackage[abs]{overpic}
\usepackage{longtable}

\usepackage{annote-equations}

\usepackage[utf8]{inputenc} % allow utf-8 input
\usepackage[T1]{fontenc}    % use 8-bit T1 fonts
\usepackage{hyperref}       % hyperlinks
\usepackage{url}            % simple URL typesetting
\usepackage{booktabs}       % professional-quality tables
\usepackage{amsfonts}       % blackboard math symbols
\usepackage{nicefrac}       % compact symbols for 1/2, etc. .
\usepackage{microtype}      
\usepackage{nameref}
\usepackage[shortlabels]{enumitem}
\usepackage{textgreek}

\usepackage{float}
\usepackage{graphicx}

\usepackage{color}
\usepackage{amsmath}
\usepackage{amssymb}
\usepackage{amsthm}
\usepackage{mathtools}
\usepackage{thmtools}
\usepackage[mathscr]{eucal}
\usepackage{bm}
\usepackage{bbm}
\usepackage{relsize}
\usepackage{graphics}
\usepackage{wrapfig}
\usepackage{multirow}
\usepackage{enumitem} 

\usepackage{array}
\usepackage{adjustbox}

\usepackage{tocloft}

\usepackage{pgf}
\usepackage{xinttools}
\usepackage{tikz}
\usepackage{bbding}
\usepackage{pifont}
\usetikzlibrary{arrows.meta}
\usepackage{textcomp}
\usetikzlibrary{calc,shapes,arrows,positioning,automata,trees}
\usepackage{placeins} 
\usepackage{tabularx}
\usepackage{graphicx}
\usepackage{subcaption} 
\usepackage{listings}
\usepackage{xargs,lipsum,caption,changepage,ifthen}

\usepackage{tikz} 
\usepackage{makecell}
\usepackage{siunitx}

\usepackage{footmisc}

\usepackage{sidecap}

\usepackage{commands/ml-defs}

\definecolor{linkcol}{RGB}{134,22,87} % dark green: {20,88,21} % light green: {84,140,47}
\definecolor{citecol}{RGB}{22,91,137} % light blue: {128,168,179}
\definecolor{urlcol}{RGB}{20,88,21}
\hypersetup{
   colorlinks=true,
   linkcolor=linkcol,
   citecolor=citecol,
   urlcolor=urlcol,
}

\newcommand{\stoptocwriting}{%
  \addtocontents{toc}{\protect\setcounter{tocdepth}{-5}}}
\newcommand{\resumetocwriting}{%
  \addtocontents{toc}{\protect\setcounter{tocdepth}{\arabic{tocdepth}}}}

\newcommand{\cellstate}[1]{\eqnmarkbox[OliveGreen]{}{#1}}

\definecolor{xLSTMRed}{RGB}{229, 46, 102}
\definecolor{xLSTMBlue}{RGB}{22, 91, 137}
\newcommand{\expGate}[1]{\eqnmarkbox[xLSTMRed]{}{\!#1}}
\newcommand{\gates}[1]{\eqnmarkbox[xLSTMBlue]{}{#1}}

\newcommand{\first}[1]{\textbf{#1}}
\newcommand{\scd}[1]{\underline{#1}}

\title{xLSTM: Extended Long Short-Term Memory}

\author{%
  \ \ \ \   Maximilian Beck$^{* \ 1,2,3}$
  \ \ \ \ \ \ \ \
  Korbinian P\"{o}ppel$^{* \ 1,2,3}$
  \ \ \ \ \ \ \ \ \ \
  Markus Spanring$^{\ 1}$ \ \ \ \\
  \ \ \ \ \ \ \ \textbf{Andreas Auer}$^{\ 1,2}$
 \ \ \ \ \ \ \
  \textbf{Oleksandra Prudnikova}$^{\ 1}$ 
\ \ \ \ \ \ \ \ \ \ \ 
  \textbf{Michael Kopp} \ \ \ \ \ \ \ \ \\
   \ \ \ \  \textbf{G\"{u}nter Klambauer}$^{\ 1,2,3}$ 
  \ \ \ \ 
  \textbf{Johannes Brandstetter}$^{\ 1,2,3}$
  \ \ \ \ \ 
  \textbf{Sepp Hochreiter}$^{\ 1,2,3}$ \\ 
{$^*$}{Equal contribution}\\ %% I'd rather have it here than in footnote
{$^1$}{ELLIS Unit, LIT AI Lab, Institute for Machine Learning, JKU Linz, Austria}\\
{$^2$}{NXAI Lab, Linz, Austria}, \ \ \ 
{$^3$}{NXAI GmbH, Linz, Austria}
}

\begin{document}

\stoptocwriting

\maketitle
\vspace{-0.3cm}
\begin{abstract}
\vspace{-0.3cm}
In the 1990s, the constant error carousel and gating were introduced
as the central ideas of the Long Short-Term Memory (LSTM). 
Since then, LSTMs have stood the test of time 
and contributed to numerous deep learning success stories,
in particular they constituted the first Large Language Models (LLMs).
However, the advent of the Transformer technology with 
parallelizable self-attention at its core 
marked the dawn of a new era, outpacing LSTMs at scale.  
We now raise a simple question: 
How far do we get in language modeling 
when scaling LSTMs to billions of parameters, 
leveraging the latest techniques from modern LLMs,
but mitigating known limitations of LSTMs? 
Firstly, we introduce exponential gating 
with appropriate normalization and stabilization techniques. 
Secondly, we modify the LSTM memory structure, obtaining: 
(i) sLSTM with a scalar memory, a scalar update, and new memory mixing,
(ii) mLSTM that is fully parallelizable 
with a matrix memory and a covariance update rule.
Integrating these LSTM extensions into residual block backbones 
yields xLSTM blocks that are then residually stacked into xLSTM architectures.
Exponential gating and modified memory structures boost xLSTM capabilities to perform favorably 
when compared to state-of-the-art Transformers and 
State Space Models, both in performance and scaling.\\
Code available at: \url{https://github.com/NX-AI/xlstm}
\end{abstract}

\begin{figure}[H]
\vspace{-0.8em}
\centering
\includegraphics[width=1.\textwidth]{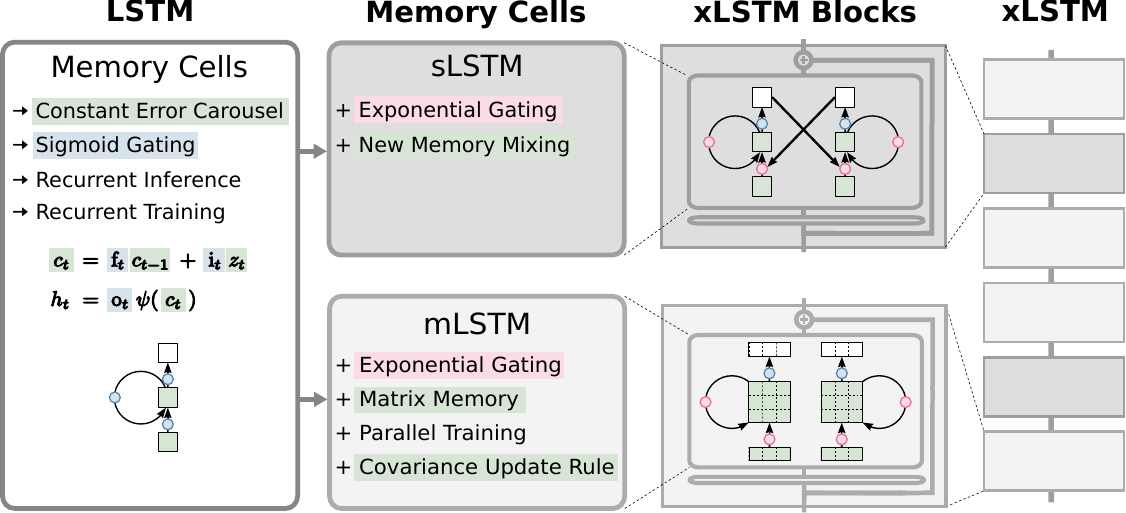}
\vspace{-0.8em}
\caption{The extended LSTM (xLSTM) family. From left to right: 
1. The original LSTM memory cell with constant error carousel and gating. 
2. New sLSTM and mLSTM memory cells that 
introduce exponential gating. 
sLSTM offers a new memory mixing technique. 
mLSTM is fully parallelizable with a novel matrix memory cell state and new
covariance update rule. 
3. mLSTM and sLSTM in 
residual blocks yield xLSTM blocks. 
4. Stacked xLSTM blocks give an 
xLSTM architecture. \label{fig:xlstm_sketch}}
\end{figure}

\section{Introduction}
\label{sec:intro}

The Long Short-Term Memory (LSTM) 
ideas~\citep{Hochreiter:91,Hochreiter:97nips,Hochreiter:97}, 
i.e., the constant error carousel and gating, 
were introduced to overcome 
the vanishing gradient problem of 
recurrent neural networks~\citep{Hochreiter:91,Hochreiter:00book}:
\begin{align}
\label{eq:lstm_idea}
  \cellstate{c_t} \ = \ \gates{\Rf_t} \ \cellstate{c_{t-1}} \ + \ 
          \gates{\Ri_t} \  \cellstate{z_t} \ , \quad  
          h_t \ = \ \gates{\Ro_t} \ \psi({\cellstate{c_t}}) \ . 
\end{align}
%Markus: It took me a bit to figure out that the color coding in eq.1 and the following eqs. comes from Figure 1. Maybe it can be stated explicitly also in the text e.g. sigmoid gating (blue) on the first occurrence?
%Sepp: the agreement with figure 1 is not so important: I would leave it.
The constant error carousel is the additive update of 
the cell state $c_{t-1}$ (green) by cell inputs $z_t$ 
and moderated by sigmoid gates (blue).
The input gate $\Ri_t$ and the forget gate $\Rf_t$ control this update,
while the output gate $\Ro_t$ controls the output of the memory cell, 
i.e.\ the hidden state $h_t$.
The cell state is normalized or squashed by $\psi$ 
and then output gating gives the hidden state.

LSTMs have been successfully applied to various domains
\citep{Hochreiter:01,Hochreiter:07,Schmidhuber:15},
and prevailed over text generation until
the dawn of Transformers in 2017~\citep{Vaswani:17}.
The effectiveness of LSTMs has been demonstrated at 
numerous sequence-related tasks such as 
generating text~\citep{Graves:13,Karpathy:15blog}, 
generating handwritings~\citep{Graves:13},
sequence-to-sequence translation~\citep{Sutskever:14nips}, 
evaluating computer programs~\citep{Zaremba:14arxiv}, 
generating image captions~\citep{Karpathy:15,Hossain:19},
generating source code~\citep{Karpathy:15blog}, 
rainfall-runoff modeling~\citep{Kratzert:18,Kratzert:19},
or hydrological models for flooding warnings~\citep{Nearing:24}.
In reinforcement learning, LSTMs are the best performing
sequence models, e.g., the AlphaStar model for StarCraft II~\citep{Vinyals:17short}, 
the OpenAI Five model for Dota 2~\citep{OpenAI:19}, and
models of the magnetic controller for nuclear fusion~\citep{Degrave:22short}.
LSTMs excel at learning abstractions, i.e., 
adeptly extracting semantic information and storing 
it in their memory cells~\citep{Karpathy:15blog}, which for example became evident
by number and syntax neurons~\citep{Lakretz:19}, 
linguistic neurons~\citep{Bau:19}, and
sentiment neurons~\citep{Radford:17arxiv}.
LSTMs are still used in 
highly relevant applications~\citep{Degrave:22short,Nearing:24}
and have stood the test of time.

\begin{wrapfigure}{r}{.6\textwidth}
 \vspace{-0.65cm}
    \includegraphics[angle=0,width=0.6\textwidth]{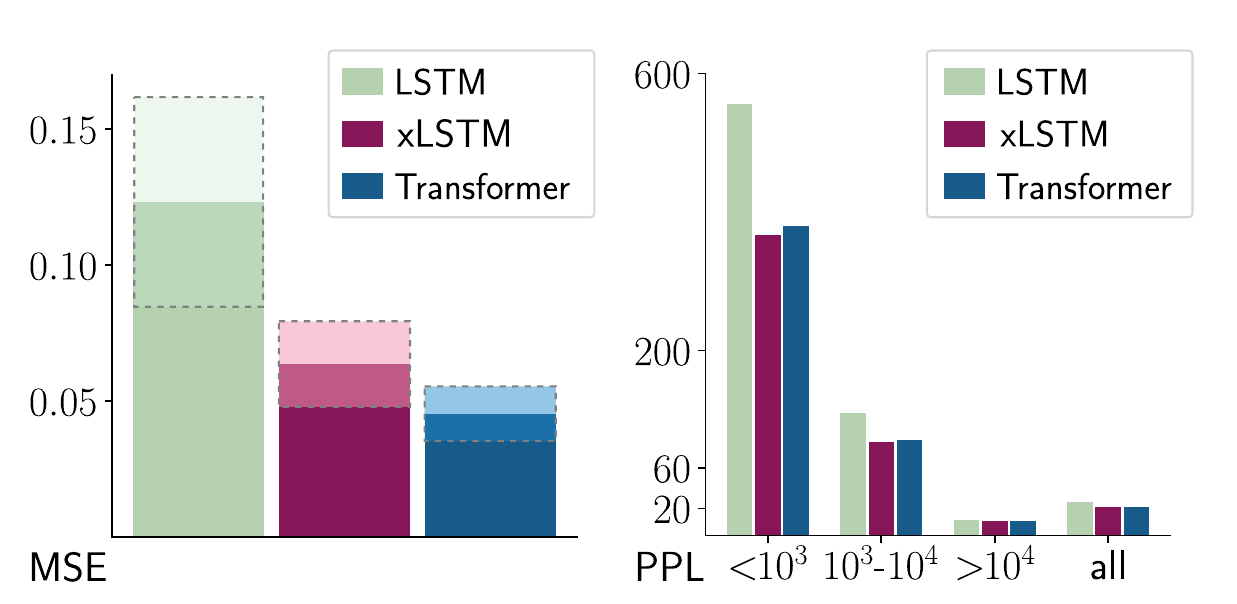}
    \caption{LSTM limitations. 
      {\bf Left}: Nearest Neighbor Search problem in terms of 
      mean squared error (MSE).
      Given a reference vector, 
      a sequence is scanned sequentially
      for the most similar vector with the objective to return its attached value at sequence end. 
      LSTM struggles to revise a 
      stored value when a more similar vector is found. 
      Our new xLSTM overcomes this limitation by exponential gating. 
      {\bf Right}: Rare Token Prediction. The perplexity (PPL)
      of token prediction on Wikitext-103, 
      in partitions of token frequency. % changed: buckets -> partitions
      LSTM performs worse on predicting rare tokens 
      because of its limited storage capacities, whereas
      our new xLSTM solves this  problem via a matrix memory.
      %SEPP: in Figure xLSTM[1:0] -> xLSTM and  xLSTM[0:1] -> xLSTM.
      %Same color for xLSTM
      \label{fig:lstmProblems}}            
 \vspace{-0.3cm}
\end{wrapfigure}

Despite their tremendous successes, 
LSTMs have three main limitations:
(i) Inability to revise storage decisions.
We exemplify this limitation via the 
{\it Nearest Neighbor Search} problem 
(see also Appendix~\ref{sec:appNearestNeighborSearch}): 
With a reference vector given, 
a sequence must be scanned sequentially
for the most similar vector in order to provide 
its attached value at sequence end.  
The left panel of Figure~\ref{fig:lstmProblems} shows 
the mean squared error at this task.
LSTM struggles to revise a stored value when a more similar 
vector is found,
while our new xLSTM remediates this limitation by exponential gating.
(ii) Limited storage capacities, i.e., 
information must be compressed into scalar cell states.
We exemplify this limitation via {\it Rare Token Prediction}. 
In the right panel of Figure~\ref{fig:lstmProblems},
the perplexity of token prediction on Wikitext-103~\citep{Merity:17}
is given for partitions of different token frequency.  % changed: buckets -> partitions
LSTM performs worse on rare tokens 
because of its limited storage capacities.
Our new xLSTM solves this problem by a matrix memory.
(iii) Lack of parallelizability due to memory mixing, i.e.,
the hidden-hidden connections between hidden states from one time step to the next,
which enforce sequential processing. 

These limitations of LSTM have paved the way for 
the emergence of Transformers~\citep{Vaswani:17} in
language modeling.
What performances can we achieve in language modeling 
when overcoming these limitations and scaling LSTMs 
to the size of current Large Language Models?

%##########################################################
%##########################################################
%##########################################################
%##########################################################
%##########################################################
%##########################################################
%##########################################################

\section{Extended Long Short-Term Memory}
\label{sec:xLSTM}

To overcome the LSTM limitations, Extended Long Short-Term Memory (xLSTM) introduces 
two main modifications to the LSTM idea of Equation~\eqref{eq:lstm_idea}.
Those modifications --- exponential gating and novel memory structures --- 
enrich the LSTM family by two members:
(i) the new sLSTM (see Section~\ref{sec:sLSTM}) with a scalar memory, a scalar update, and memory mixing, and 
(ii) the new mLSTM (see Section~\ref{sec:mLSTM}) with a matrix memory and a covariance (outer product) update rule,
which is fully parallelizable.
Both sLSTM and mLSTM enhance the LSTM through exponential gating.
To enable parallelization, the mLSTM abandons memory mixing, i.e., 
the hidden-hidden recurrent connections.
Both mLSTM and sLSTM can be extended to multiple memory cells, 
where sLSTM features memory mixing across cells.
Further, the sLSTM can have multiple heads without 
memory mixing across the heads, but only memory mixing across cells within each head.
This introduction of heads for sLSTM together with exponential gating
establishes a new way of memory mixing.
For mLSTM multiple heads and multiple cells are equivalent.

Integrating these new LSTM variants into residual block modules 
results in xLSTM blocks (see Section~\ref{sec:xLSTMblock}). Residually stacking those xLSTM blocks
in architectures provides xLSTM architectures (see Section~\ref{sec:xLSTMarchitecure}).
See Figure~\ref{fig:xlstm_sketch} for the xLSTM architecture with its components.

%##########################################################
%##########################################################
%##########################################################
\subsection{Review of the Long Short-Term Memory}
\label{sec:orgLSTM}

The original LSTM idea~\citep{Hochreiter:91,Hochreiter:97nips,Hochreiter:97}
introduced the scalar memory cell as a central processing and storage unit 
that avoids vanishing gradients~\citep{Hochreiter:91,Hochreiter:00book}
through the constant error carousel (the cell state update).
The memory cell contains three gates: input, output, and forget gate.
The forget gate has been introduced by \citet{Gers:00}. 
The update rules of the LSTM memory cell at time step $t$ are:
\begin{align}
\cellstate{c_t} \ &= \  \gates{\Rf_t} \ \cellstate{c_{t-1}} \ + \ \gates{\Ri_t} \ \cellstate{z_t}
  &  & &\text{cell state} \\
h_t  \ &= \ \gates{\Ro_t} \ \tilde{h}_t \ , 
  & \tilde{h}_t \ &= \ \psi \left( \cellstate{c_t}\right)
  &\text{hidden state} \\
\cellstate{z_t} \ &= \ \varphi \left( \tilde{z}_t \right) \ , 
  &\tilde{z}_t \ &=  \ \Bw^\top_{z} \ \Bx_t \ + \
  r_{z}  h_{t-1} \ + \  b_{z} \ \
  &\text{cell input} \\
\gates{\Ri_t} \ &= \ \sigma \left( \tilde{\Ri}_t  \right) \ , 
  &\tilde{\Ri}_t \ &= \ \Bw^\top_{\Ri} \ \Bx_t \ + \
  r_{\Ri}  \ h_{t-1} \ + \  b_{\Ri} \ \
  &\text{input gate} \\
\gates{\Rf_t} \ &= \ \sigma \left(  \tilde{\Rf}_t \right) \ , 
  &\tilde{\Rf}_t \ &= \ \Bw^\top_{\Rf} \ \Bx_t  \ + \
  r_{\Rf}  \ h_{t-1} \ + \  b_{\Rf} \ \
  &\text{forget gate} \\
\gates{\Ro_t} \ &= \ \sigma \left( \tilde{\Ro}_t \right) \ , 
  &\tilde{\Ro}_t  \ &= \ \Bw^\top_{\Ro} \ \Bx_t \ + \
  r_{\Ro}  \ h_{t-1} \ + \  b_{\Ro} \ \
  &\text{output gate} 
\end{align}
The weight vectors $\Bw_{z}$, $\Bw_{\Ri}$,
$\Bw_{\Rf}$, and $\Bw_{\Ro}$ correspond to the
input weight vectors between inputs $\Bx_t$ and 
cell input, input gate, forget gate, and
output gate, respectively.
The weights $r_{z}$, $r_{\Ri}$,
$r_{\Rf}$, and $r_{\Ro}$ correspond to 
the recurrent weights between hidden state $h_{t-1}$
and cell input, input gate, forget gate, and
output gate, respectively.
$b_{z}$, $b_{\Ri}$,
$b_{\Rf}$, and $b_{\Ro}$ are the corresponding bias terms.
$\varphi$ and $\psi$ are the cell input and hidden state 
activation functions (typically $\tanh$). 
$\psi$ is used to normalize or squash the cell state, which would be
unbounded otherwise.
All gate activation functions are sigmoid, i.e.,
$\sigma \left( x \right)= 1/(1+\exp(-x))$.
In later formulations, 
multiple scalar memory cells $\Bc_t \in \dR^{d}$ were combined in a vector, which  
allows the usage of recurrent weight matrices $\BR \in \dR^{d \times d}$ 
to mix the cell outputs of memory cells~\citep{Greff:15}, for more details see Appendix~\ref{sec:apporgLSTM}. 
Ablation studies showed that all components of 
the memory cell are crucial~\citep{Greff:15}. 

\vspace{-0.5cm}

%##########################################################
%##########################################################
%##########################################################
\subsection{sLSTM}
\label{sec:sLSTM}
%\subsection{Exponential gating}
%\label{sec:xGating}
\vspace{-0.3cm}
To empower LSTMs with the ability to revise storage decisions, 
we introduce exponential
gates (red) together with normalization and stabilization. 
In particular, input and forget gates can have exponential activation functions.
For normalization, we introduce a normalizer state that sums up
the product of the input gate times all future forget gates.

The scalar sLSTM forward pass is:
\begin{align}
\label{eq:slstmforward}
\cellstate{c_t} \ &= \  \gates{\Rf_t} \ \cellstate{c_{t-1}} \ + \ \gates{\Ri_t} \ \cellstate{z_t} & & 
 &\text{cell state} \\
\cellstate{n_t} \ &= \  \gates{\Rf_t} \ \cellstate{n_{t-1}} \ + \ \gates{\Ri_t}  & & 
 &\text{normalizer state} \\
h_t \ &= \ \gates{\Ro_t} \ \tilde{h}_t \ ,
  &\tilde{h}_t \ &= \ \cellstate{c_t} / \cellstate{n_t}
&\text{hidden state} \\
\cellstate{z_t} \ &= \ \varphi \left( \tilde{z}_t \right) \ , 
  &\tilde{z}_t \ &=  \ \Bw^\top_{z} \ \Bx_t \ + \
  r_{z}  h_{t-1} \ + \  b_{z}
  &\text{cell input} \\
\gates{\Ri_t} \ &= \ \expGate{\exp} \left( \tilde{\Ri}_t  \right) \ , 
  &\tilde{\Ri}_t \ &= \ \Bw^\top_{\Ri} \ \Bx_t \ + \
  r_{\Ri}  \ h_{t-1} \ + \  b_{\Ri}
  &\text{input gate} \\
\gates{\Rf_t} \ &= \ \sigma \left(  \tilde{\Rf}_t \right) \ \text{OR} \
\expGate{\exp} \left(  \tilde{\Rf}_t \right) \ ,
  &\tilde{\Rf}_t \ &= \ \Bw^\top_{\Rf} \ \Bx_t  \ + \
  r_{\Rf}  \ h_{t-1} \ + \  b_{\Rf}
  &\text{forget gate} \\
\gates{\Ro_t} \ &= \ \sigma \left( \tilde{\Ro}_t \right) \ , 
  &\tilde{\Ro}_t  \ &= \ \Bw^\top_{\Ro} \ \Bx_t \ + \
  r_{\Ro}  \ h_{t-1} \ + \  b_{\Ro}
  &\text{output gate} 
\end{align}

% SH: pick up on normalizing celll state again
% LSTM: norm with psi; here with n 

We transfer the original LSTM gating techniques, i.e., input- and/or hidden-dependent gating plus bias term, to the new architectures. Exponential activation functions can lead to large values that cause 
overflows. Therefore, we stabilize gates 
with an additional state \mbox{$\cellstate{m_t}$~\citep{Milakov:18arxiv}}:
\begin{align}
\label{eq:slstmstabil}
\cellstate{m_t} \ &= \ \max \left( \log ( \gates{\Rf_t} ) + \cellstate{m_{t-1}} , \log ( \gates{\Ri_t} ) \right) &\text{stabilizer state} \\
\gates{\Ri'_t} \ &= \ \expGate{\exp} \left( \log \left ( \gates{\Ri_t} \right) - \cellstate{m_t} \right) \ = \ \expGate{\exp} \left( \tilde{\Ri}_t - \cellstate{m_t} \right) \ \, 
  &\text{stabil. input gate} \\
\gates{\Rf'_t} \ &= \ \expGate{\exp} \left( \log \left( \gates{\Rf_t} \right) + \cellstate{m_{t-1}} - \cellstate{m_t} \right) \ \, 
  &\text{stabil. forget gate}
\end{align}

We show in Appendix~\ref{sec:appsLSTM}, that replacing 
$\Rf_t$ by $\Rf'_t$ and $\Ri_t$ by $\Ri'_t$ 
in the forward pass does neither
change the output of the whole network nor
the derivatives of the loss with respect to the
parameters.

\paragraph{New Memory Mixing.} 
sLSTM can have multiple memory cells like the original LSTM
(see Appendix~\ref{sec:appsLSTM}).
Multiple memory cells enable memory mixing via 
recurrent connections $\BR_{\Bz}$, $\BR_{\bfi}$, $\BR_{\bff}$, $\BR_{\bfo}$ 
from hidden state vector $\Bh$ to memory cell input $\Bz$ and the gates
$\bfi$, $\bff$, $\bfo$, respectively.
A new aspect in memory mixing is the effect of exponential gating.
The new sLSTM can have multiple heads with memory mixing
within each head but not across heads.
The introduction of heads for sLSTM together with exponential gating
establishes a new way of memory mixing.

%##########################################################
%##########################################################
%##########################################################
\subsection{mLSTM}
\label{sec:mLSTM}

To enhance storage capacities of LSTMs,  
we increase the LSTM memory cell from a scalar $c \in \dR$
to a matrix $\BC \in \dR^{d \times d}$. 
Hence, retrieval is performed via a matrix multiplication.
At time $t$, we want to store a pair of vectors, 
the key $\Bk_t \in \dR^d$ and the value $\Bv_t \in \dR^d$
(we use the Transformer terminology).  
Later at time $t + \tau$, the value $\Bv_t$ should be retrieved 
by a query vector $\Bq_{t+\tau} \in \dR^d$.
This is the setting of Bidirectional Associative Memories (BAMs)
\citep{Kohonen:72,Anderson:72,Nakano:72,Anderson:77}.
%\citep{Willshaw:69,Kohonen:72,Anderson:72,Nakano:72,Anderson:77,Kosko:88,Palm:13}.
The covariance update rule~\citep{Sejnowski:77,Dayan:91} 
for storing a key-value pair is
\begin{align}
\BC_t \ &= \ \BC_{t-1} \ + \ \Bv_{t} \ \Bk_t^\top \ .
\end{align}
We assume a layer-norm
before projecting inputs
to keys and values, therefore they have zero mean.
The covariance update rule is optimal~\citep{Dayan:91} for a 
maximal separability of retrieved binary vectors, 
which is equivalent to a maximal signal/noise ratio. 
Higher separability is possible when limiting retrieval 
to pairwise interactions and conceding quadratic complexity
like attention~\citep{Krotov:16,Krotov:17,Ramsauer:21}.
The covariance update rule is equivalent to Fast Weight Programmers
\citep{Schmidhuber:92ncfastweights,Schlag:21}, which
have later been equipped with 
a constant decay rate multiplied to $\BC_{t-1}$ and 
a constant learning rate multiplied to 
$\Bv_{t} \Bk_t ^\top$~\citep{Ba:16}.
In this spirit,
we integrate the covariance update rule into the LSTM framework, where
the forget gate corresponds to decay rate and the input gate to the learning rate,
while the output gate scales the retrieved vector.

For this matrix memory, the normalizer state is the weighted sum of 
key vectors, where each key vector is weighted by the input gate
and all future forget gates.
Again, the normalizer state keeps record of the strength of the gates.
Since the dot product between query and normalizer state 
can be close to
zero, we use the absolute value of this dot product 
and lower bound it by a threshold (typically 1.0) as
done previously~\citep{Sun:23arxiv}.
The mLSTM forward pass is: 
% Max: We actually use $\Bk_t/\sqrt{d}$. Should we add it? - We added it in (23)
\begin{align}
\label{eq:mlstm_recurrent_begin}
\cellstate{\BC_t} \ &= \  \gates{\Rf_t} \ \cellstate{\BC_{t-1}} \ + \ 
  \gates{\Ri_t} \ \cellstate{\Bv_t \ \Bk_t^\top} & &  &\text{cell state} \\
\cellstate{\Bn_t} \ &= \  \gates{\Rf_t} \ \cellstate{\Bn_{t-1}} \ + \ 
  \gates{\Ri_t} \ \cellstate{\Bk_t} & &  &\text{normalizer state} \\
\Bh_t  \ &= \ \gates{\bfo_t} \ \odot \ \tilde{\Bh}_t
\ , \qquad \qquad 
\tilde{\Bh}_t \ = \   \cellstate{\BC_t} \cellstate{\Bq_t} \ / \ 
  \max \left\{ \ABS{\cellstate{\Bn_t^\top} \cellstate{\Bq_t}}, 1 \right\} 
   &&&\text{hidden state} \\
\Bq_t \ &= \ \BW_q \ \Bx_t \ + \ \Bb_q  & &  &\text{query input} \\
\Bk_t \ &= \ \frac{1}{\sqrt{d}} \BW_k \ \Bx_t \ + \ \Bb_k  & &  &\text{key input} \\
\Bv_t \ &= \ \BW_v \ \Bx_t \ + \ \Bb_v  & &  &\text{value input} \\
\gates{\Ri_t} \ &= \ \expGate{\exp} \!  \! \left( \tilde{\Ri}_t  \right) \ , 
 \qquad \qquad \ \ \ \ \,
  \tilde{\Ri}_t \ = \ \Bw^\top_{\Ri} \ \Bx_t \ + \  b_{\Ri} \ \ \ 
  &&&\text{input gate} \\
\gates{\Rf_t} \ &= \ \sigma \!  \left(  \tilde{\Rf}_t \right) \ \text{OR} \ \expGate{\exp} \!  \! \left(  \tilde{\Rf}_t \right) , \ \ \: \!
\tilde{\Rf}_t \ = \ \Bw^\top_{\Rf} \ \Bx_t  \ + \
  b_{\Rf}
  &&& \text{forget gate} \\
\label{eq:mlstm_recurrent_end}
\gates{\bfo_t} \ &= \ \sigma \left( \tilde{\bfo}_t \right) \ , \qquad \qquad \quad \ \ \ \,
  \tilde{\bfo}_t  \ = \ \BW_{\bfo} \ \Bx_t \ + \
  \Bb_{\bfo}
  &&&\text{output gate} 
\end{align}

mLSTM can have multiple memory cells like the original LSTM.
For mLSTM, multiple heads and multiple cells are equivalent as
there is no memory mixing. In order to stabilize the exponential 
gates of mLSTM, we use the same stabilization 
techniques as for sLSTM (see Equation~\ref{eq:slstmstabil}). 
Since the mLSTM has no memory mixing, 
this recurrence can be reformulated in a parallel version. 
For more details we refer to Appendix~\ref{sec:appmLSTM}.

%##########################################################
%##########################################################
%##########################################################
\subsection{xLSTM Architecture}
\label{sec:xLSTMarch}

\begin{wrapfigure}{r}{.6\textwidth}
 \vspace{-0.3cm}
 \includegraphics[angle=0,width=0.29\textwidth]{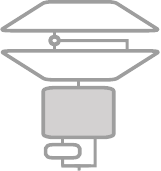}
  \hfill
  \includegraphics[angle=0,width=0.29\textwidth]{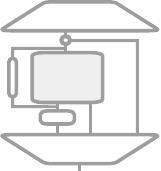}
       \caption{xLSTM blocks. {\bf Left}: A residual sLSTM block 
       with post up-projection 
       (like Transformers): 
       The input is fed into an sLSTM 
        --- with an optional convolution --- 
       followed by a gated MLP. 
        {\bf Right}: A residual mLSTM block with pre up-projection 
        (like State Space models): 
        mLSTM is wrapped inside two MLPs, via a convolution, 
        a learnable skip connection, and 
        an output gate that acts component-wise. 
        See Figure~\ref{fig:appsLSTM_detailed} and 
        Figure~\ref{fig:appmLSTM_detailed} in the appendix for details.
        \label{fig:Backbones} }
        %Markus: What information is this figure supposed to add? It feels a bit overloaded if it is only to show where the up and down projections are located. For details, e.g. it is not clear how I would see an optional convolution in the figure, the reader would anyway need to jump to the detailed one. At this point I would avoid tempting the reader to jump to the appendix and only give the general concept - he should be itching to go to the experiments and results by now.
        %SEPP: yes the up and down projections are the most important. They are the main differences
        % between the blocks. That with the convolution is indeed unclear. Also gated MLP becomes not clear.
  \vspace{-0.6cm}
\end{wrapfigure}

\paragraph{xLSTM Blocks.}
\label{sec:xLSTMblock}
An xLSTM block should non-linearly summarize the past in a high-dimensional
space to better separate different histories or contexts. 
Separating histories is the prerequisite to correctly predict the
next sequence element such as the next token.
We resort to Cover’s Theorem~\citep{Cover:65}, which states that
in a higher dimensional space non-linearly embedded patterns
can more likely be linearly separated
than in the original space.  
We consider two residual block architectures:
(i) A residual block with post up-projection
(like Transformers), which
non-linearly summarizes the past in the original space,
then linearly maps into a high-dimensional space,
applies a non-linear activation function, and 
linearly maps back to the original space;
see left panel of Figure~\ref{fig:Backbones}
and third column in Figure~\ref{fig:xlstm_sketch}. 
A more detailed version is depicted in Figure~\ref{fig:appsLSTM_detailed}
in the appendix.
(ii) A residual block with pre up-projection 
(like State Space Models), which 
linearly maps to a high-dimensional space, 
%and then 
non-linearly summarizes the past in the high-dimensional space
and
then linearly maps back to the original space.
For an xLSTM block containing an sLSTM,
we use the post up-projection block. 
For an xLSTM block containing an mLSTM,
we use the pre up-projection block since
the memory capacity becomes larger in the high-dimensional space.
We refer to the left panel of Figure~\ref{fig:Backbones}
and third column in Figure~\ref{fig:xlstm_sketch}, 
or Figure~\ref{fig:appsLSTM_detailed} in the appendix for more details.

\paragraph{xLSTM Architecture. }
\label{sec:xLSTMarchitecure}
An xLSTM architecture is constructed by residually stacking building blocks~\citep{Srivastava:15,He:16}. We rely on the most commonly used pre-LayerNorm~\citep{Ba:16b} residual backbones as used in contemporary Large Language Models.
See last column in Figure~\ref{fig:xlstm_sketch}. 
%SH Do we have something in the appendix concerning xLSTM architecture?

%##########################################################
%##########################################################
%##########################################################
\subsection{Memory and Speed Considerations}

Contrary to Transformers, xLSTM networks have a linear 
computation and a constant memory complexity
with respect to the sequence length.
Since the xLSTM memory is compressive, it is well suited
for industrial applications and implementations on the edge.

The memory of mLSTM does not require parameters but is
computationally expensive through its $d \times d$ matrix memory
and $d \times d$ update. We trade off memory 
capacity against computational complexity. 
Nevertheless, the computations can be done in parallel on GPUs, therefore
these computations have only a minor effect on the wall clock time.

While mLSTM is parallelizable 
analog to FlashAttention~\citep{Dao:22,Dao:23} or GLA~\citep{Yang:23arxiv},
sLSTM is not parallelizable due to 
the memory mixing (hidden-hidden connections).
However, we developed a fast CUDA implementation with 
GPU memory optimizations to the register level which is
typically less than two times slower than mLSTM.

%##########################################################
%##########################################################
%##########################################################
%##########################################################
%##########################################################
%##########################################################
\section{Related Work}
\label{sec:related}

\paragraph{Linear Attention.}
Several methods have been suggested to overcome the
quadratic complexity in terms of context length of the Transformer
and make attention linear in the context length.
The Synthesizer learns synthetic attention weights
without token--token interactions~\citep{Tay:20arxiv}.
Linformer realizes self-attention by a low-rank matrix and even 
linearly approximates it~\citep{Wang:20arxiv}.
Linear Transformer linearizes the attention mechanism
\citep{Katharopoulos:20}.
Performer linearly approximates the attention softmax by 
positive orthogonal random features approach~\citep{Choromanski:21}.
Attention has been replaced by fast long convolutions
in the Structured Global Convolution (SGConv)~\citep{Li:22}
and the Hyena Hierarchy~\citep{Poli:23}.

\paragraph{State Space Models.}
Recently, State Space Models (SSMs) became very popular since they
are linear in the context length and show promising performance compared
to Transformers.
One of the first proposed models was 
Structured State Space sequence model (S4)~\citep{Gu:21}, followed by
Diagonal State Space (DSS) model~\citep{Gupta:22},
Gated State Space (GSS) models~\citep{Mehta:22},
S5 model~\citep{Smith:22}, 
Bidirectional Gated SSM (BiGS)~\citep{Wang:22},
H3 model~\citep{Fu:23},
and Mamba~\citep{Gu:24arxiv}.

\paragraph{Recurrent Neural Networks.}
Recurrent Neural Networks (RNNs) have been suggested to
replace Transformer and attention due to their linearity
in the context length.
RNNs with Deep Linear Recurrent Units (LRUs) showed
promising results for language modeling~\citep{Orvieto:23, De:24arxiv},
as did Hierarchically Gated Linear RNN (HGRN)
\citep{Qin:23} and HGRN2~\citep{Qin:24arxiv}.
A well-known RNN approach to large language modeling
is RWKV~\citep{Peng:23arxivshort,Peng:24arxivshort}, 
showcasing competitive performance to Transformers.

\paragraph{Gating.}
One of the key ideas of LSTM is gating, which
was rediscovered and reinterpreted in many recent approaches.
Gating was used in HGRN~\citep{Qin:23},
HGRN2~\citep{Qin:24arxiv}, Gated Linear Attention (GLA)~\citep{Yang:23arxiv},
Gated State Space (GSS) models~\citep{Mehta:22},
Bidirectional Gated SSM (BiGS)~\citep{Wang:22},
Moving Average Equipped Gated Attention (MEGA)~\citep{Ma:22},
RWKV~\citep{Peng:23arxivshort},
and Mamba~\citep{Gu:24arxiv}.

\paragraph{Covariance Update Rule.}
To enhance storage capacities, we equipped the mLSTM cell 
with a matrix memory with a covariance update rule.
Other methods which build on such an update mechanism are
Fast Weight Programmers 
\citep{Schmidhuber:92ncfastweights,Schlag:21},
RWKV-5 and RWKV-6~\citep{Peng:24arxivshort},
Retention~\citep{Sun:23arxiv},
Linear Transformer~\citep{Katharopoulos:20},
and HGRN2~\citep{Qin:24arxiv}.

\paragraph{Most Related.}
Conceptually the closest models to xLSTM are Retention~\citep{Sun:23arxiv},
RWKV~\citep{Peng:23arxivshort,Peng:24arxivshort}, 
and HGRN2~\citep{Qin:24arxiv}. 
These models share the concepts matrix memory and/or gating.
However, in contrast to the new sLSTM, these approaches do not allow memory mixing.
Memory mixing enables to solve state tracking problems,
and therefore LSTMs are more expressive 
than State Space Models (SSMs) and Transformers~\citep{Merrill:24,Deletang:23}.
State tracking is required to evaluate code or 
to track entities in a long narrative.

\paragraph{Residually Stacking Architectures.}
Like almost all contemporary large deep learning models,
xLSTM architectures are constructed 
by residually stacking building blocks~\citep{Srivastava:15,He:16}.
%The ResNet architecture enables constant error flow 
%through the layers via the residual connections (weights of $1.0$, identity activation)
%similar to LSTM's constant error carousel (weights of $1.0$, identity %activation).
This construction enabled
deep convolutional networks~\citep{He:16} and
Transformers~\citep{Vaswani:17}. 
Transformers are the ultimate force behind  
Large Language Models (LLMs) like 
GPT-3~\citep{Brown:20short}, 
ChatGPT~\citep{Schulman:22short},
GPT-4~\citep{Achiam:23gpt4short},
Megatron-LM~\citep{Shoeybi:19},
Gopher~\citep{Rae:21short},
ERNIE 3.0 Titan~\citep{Wang:21arxivshort},
GLaM ~\citep{Du:21glamshort},
Chinese M6~\citep{Lin:21},
mutilingual AlexaTM 20B~\citep{Soltan:22},
OPT~\citep{Zhang:22opt},
Chinchilla~\citep{Hoffmann:22short},
BLOOM~\citep{Scao:22arxivshort},
GLM-130B~\citep{Zeng:22arxivshort},
LaMDA~\citep{Thoppilan:22short},
PaLM~\citep{Chowdhery:22short},
Llama~\citep{Touvron:23arxiv},
Gemini~\citep{GeminiTeam:23arxiv,Reid:24arxiv}.

%Nowadays, Transformer models are also widely used in computer vision, e.g.,
%VisionTransformer~\citep{Dosovitskiy:21},
%SimCLR~\citep{Chen:20},
%MoCo~\citep{He:20moco},
%Dino~\citep{Caron:21},
%Florence~\citep{Yuan:21},
%CLIP~\citep{Radford:21}, 
%Flamingo~\citep{Alayrac:22short}, 
%DALL-E 2~\citep{Ramesh:22},
%SORA~\citep{Brooks:24},
%or diffusion models~\citep{Rombach:22,Meng:22}. 
%See also methods in the
%overviews~\citep{Han:12arxiv,Khan:21arxiv}.

%##########################################################
%##########################################################
%##########################################################
%##########################################################
%##########################################################
%##########################################################
%##########################################################

\section{Experiments}
\label{sec:Exp}

In this section, we experimentally evaluate xLSTM 
and compare it to existing methods with a focus on language modeling.
We investigate xLSTM's specific capabilities on
synthetic tasks in Section~\ref{sec:ExpSyntethic}. 
In Section~\ref{sec:ExpComparison}, we compare the validation set perplexity
of various current language modeling methods that 
were trained on 15B tokens from SlimPajama~\citep{Soboleva:23}.
On the same dataset, we perform ablation studies for xLSTM.
Then, we assess the scaling behavior of the different methods
analogous to \citet{Kaplan:20} and \citet{Brown:20short}.
In Section~\ref{sec:ExpLanguage}, 
we conduct a more thorough language modeling experiment.
We compare xLSTM and the best performing methods 
from Section~\ref{sec:ExpComparison} after being trained on 
300B tokens from SlimPajama~\citep{Soboleva:23}.
First, we assess how well the methods perform in extrapolating to longer contexts,
secondly we test the methods via validation perplexity and performance 
on downstream tasks~\citep{Sutawika:24},
thirdly we evaluate the methods on 571 text domains of 
the PALOMA language benchmark dataset~\citep{Magnusson:23arxivshort},
fourthly we again assess the scaling behavior of the different methods, but
now with 20 times more training data.

\label{sec:xlstm_blocks_notation}
For all experiments, we use the 
notation xLSTM[$a$:$b$]
for the ratio $a/b$ of mLSTM-based versus sLSTM-based xLSTM blocks. 
For example, xLSTM[7:1] means that out of eight blocks, 
seven are mLSTM-based blocks and one is an sLSTM-based block. 
For a common total block number of 48, this translates to 
6 sLSTM-based blocks and 42 mLSTM-based blocks. 
Further, for all experiments, we use 
pre and post up-projection blocks for mLSTM and sLSTM, respectively. 
%Markus I think it is nowhere stated that the GPT-2 tokenizer is used

%##########################################################
%##########################################################
%##########################################################

\subsection{Synthetic Tasks and Long Range Arena}
\label{sec:ExpSyntethic}
First, we test the effectiveness of xLSTM's 
new exponential gating with memory mixing on 
formal languages~\citep{Deletang:23}.
Then, we assess the effectiveness of xLSTM's new matrix memory
on the Multi-Query Associative Recall task~\citep{Arora:23arxiv}.
Finally, xLSTM's performance at processing 
long sequences in the Long Range Arena is evaluated~\citep{Tay:21}.

\begin{figure}[h]
\includegraphics[width=\textwidth]{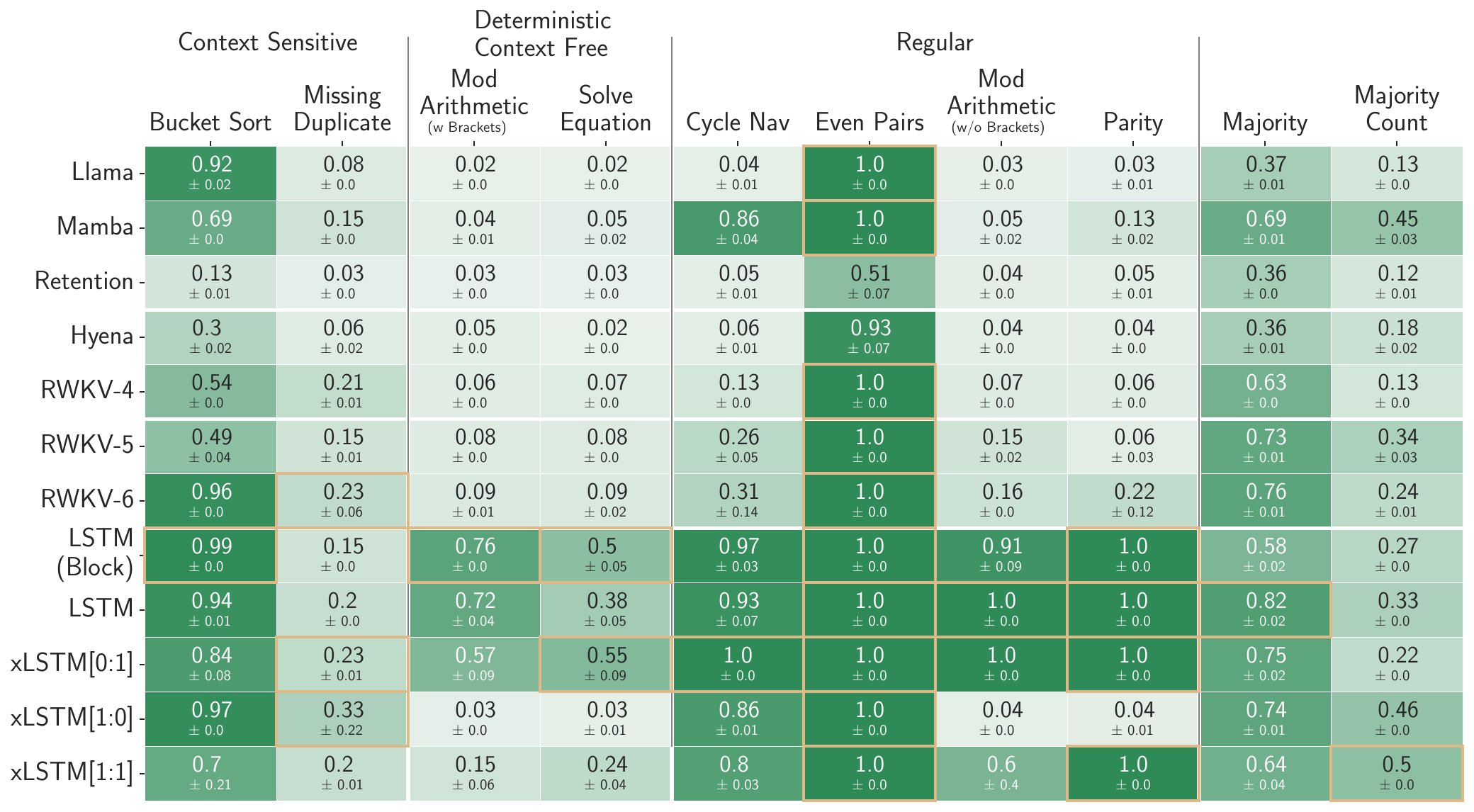}
\caption{Test of xLSTM's exponential gating with memory mixing. 
   Results are given by the scaled accuracy of different models at 
   solving formal language tasks, of which some require state tracking. 
   The different tasks are grouped by 
   the Chomsky hierarchy. \label{fig:formal-main}}
\end{figure}

\paragraph{Test of xLSTM's Exponential Gating with Memory Mixing.}
We test xLSTM's new exponential gating with memory mixing,
which should enable it to solve state tracking problems~\citep{Merrill:24, Merrill:23}.
We implement and extend the formal language tasks from \citet{Deletang:23}
to enable multi-length training for length extrapolation. 
For a detailed description of all tasks and extended results see 
Appendix~\ref{sec:appExpSynthetic-formalLang}. 
We compare xLSTM to other methods including Transformers, State Space Models,
and Recurrent Neural Networks.
The accuracy of the tested methods is evaluated
on those tokens relevant to the task.
The accuracy is scaled between 0 (random) and 1 (perfect).
We compare 2-block architectures of the following methods on these tasks: 
xLSTM[0:1] (i.e., only sLSTM), 
xLSTM[1:0] (i.e., only mLSTM), 
xLSTM[1:1],
Llama, 
Mamba,
RWKV, 
Retention, 
Hyena, 
LSTM, and
LSTM in Transformer blocks (LSTM (Block)).
The results of this experiment are shown in Figure~\ref{fig:formal-main}.
Models such as Transformers or State Space Models without memory mixing (no state tracking) 
cannot solve, e.g.\ regular grammars like the parity task. 
This result is in agreement with findings that 
Transformers and State Space models are fundamentally 
less powerful than RNNs~\citep{Merrill:24, Merrill:23, Deletang:23}. 
% JB: do we need this? are we opening a box with Mamba here? Maybe re-write this a bit.
% SH: yes you can rewrite. But this is an agrument why we want to scale up LSTMs
%     In principle they are powerful ...

\begin{figure}[h]
    \centering
    \includegraphics[width=\textwidth]{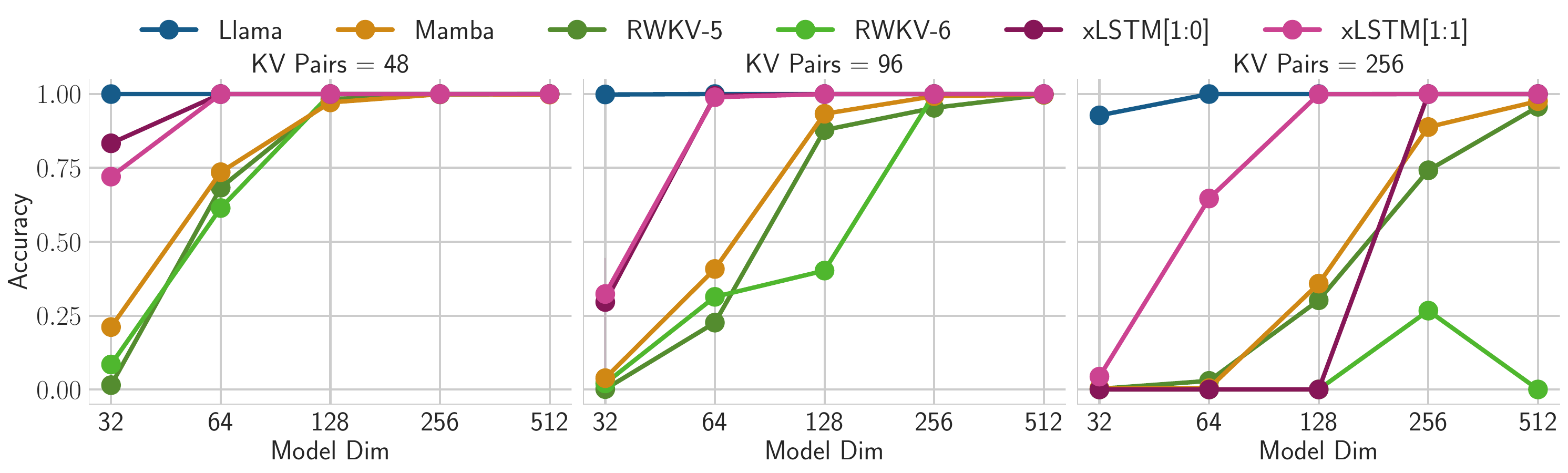}
    \caption{Test of memory capacities of different models at 
    the Multi-Query Associative Recall task with context length 2048. 
    Each panel is dedicated to a different number of key-value pairs. 
    The $x$-axis displays the model size and 
    the $y$-axis the validation accuracy.  \label{fig:mqar-main} }
\end{figure}

\paragraph{Test of xLSTM's Memory Capacities on Associative Recall Tasks.}
In this experiment, we test xLSTM's new matrix memory in terms of the
memory capacity on the Multi-Query Associative Recall task~\citep{Arora:23arxiv}:
For each sequence, key-value pairs are randomly chosen
from a large vocabulary,
which must be memorized for later retrieval.
To enhance the difficulty of the original task, 
we increase the number of key-value pairs up 
to 256 and extend the context length up to 2048.
Thus, we have broader tests for the memory capacities of different models.
We compare 2-block architectures of 
Llama, Mamba, RWKV-5, RWKV-6, xLSTM[1:1] and xLSTM[1:0]. 
The models are evaluated by the accuracy at recalling the pairs.
Since Transformers (e.g.\ Llama) have a memory that is 
exponential in the coding dimension~\citep{Ramsauer:21}, 
they constitute the gold standard at this task.
Results are shown in Figure~\ref{fig:mqar-main}. 
xLSTM[1:1] performs best among all non-Transformer models, 
also for small models.
Interestingly, the sLSTM block does not diminish the memory capacity but rather 
leverages it, which becomes evident at the most difficult 
task with 256 key-value pairs.
Additional results are presented in Appendix~\ref{sec:appExpSynthetic-mqar},
where extrapolation analyses
indicate that xLSTM's enhanced memory capacities also allow for extrapolating to contexts that are longer than those seen during training.

\paragraph{Test of xLSTM's Long Context Capabilities on Long Range Arena.}
To assess xLSTM's performance on long sequences and large contexts, 
we compare different methods on 
the Long Range Arena~\citep{Tay:21}. 
xLSTM demonstrates consistent strong performance on all of the tasks, 
suggesting that the xLSTM architecture is remarkably 
efficient in handling different aspects of long context problems.
% xLSTM performs well when compared to other methods.
% These experiments demonstrate that xLSTM's new exponential gating 
% and new memory structures do not diminish the performance
% of the original LSTM on long sequences.
For more details, see Appendix~\ref{sec:appExpSynthetic-lra}.

%##########################################################
%##########################################################
%##########################################################

\subsection{Method Comparison and Ablation Study}
\label{sec:ExpComparison}
% The main question of this paper is, 
% what can we achieve in language modeling when scaling up 
% the new LSTM variants.
% Therefore, we train xLSTMs, Transformers, State Space Models, and
% other methods on 15B tokens from SlimPajama 
% in an auto-regressive language modeling setting.
% \MB{For all models we use the GPT2 tokenizer.}
% \KP{Now in the appendix}
% Sepp: put it into comparison para.
% We compare the trained models on the validation set.
% Finally, we perform ablation studies for xLSTM.
%SH I reformulated "The main question of this paper is, 
%   if we can match Transformer's performance on Language Modeling."
% We do not directly compare to Transformer according to our story.
To address the main question of our paper, i.e. what can our new LSTM variants achieve when scaled
up in language modelling, we train xLSTMs, Transformers, State Space Models, and other methods
on 15B tokens from SlimPajama in the same auto-regressive setting. We compare the trained models
on the validation set and perform ablation studies for the xLSTMs.

\begin{wraptable}{r}{0.42\textwidth}
    \vspace{-0.4cm}
    \centering
    \begin{tabular}{lcr}
    \toprule
    Model               & \thead{\#Params                              \\ M} & \thead{SlimPajama \\ (15B) ppl $\downarrow$}\\
    \midrule
    GPT-3               & 356             & 14.26                      \\
    Llama               & 407             & \underline{14.25}          \\
    % \midrule
    % LSTM-block          & 506             & \underline{26.07}          \\
    \midrule
    H3                  & 420             & 18.23                      \\
    Mamba               & 423             & \underline{13.70}          \\
    \midrule
    Hyena               & 435             & 17.59                      \\
    RWKV-4              & 430             & 15.62                      \\
    RWKV-5              & 456             & \underline{14.25}          \\
    RWKV-6              & 442             & 15.03                      \\
    RetNet              & 431             & 16.23                      \\
    HGRN                & 411             & 17.59                      \\
    GLA                 & 412             & 16.15                      \\
    HGRN2               & 411             & 14.32                      \\
    \midrule
    % \textbf{xLSTM[0:1]} & 402             & 18.08                      \\
    \textbf{xLSTM[1:0]} & 409             & \underline{\textbf{13.43}} \\
    \textbf{xLSTM[7:1]} & 408             & 13.48                      \\
    \bottomrule
\end{tabular}

    \caption{Method comparison on next token prediction when trained on 15B tokens from SlimPajama. Best validation perplexities within model classes, i.e., Transformers, LSTMs,
   SSMs, RNNs, and linear Transformers are underlined
   and overall best is in bold.
   For each model class, the best performing methods are later used in Section~\ref{sec:ExpLanguage} for LLM training.
   \mbox{xLSTMs} with new memory
  (xLSTM[1:0] and \mbox{xLSTM[7:1]}) perform best. \label{tab:spaj15b_model_results}
}
% \vspace*{-\baselineskip}
% \vspace{0.4cm}
\end{wraptable}

\paragraph{Comparing xLSTM to Other Methods.}
For comparison, we train models on 15B tokens from SlimPajama~\citep{Soboleva:23}.
The trained models are evaluated 
by their perplexity on the validation set. 
We compare the following methods:
xLSTM (our new method),
GPT-3 (Transformer)~\citep{Brown:20short}, 
Llama (Transformer)~\citep{Touvron:23arxiv},
H3 (SSM)~\citep{Fu:23}, 
Mamba (SSM)~\citep{Gu:24arxiv},
RWKV-4 (RNN)~\citep{Peng:23arxivshort}, 
RWKV-5 (RNN)~\citep{Peng:24arxivshort}, 
RWKV-6 (RNN)~\citep{Peng:24arxivshort}, 
GLA (linear Transformer)~\citep{Yang:23arxiv}, 
HGRN (RNN)~\citep{Qin:23},
HGRN2 (RNN)~\citep{Qin:24arxiv}. 
RetNet (linear Transformer)~\citep{Sun:23arxiv}, 
Hyena (linear Transformer)~\citep{Poli:23},
xLSTM[1:0], and
xLSTM[7:1] (see Section~\ref{sec:xlstm_blocks_notation}). 
The models were trained with mixed precision, 
for RWKV-5, RWKV-6, GLA, HGRN2, the mixed-
precision training did not utilize the PyTorch automated mixed precision (see also Appendix Section~\ref{sec:appGenTrainProc}).
We categorize the methods into (a) Transformers, 
(b) State Space Models (SSMs), and
(c) Recurrent Neural Networks (RNNs) together with linear Transformers. 
Linear Transformers are linear methods that substitute the Transformer attention mechanism.
The models match a GPT-3 model with 350M parameters in size,
i.e.\ embedding dim 1024 and 24 residual blocks. 
Only GPT-3 uses shared weights for token and output embeddings,
therefore has fewer parameters.
The results in Table~\ref{tab:spaj15b_model_results} show that xLSTM
outperforms all existing methods in validation perplexity.
For details see Appendix~\ref{sec:appExpComparison}.
Figure~\ref{fig:temp_sclaw15B} shows the scaling behaviour
for this experiment, indicating that xLSTM will 
also perform favorably for larger models.

\begin{SCfigure}[30][htp]
   \centering
    \includegraphics[width=0.65\textwidth]{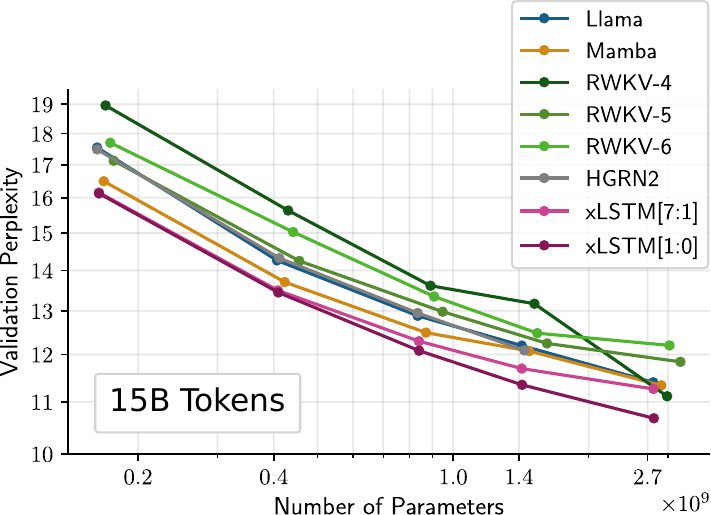}
    \vspace{-0.465cm}
    \caption{Method comparison on next token prediction when trained on 15B tokens from SlimPajama. 
    %\KP{Validation perplexities for the best methods of each model class} 
    Performance measure in validation perplexity for the best methods of each model class (see Table~\ref{tab:spaj15b_model_results}) are reported.
    The performance degradation of xLSTM[7:1] at 2.7B
    is due to initially slower training convergence that leads to 
    an especially undertrained model.
    xLSTM is the best method at all sizes.
    %The scaling indicate that xLSTM has high potential for larger models. 
    \label{fig:temp_sclaw15B} } 
\end{SCfigure}
%SEPP: x10^9 is strange placed ...
    %\vspace{0.8cm}

\paragraph{Ablation Studies.}
Table~\ref{tab:spaj15b_model_results} and Figure~\ref{fig:temp_sclaw15B} demonstrate that
xLSTM achieves excellent results at language modeling
when being trained on 15B tokens from SlimPajama.
To ablate the changes from
LSTM to xLSTM, we morph a vanilla LSTM architecture step-by-step into an xLSTM architecture.
Firstly, we integrate LSTM layers into pre-LayerNorm residual backbones. Secondly, we extend this to a post up-projection block. Finally, we add exponential gating and matrix memory. 
% Thus, it is only natural to ask which of %the changes 
% the elements of xLSTM is responsible for the 
% improvements over vanilla LSTM performances, 
% evoking an ablation study of the individual new xLSTM components.
% %(i) the exponential gating, (ii) the memory structures, and 
% %(iii) the particular gating technique.
% For doing so, we morph a vanilla LSTM architecture step-by-step 
% into an xLSTM architecture. First, we integrate LSTM layers into pre-LayerNorm 
% residual backbones, second we extend this to a post up-projection block, then we add exponential gating, and finally the matrix memory.
%(i) xLSTM with exponential gating (without matrix memory)
%to the original LSTM,
%(ii) xLSTM with matrix memory (mLSTM via xLSTM[1:0]) to 
%xLSTM without matrix memory (sLSTM via xLSTM[0:1]) and to the original LSTM, 
%(iii) xLSTM with matrix memory (xLSTM[1:0]) to xLSTM with matrix memory 
%(xLSTM[1:0]) but with different gating techniques.
The results are shown in Table~\ref{tab:ablstudies} (top). 
% The ablation studies attribute the strong performance 
% improvement to both the exponential gating and the matrix memory. 
% Additionally, since gating is an ever-occuring topic in RNNs and State 
% Space Models, we ablate different gating mechanisms.
The ablation studies attribute the strong performance improvement
to both the exponential gating and the matrix memory. Additionally, due to the importance of gating in
RNNs and State Space Models, we ablate different gating mechanisms.
In Table~\ref{tab:ablstudies} (bottom), we conclude that having each gate learnable and 
influenced by the input has an incremental positive effect. Additional 
studies on the individual backbone components are discussed in Appendix~\ref{sec:appAblDetails}.

% \begin{table}[htbp]
%     \centering
% \input{tables/spaj15b_ablations_small}
% \caption{xLSTM ablation studies for training on 15B tokens from SlimPajama. 
% We ablate exponential gating and new matrix memory to assess
% their contribution to xLSTM's performance.
% We observe a considerable improvement 
% from both the exponential gating and the matrix memory. 
% \KP{Should we use No... as ablation, the other way round or as it is now? Otherwise some things might be unclear}
% \MB{I would reformulate it. Line three also has matrix memory.. the difference between line 3 and 4 is post up-projection block vs. pre up-projection block.}
% \MB{Suggestion: I would split these table into two. (1) mLSTM only: Just for xLSTM[1:0] with rows 3+4 plus the convolution ablation. (2) (s)LSTMonly: rows: LSTM (multilayer) ppl very high, does not train, then LSTM in block structure, }
% \label{tab:spaj15b_abl_expgate} }
% \end{table}

\begin{table}[htbp]
    \centering
    \caption*{Ablation studies on the new xLSTM components.}
    \begin{adjustbox}{width=1.\textwidth}
        \begin{tabular}{llcccr}
    \toprule
    Model                       &  Modification                              & \thead{Exponential                                     \\ Gating}&\thead{Matrix \\ Memory}& \thead{\#Params         \\ M} & \thead{SlimPajama \\ (15B) ppl $\downarrow$}  \\
    \midrule
    \multirow{3}{*}{LSTM}       & Vanilla Multi-Layer LSTM            & \ding{55}          & \ding{55} & 607.8 & 2417.86       \\ % (1)
                                & Adding Resnet Backbone             & \ding{55}          & \ding{55} & 506.1 & 35.46         \\ % (2)
    % & Pre-Layernorm + Skip Connections      & \ding{55}          & \ding{55}  & 607.9 & 30.29         \\ % (3)
                                & Adding Up-Projection Backbone & \ding{55}          & \ding{55} & 505.9 & 26.01         \\ % (4)
    \midrule %\cmidrule{2-6}
    \multirow{1}{*}{xLSTM[0:1]} & Adding Exponential Gating & \ding{51}          & \ding{55} & 427.3 & 17.70         \\ % (5) slstmv1
    % & Pre Up-Projection Block               & \ding{51}          & \ding{55}  & 418.3 & 16.82         \\ % (6) slstmv2
    % \midrule
    % & Pre Up-Projection Block               & \ding{51}          & \ding{51}  & 410.4 & 13.64         \\ % (7) slstmv2
    \multirow{1}{*}{xLSTM[7:1]} & Adding Matrix Memory & \ding{51}          & \ding{51} & 408.4 & \first{13.48} \\ % (8) slstmv1 % this is the reevaluation number (others are during training) 
    \bottomrule
\end{tabular}
    \end{adjustbox}
    %\vspace{0.1cm}
    % \label{tab:spaj15b_abl_expgate_memory2v2}
    %\bigskip
    %\caption*{Ablation studies memory mixing.}
    %\begin{adjustbox}{width=1.\textwidth}
       % \input{./tables/abl_spaj15b_mlstm_blockconv}
    %\end{adjustbox}
    \vspace{0.1cm}
    %\bigskip
    \caption*{Ablation studies on different gating techniques.}
    \begin{adjustbox}{width=1.\textwidth}
        \begin{tabular}{lcccccccr}
    \toprule
    \multirow{3}{*}{Learnable Gates} & \multicolumn{3}{c}{Forget Gate} & & \multicolumn{3}{c}{Input Gate} & \multirow{3}{*}{\thead{SlimPajama \\ (15B) ppl $\downarrow$}}  \\ 
    & \thead{Input \\ Dependent} & \thead{Learnable \\ Bias} & \thead{Bias \\ Init} & & \thead{Input \\ Dependent} & \thead{Learnable \\ Bias} & \thead{Bias \\ Init} \\

    \midrule

    No Gates                                                    & \ding{55}    & \ding{55} & $+\infty$ & & \ding{55} & \ding{55} & $0$                  & NaN   \\
    No Gates                                                    & \ding{55}    & \ding{55} & $[3,6]$   & & \ding{55} & \ding{55} & $0$                  & 13.95 \\
    Forget Gate                                                 & \ding{51}    & \ding{51} & $[3,6]$   & & \ding{55} & \ding{55} & $0$                  & 13.58 \\
    Input Gate                                                  & \ding{55}    & \ding{55} & $[3,6]$   & & \ding{51} & \ding{51} & $\mathcal{N}(0,0.1)$ & 13.69 \\
    Forget Gate Bias                                            & \ding{55}    & \ding{51} & $[3,6]$   & & \ding{55} & \ding{55} & $0$                  & 13.76 \\
    Forget + Input Gate Bias                                    & \ding{55}    & \ding{51} & $[3,6]$   & & \ding{55} & \ding{51} & $\mathcal{N}(0,0.1)$ & 13.73 \\
    Forget Gate + Input Gate Bias                               & \ding{51}    & \ding{51} & $[3,6]$   & & \ding{55} & \ding{51} & $\mathcal{N}(0,0.1)$ & 13.55 \\
    \midrule
    Forget Gate + Input Gate                                    & \ding{51}    & \ding{51} & $[3,6]$   & & \ding{51} & \ding{51} & $\mathcal{N}(0,0.1)$ & \bf{13.43} \\
    \bottomrule
\end{tabular}

    \end{adjustbox}
    \vspace{0.1cm}

\caption{Ablation studies. 
\textbf{Top:} 
Ablation studies on the new xLSTM components, contributing the strong performance improvement of xLSTM over vanilla LSTM to both the exponential gating and the matrix memory.
%Architecture Modifications from LSTM to xLSTM. We show the design decisions extending the LSTM to become an xLSTM. 
%We start with a large default PyTorch LSTM. Due to a lack of skip-connections and layernorms an LSTM of this size is not trainable.
%Adding skip-connections and pre-layernorms fixes this. Additionally placing MLPs in between the LSTM layers further improves performance.
%By adding exponential gating we outperform the sigmoid gating LSTM by a large margin. 
%Finally, adding the best matrix memory from Table~\ref{tab:spaj15b_abl_matrixmemory} yields the xLSTM[7:1].
%Note that for xLSTM[0:1] the Pre Up-Projection Block performs better, while for xLSTM[7:1] the Post Up-Projection block yields better performance. 
%We hypothesize that this is due to the fact that the Pre Up-Projection Block yields a larger memory state vector and consequently a higher memory capacity. 
%For xLSTM[7:1] the memory is dominated by the mLSTMs matrix memory, hence a smaller memory state vector in the sLSTM is sufficient. \MB{Use reevaluation ppl numbers (will not change the results, just numerical).} 
\textbf{Bottom:} 
Ablation studies on different gating techniques. 
We consider an xLSTM[1:0] with sigmoid forget gate and 
exponential input gate. Bias initialization $\infty$ means that the forget gate is set to one, 
$[3,6]$ indicates that values are taken equidistant in the respective interval, 
and $\cN(0,0.1)$ that values are randomly chosen from a Gaussian with mean $0$ and std $0.1$. PPL denotes validation perplexity.
The first two lines correspond to models similar to linearized attention,
line four to Retention, 
line five to RWKV-5, and line six to RWKV-6. 
Dependencies of the gates on the input lead to better performance.
\label{tab:ablstudies}
}
\end{table}

\clearpage

%##########################################################
%##########################################################
%##########################################################
\clearpage

\subsection{xLSTM as Large Language Model}
\label{sec:ExpLanguage}

We culminate this study in large-scale language modeling experiments, testing the potential of xLSTM as an LLM. 
We therefore increase the amount of training data and train on 300B tokens from SlimPajama. The same number of tokens is used in, e.g., Mamba~\citep{Gu:24arxiv} and Griffin~\citep{De:24arxiv}.
% We compare xLSTM, RWKV-4, Llama, and Mamba, which were selected 
% as the best-performing methods in
% their respective method classes in the 
% model comparison in Section~\ref{sec:ExpComparison}.
We compare xLSTM
to RWKV-4, Llama, and Mamba – one method from each respective method class in Section~\ref{sec:ExpComparison}.
We select RWKV-4 as RNN representative since for RWKV-5, RWKV-6 and HGRN2 a reasonable
training precision setting has been found only after the training start of the
300B token experiments (see Appendix~\ref{sec:appGenTrainProc}).
We train different model sizes (125M, 350M, 760M, 1.3B), 
test all models for length extrapolation capabilities 
and evaluate their performance on the validation set.
We assess their performance on downstream tasks,
test their performance in language modeling 
on 571 text domains of the PALOMA benchmark, 
and, finally, investigate their scaling law behavior.

\begin{figure}[htbp]
    \vspace{-0.4cm}
    \centering
    \begin{minipage}{0.69\textwidth}
    \hspace{-1em}
    \includegraphics[width=1.15\textwidth]{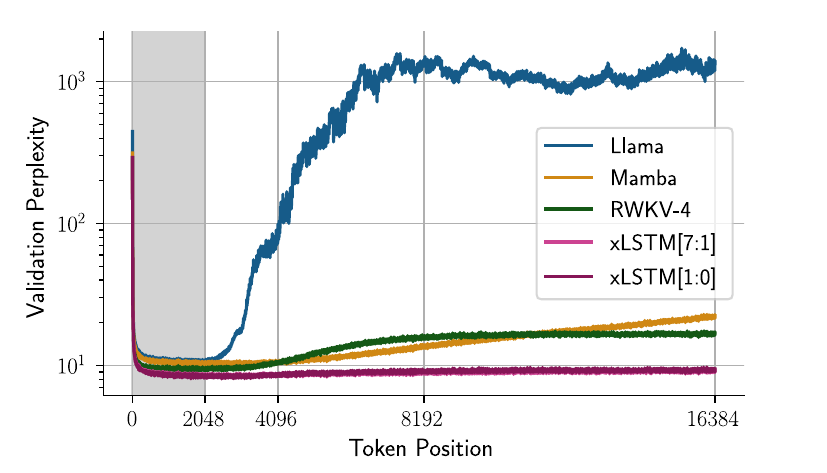}    
    \end{minipage}
    \hfill
    \begin{minipage}{0.27\textwidth}
         \centering
         \vspace{-0.7em}
         \begin{adjustbox}{width=\textwidth}
        \begin{tabular}{lr}
            \toprule
            \vspace{0.1em}
            Model & \thead{SlimPajama  \\ (300B) ppl $\downarrow$ \\ at 16k} \\
            \midrule
            Llama & 337.83 \\
            Mamba & 14.00 \\
            RWKV-4 & 13.75 \\
            xLSTM[7:1] & \first{8.92} \\
            xLSTM[1:0] & \scd{9.01} \\
            \bottomrule
        \end{tabular}
        \end{adjustbox}
    \end{minipage}
    \caption{Sequence extrapolation in language modeling. 
    This is a comparison of 1.3B-sized, large models of xLSTM, RWKV-4, Llama, and Mamba
    at next token prediction on the SlimPajama validation set after
    training on 300B tokens from SlimPajama. 
    Models are trained with context length 2048 
    and then tested for context lengths up to 16384.
    \textbf{Left:} 
    Token perplexities evaluated at different context lengths. 
    In contrast to other methods, 
    xLSTM models remain at low perplexities for longer contexts.
    \textbf{Right:} Prediction quality when
    extrapolating to long context sizes in terms of validation
    perplexity (PPL). xLSTM yields the best PPL values (best in bold, second best underlined).
    \label{fig:spaj300B_seq_extrapolation}
   }
\end{figure}

\vspace{-0.2cm}
\paragraph{Sequence Length Extrapolation.}
\label{sec:spaj300B_ctx_extrapolation}
Firstly, we test the sequence length extrapolation 
for 1.3B-sized, large models of xLSTM, RWKV-4, Llama, and Mamba.  
All models are trained on context length 2048, 
and then tested for context lengths up to 16384.
See Figure~\ref{fig:spaj300B_seq_extrapolation} for the results.
In contrast to other methods, xLSTM models maintain
low perplexities for longer contexts.

\vspace{-0.25cm}
\paragraph{Validation Perplexity and Downstream Tasks.}
\label{sec:spaj_downstream_eval}
Secondly, for all model sizes, we evaluate the 
performance of xLSTM, RWKV-4, Llama, and Mamba models
on the SlimPajama validation set for next token prediction and 
on downstream tasks that measure common sense reasoning.
%SH: downstream is not token prediction but most question answereing
The third column of 
Table~\ref{tab:lmeval} lists the validation set perplexities of different methods.
Both xLSTM[1:0] and xLSTM[7:1] are the best models for
all model sizes
with respect to the validation set perplexity.
The other columns of 
Table~\ref{tab:lmeval} provide the performance on downstream tasks.
In the vast majority of tasks and across all model sizes 
xLSTM is the best method --- only on the ARC task Mamba is
in some cases the best method.
For details see Appendix~\ref{sec:appExpLanguage}.

\begin{table}[htbp]
    \centering
    \begin{adjustbox}{width=\textwidth}
        \begin{tabular}{llcrrrrrrrrr}
    \toprule
                                                      & Model      & \thead{\#Params                                                                                                                                                 \\ M} & \thead{SlimPajama \\ (300B) ppl $\downarrow$} & \thead{LAMBADA \\ ppl $\downarrow$} & \thead{LAMBADA \\ acc $\uparrow$} & \thead{HellaSwag \\ acc $\uparrow$} & \thead{PIQA \\ acc $\uparrow$} & \thead{ARC-E \\ acc $\uparrow$} & \thead{ARC-C \\ acc $\uparrow$} & \thead{WinoGrande \\ acc $\uparrow$} & \thead{Average \\ acc $\uparrow$} \\
    \midrule
    \multirow{5}{*}{{\rotatebox[origin=c]{90}{125M}}} & RWKV-4     & 169.4           & 16.66         & 54.72         & 23.77         & 34.03         & 66.00         & 47.94         & 24.06         & 50.91         & 41.12         \\
                                                      & Llama      & 162.2           & 15.89         & 39.21         & 31.54         & 34.09         & 65.45         & 45.33         & 23.63         & 50.67         & 41.78         \\
                                                      & Mamba      & 167.8           & 15.08         & 27.76         & 34.14         & 36.47         & \scd{66.76}   & \first{48.86} & 24.40         & 51.14         & 43.63         \\
                                                      & xLSTM[1:0] & 163.8           & \scd{14.63}   & \first{25.98} & \first{36.52} & \scd{36.74}   & 65.61         & 47.81         & \scd{24.83}   & \first{51.85} & \scd{43.89}   \\
                                                      & xLSTM[7:1] & 163.7           & \first{14.60} & \scd{26.59}   & \scd{36.08}   & \first{36.75} & \first{66.87} & \scd{48.32}   & \first{25.26} & \scd{51.70}   & \first{44.16} \\
    \midrule
    \multirow{5}{*}{{\rotatebox[origin=c]{90}{350M}}} & RWKV-4     & 430.5           & 12.62         & 21.57         & 36.62         & 42.47         & 69.42         & 54.46         & 25.43         & 51.22         & 46.60         \\
                                                      & Llama      & 406.6           & 12.19         & 15.73         & 44.19         & 44.45         & 69.15         & 52.23         & 26.28         & 53.59         & 48.32         \\
                                                      & Mamba      & 423.1           & 11.64         & 12.83         & 46.24         & 47.55         & \scd{69.70}   & 55.47         & \scd{27.56}   & \scd{54.30}   & 50.14         \\
    %   & Mamba      & 371.5           & 11.66         & 12.98         & 45.88         & 47.51         & \scd{70.08}   & \scd{56.19}   & \first{28.75} & \scd{53.59}   & 50.33         \\
                                                      & xLSTM[1:0] & 409.3           & \first{11.31} & \first{11.49} & \first{49.33} & \first{48.06} & 69.59         & \scd{55.72}   & 26.62         & \first{54.38} & \scd{50.62}   \\
                                                      & xLSTM[7:1] & 408.4           & \scd{11.37}   & \scd{12.11}   & \scd{47.74}   & \scd{47.89}   & \first{71.16} & \first{56.61} & \scd{27.82}   & 53.28         & \first{50.75} \\
    \midrule
    \multirow{5}{*}{{\rotatebox[origin=c]{90}{760M}}} & RWKV-4     & 891.0           & 10.55         & 10.98         & 47.43         & 52.29         & \scd{72.69}   & 58.84         & 28.84         & 55.41         & 52.58         \\
                                                      & Llama      & 834.1           & 10.60         & 9.90          & 51.41         & 52.16         & 70.95         & 56.48         & 28.75         & 56.67         & 52.74         \\
                                                      & Mamba      & 870.5           & 10.24         & 9.24          & 50.84         & 53.97         & 71.16         & 60.44         & \scd{29.78}   & \scd{56.99}   & 53.86         \\
                                                      & xLSTM[1:0] & 840.4           & \first{9.86}  & \scd{8.09}    & \scd{54.78}   & \scd{55.72}   & \scd{72.69}   & \first{62.75} & \first{32.59} & \first{58.17} & \first{56.12} \\
                                                      & xLSTM[7:1] & 839.7           & \scd{9.91}    & \first{8.07}  & \first{55.27} & \first{56.12} & \first{72.74} & \scd{61.36}   & 29.61         & 56.43         & \scd{55.26}   \\
    \midrule
    \multirow{5}{*}{{\rotatebox[origin=c]{90}{1.3B}}} & RWKV-4     & 1515.2          & 9.83          & 9.84          & 49.78         & 56.20         & \scd{74.70}   & 61.83         & 30.63         & 55.56         & 54.78         \\
                                                      & Llama      & 1420.4          & 9.44          & 7.23          & \scd{57.44}   & 57.81         & 73.12         & 62.79         & 31.74         & 59.04         & 56.99         \\
                                                      & Mamba      & 1475.3          & 9.14          & 7.41          & 55.64         & \scd{60.45}   & 74.43         & \first{66.12} & \first{33.70} & \scd{60.14}   & \scd{58.41}   \\
                                                      & xLSTM[1:0] & 1422.6          & \first{8.89}  & \first{6.86}  & \first{57.83} & \first{60.91} & 74.59         & 64.31         & \scd{32.59}   & \first{60.62} & \first{58.48} \\
                                                      & xLSTM[7:1] & 1420.1          & \scd{9.00}    & \scd{7.04}    & 56.69         & 60.26         & \first{74.92} & \scd{65.11}   & 32.34         & 59.27         & 58.10         \\
    \bottomrule
\end{tabular}
% 423.1M & 11.64 & 12.83 & 46.24 & 47.55 & 69.7 & 55.47 & 27.56 & 54.3 & 50.14 \\
    \end{adjustbox}
    \vspace{0.1cm}
    \caption{Validation set perplexity and downstream tasks. 
    Comparison of xLSTM, RWKV-4, Llama, and Mamba
    on the validation set at next token prediction and on downstream tasks 
    after training on 300B tokens from SlimPajama. 
    Model sizes are 125M, 250M, 760M, 
    and 1.3B. The first column shows the methods and
    the second the actual number of parameters. 
    The third column lists the validation set perplexities,
    while the remaining columns
    show the performance on downstream tasks.
    Best model per model size is depicted bold and the second best 
    is underlined.
    In the vast majority of tasks and 
    across all model sizes xLSTM is the best method --- 
    only on the ARC task Mamba is 
    in some cases the best method.
    xLSTM[1:0] and xLSTM[7:1] are the two best models
    with respect to validation set perplexity.
    \label{tab:lmeval}
}  
\end{table}

\begin{table}[htbp]
    \centering
    \begin{adjustbox}{width=\textwidth}
        \begin{tabular}{llcrrrrrrrrrrrrr}
    \toprule
                                                      & Model      & \thead{\#Params                                                                                                                                                                                                               \\ M} & C4    & \thead{MC4 \\ EN} & \thead{Wikitext \\ 103} & \thead{Penn \\ Treebank} & \thead{Red \\ Pajama} & \thead{Refined \\ Web} & Dolma & \thead{M2D2 \\ S2ORC} & \thead{M2D2 \\ Wikipedia} & \thead{C4 \\ Domains} & \thead{Dolma \\ Subreddits} & \thead{Dolma \\ Coding} & \thead{Average} \\
    \midrule
    \multirow{5}{*}{{\rotatebox[origin=c]{90}{125M}}} & RWKV-4     & 169.4           & 26.25         & 22.33         & 29.18         & 38.45         & 8.99         & 32.47         & 17.04         & 23.86         & 21.42         & 22.68         & 37.08         & 5.12         & 23.74         \\
                                                      & Llama      & 162.2           & 24.64         & 17.23         & 23.16         & 31.56         & 8.26         & 29.15         & 15.10         & 19.71         & 20.41         & 21.45         & 36.73         & \scd{3.61}   & 20.92         \\
                                                      & Mamba      & 167.8           & 23.12         & 17.04         & 22.49         & 30.63         & 7.96         & 27.73         & 14.60         & 19.38         & 19.36         & 20.14         & 34.32         & 3.77         & 20.05         \\
                                                      & xLSTM[1:0] & 163.8           & \scd{22.54}   & \scd{16.32}   & \scd{21.98}   & \scd{30.47}   & \scd{7.80}   & \scd{27.21}   & \scd{14.35}   & \scd{19.02}   & \scd{19.04}   & \scd{19.65}   & \scd{34.15}   & 3.64         & \scd{19.68}   \\
                                                      & xLSTM[7:1] & 163.7           & \first{22.39} & \first{16.13} & \first{21.47} & \first{30.01} & \first{7.75} & \first{26.91} & \first{14.13} & \first{18.6}  & \first{18.84} & \first{19.52} & \first{33.9}  & \first{3.59} & \first{19.44} \\
    \midrule
    \multirow{5}{*}{{\rotatebox[origin=c]{90}{350M}}} & RWKV-4     & 430.5           & 19.55         & 15.82         & 19.64         & 27.58         & 6.97         & 24.28         & 12.94         & 17.59         & 15.96         & 16.98         & 29.40         & 3.90         & 17.55         \\
                                                      & Llama      & 406.6           & 18.38         & 13.28         & 16.41         & \first{21.82} & 6.56         & 22.09         & 11.76         & 15.05         & 15.25         & 15.99         & 28.30         & 3.12         & 15.67         \\
                                                      & Mamba      & 423.1           & 17.33         & 13.05         & 16.11         & 22.24         & 6.34         & 21.04         & 11.42         & 14.83         & 14.53         & \scd{15.16}   & 27.02         & 3.20         & \scd{15.19}   \\
    %   & Mamba      & 371.5           & 17.39         & 13.06         & 16.10         & 22.65         & 6.35         & 21.09         & 11.43         & 14.92         & 14.56         & 15.17         & 27.05         & 3.19         & \scd{15.25}   \\
                                                      & xLSTM[1:0] & 409.3           & \scd{17.01}   & \first{12.55} & \first{15.17} & 22.51         & \first{6.20} & \first{20.66} & \first{11.16} & \first{14.44} & \first{14.27} & \first{14.85} & \scd{26.70}   & \first{3.08} & \first{14.88} \\
                                                      & xLSTM[7:1] & 408.4           & \first{16.98} & \scd{12.68}   & \scd{15.43}   & \scd{21.86}   & \scd{6.23}   & \scd{20.70}   & \scd{11.22}   & \scd{14.62}   & \scd{14.30}   & \first{14.85} & \first{26.61} & \scd{3.11}   & \first{14.88} \\
    \midrule
    \multirow{5}{*}{{\rotatebox[origin=c]{90}{760M}}} & RWKV-4     & 891.0           & 15.51         & 12.76         & 14.84         & 21.39         & 5.91         & 19.28         & 10.70         & 14.27         & 13.04         & 13.68         & 24.22         & 3.32         & 14.08         \\
                                                      & Llama      & 834.1           & 15.75         & 11.59         & 13.47         & 18.33         & 5.82         & 19.04         & 10.33         & 13.00         & 13.05         & 13.76         & 24.80         & 2.90         & 13.49         \\
                                                      & Mamba      & 870.5           & 15.08         & 11.54         & 13.47         & 19.34         & 5.69         & 18.43         & 10.15         & 13.05         & 12.62         & 13.25         & 23.94         & 2.99         & 13.30         \\
                                                      & xLSTM[1:0] & 840.4           & \first{14.60} & \first{11.03} & \first{12.61} & \scd{17.74}   & \first{5.52} & \first{17.87} & \first{9.85}  & \first{12.50} & \first{12.20} & \first{12.81} & \scd{23.46}   & \first{2.87} & \first{12.76} \\
                                                      & xLSTM[7:1] & 839.7           & \scd{14.72}   & \scd{11.11}   & \scd{12.68}   & \first{17.61} & \scd{5.55}   & \scd{18.01}   & \scd{9.87}    & \scd{12.59}   & \scd{12.25}   & \scd{12.89}   & \first{23.43} & \scd{2.88}   & \scd{12.80}   \\
    \midrule
    \multirow{5}{*}{{\rotatebox[origin=c]{90}{1.3B}}} & RWKV-4     & 1515.2          & 14.51         & 12.04         & 13.73         & 19.37         & 5.62         & 18.25         & 10.11         & 13.46         & 12.10         & 12.87         & 22.85         & 3.25         & 13.18         \\
                                                      & Llama      & 1420.4          & 13.93         & 10.44         & 11.74         & \first{15.92} & 5.29         & 17.03         & 9.35          & \scd{11.61}   & 11.53         & 12.24         & 22.63         & \scd{2.74}   & 12.04         \\
                                                      & Mamba      & 1475.3          & 13.35         & 10.40         & 11.76         & 16.65         & 5.21         & 16.50         & 9.17          & 11.73         & 11.18         & 11.83         & \scd{21.43}   & 2.83         & 11.84         \\
                                                      & xLSTM[1:0] & 1422.6          & \first{13.13} & \first{10.09} & \scd{11.41}   & \first{15.92} & \first{5.10} & \first{16.25} & \first{9.01}  & \first{11.43} & \first{10.95} & \first{11.60} & \first{21.29} & \first{2.73} & \first{11.58} \\
                                                      & xLSTM[7:1] & 1420.1          & \scd{13.31}   & \scd{10.21}   & \first{11.32} & \scd{16.00}   & \scd{5.16}   & \scd{16.48}   & \scd{9.11}    & \scd{11.61}   & \scd{11.10}   & \scd{11.76}   & 21.50         & 2.75         & \scd{11.69}   \\
    \bottomrule
\end{tabular}
    \end{adjustbox}
    \vspace{0.1cm}
    \caption{Performance on PALOMA Language Modeling Tasks. 
    Comparison of xLSTM, RWKV-4, Llama, and Mamba by the perplexity
    of next token prediction on the PALOMA language benchmark 
    after training on 300B tokens from SlimPajama. 
    Model sizes are 125M, 250M, 760M, and 1.3B.
    The second column shows the actual number of parameters.
    The 571 text domains are grouped into language modeling (next seven columns) 
    and fine-grained domain benchmarks (further 5 columns). 
    The last column shows the average perplexity across all of these tasks.
    Best model per model size is given in bold and the second best 
    is underlined.
    xLSTM yields the best performance.
    %SEPP: Llama is best togehter with xLSTM[1:0] for 1.3B on Penn TreeBank. At M2D2
%S2ORC is Llama also second best. at Dolma
%Coding is Llama the secon best.
    \label{tab:ppleval}
}  
\end{table}

\vspace{-0.2cm}
\paragraph{Performance on PALOMA Language Tasks.}
Thirdly, for all model sizes,
we test the next token prediction 
performance of xLSTM, RWKV-4, Llama, and Mamba models 
on PALOMA language tasks~\citep{Magnusson:23arxivshort}.
We measure the performance by the perplexity for next token prediction
on 571 text domains, which range from 
nytimes.com to r/depression on Reddit.
Table~\ref{tab:ppleval} shows token prediction perplexity
grouped into language modeling (first seven columns) and 
fine-grained domain benchmarks (last 5 columns). 
xLSTM[1:0] performs better than xLSTM[7:1] on these language tasks.
%SEPP: do not know whether to report only the xLSTM[1:0] results.
%xLSTM[7:1] has 
%in 446 out of 571 (78.1\%) text domains a lower perplexity than Mamba, 
%in 390 out of 571 (68.3\%) a lower perplexity than Llama,
%in 571 out of 571 (100\%) a lower perplexity than RWKV-4.
xLSTM[1:0] has 
in 568 out of 571 (99.5\%) text domains a lower perplexity than Mamba, 
in 486 out of 571 (85.1\%) a lower perplexity than Llama,
in 570 out of 571 (99.8\%) a lower perplexity than RWKV-4.
For details see Appendix~\ref{sec:appExpLanguage}. 

\begin{figure}[htp]
    \centering
    \includegraphics[width=1.0\textwidth]{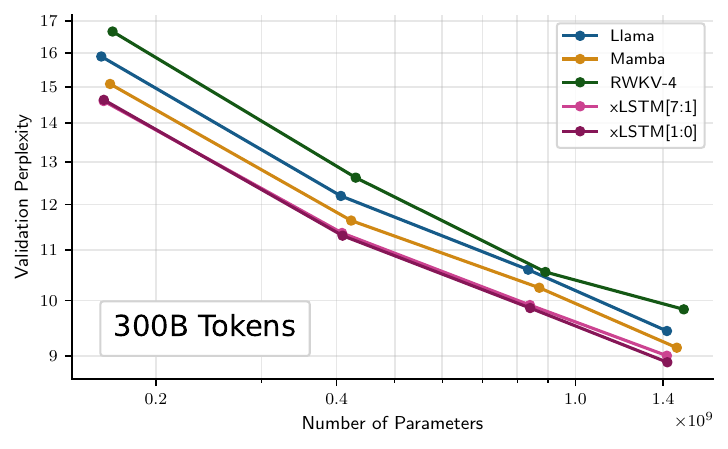}
    \caption{Scaling laws. Next token prediction perplexity of 
    xLSTM, RWKV-4, Llama, and Mamba
    on the SlimPajama validation set
    when trained
    on 300B tokens from SlimPajama. 
    Model sizes are 125M, 350M, 760M, 
    and 1.3B. Best models for each model class, see Table~\ref{tab:spaj15b_model_results}, were selected. The scaling laws indicate that 
    for larger models xLSTM will perform well too.
   \label{fig:temp_sclaw300B}}
\end{figure}

\paragraph{Scaling Laws.}
Fourthly, we assess  
the power-law scaling behavior, which allows to extrapolate the performance to
larger model sizes~\citep{Kaplan:20,Brown:20short}.
Figure~\ref{fig:temp_sclaw300B} presents the scaling behavior.
All models share a similar scaling behavior but with different offsets. 
RWKV-4 performs worst, followed by Llama and Mamba. 
xLSTM is better than Mamba with a similar margin to Mamba as Mamba has to Llama. 
The scaling behavior indicates that for larger models xLSTM will continue to perform favourable compared to Transformers and State-Space models.
% CAMERA-READY ADDED KP
% For a comparison of inference times see Appendix Section~\ref{sec:appInferenceTime}.

\begin{figure}[H]
\centering
    \begin{minipage}{0.51\textwidth}
\includegraphics[width=\textwidth]{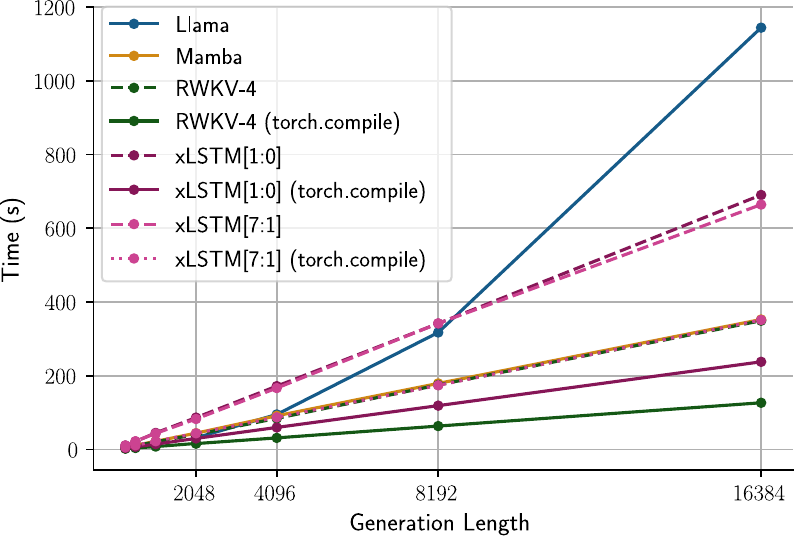}
\end{minipage}
\hspace{0.5em}
\begin{minipage}{0.43\textwidth}
    \includegraphics[width=\textwidth]{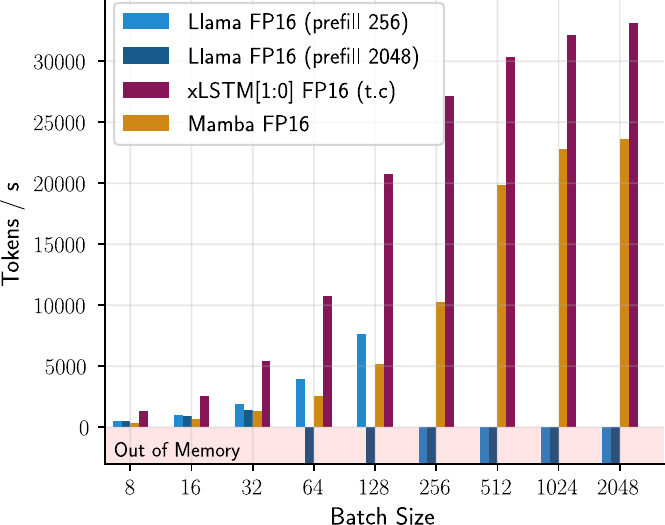}
\end{minipage}
\caption{Inference Generative Speed. \textbf{Left:} Generation times of different 1.3B models for a pre-fill context of 16 tokens (to mitigate cache initialization). The recurrent models (xLSTM[1:0], xLSTM[7:1], Mamba and RWKV-4) show linear behavior, whereas the Transformer (Llama) inference/decoding time is quadratic in sequence length. \label{fig:inference_time_generation}
\textbf{Right:} Token throughput for different batch sizes on a A100-80GB GPU for 1.3B sized models. Note that the Transformer / Llama model goes out of memory (OOM) already for small batch sizes, whereas xLSTM and Mamba can sustain very large batch sizes. xLSTM[1:0]  consistently outperforms Mamba in throughput.
\label{fig:inference_time_decoding_throughput}}
\end{figure}

\textbf{Generation Times and Maximal Throughput.}
Finally, we measure the text generation time in Figure~\ref{fig:inference_time_generation} and the maximal throughput in Figure~\ref{fig:inference_time_decoding_throughput} (left) for our xLSTM variants at 1.3B scale. 
We compare against similar sized Mamba, Llama and RWKV implementions from HuggingFace, including a static key-value cache for the Llama model. 
At the time of the experiments, both full cache compilation of the Transformer Model and compilation of the Mamba model with \texttt{torch.compile} did not work. 
For the text generation experiments all of the models are tested at batch size 1 and pre-fill 16. 
This pre-fill should be maximally favorable for the Transformer.
Figure~\ref{fig:inference_time_generation} shows the linear scaling of the xLSTM and the other recurrent models Mamba and RWKV-4 compared to the quadratic scaling of Llama. 
For the decoding throughput we measure different batch sizes and prefill for the Llama model. 
Figure~\ref{fig:inference_time_decoding_throughput} (right) shows that xLSTM can use much higher batch sizes than Llama due to its constant memory and thus achieves the highest throughput.
% Preliminary results indicate that xLSTM[1:0] inference kernels in Triton can give a further strong improvement.

%##########################################################
%##########################################################
%##########################################################
%##########################################################
%##########################################################
%##########################################################
%##########################################################
%\clearpage

\section{Limitations}

(i) In contrast to mLSTM, memory mixing of the sLSTM prohibits parallelizable operations, and therefore does not allow a fast parallel implementation.
Nevertheless, we developed a fast CUDA kernel for sLSTM, which is currently less than two times slower than our parallel mLSTM implementation.
(ii) The CUDA kernels for mLSTM are not optimized, and therefore the current implementation is about 4 times slower than FlashAttention or
the scan used in Mamba. Faster CUDA kernels could be obtained in the vein of FlashAttention. 
(iii) The matrix memory of mLSTM has high computation complexity since $d \times d$ matrices must be processed. Still, the memory update
and retrieval does not use parameters and can be parallelized using standard matrix operations, therefore the wall clock time overhead due to the complex memory is minor.
(iv) The initialization of the forget gates must be chosen carefully.
(v) Since the matrix memory is independent of the sequence length, increasing the sequence length might overload the memory for longer context sizes. Still, this does not appear to be a limitation for contexts up to 16k, 
see Section~\ref{sec:spaj300B_ctx_extrapolation}.
(vi) Due to the expensive computational load for large language experiments, 
we did neither fully optimize the architecture nor the hyperparameters, especially for larger xLSTM architectures.  
We anticipate that an extensive optimization process is needed for xLSTM to reach its full potential.

%##########################################################
%##########################################################
%##########################################################
%##########################################################
%##########################################################
%##########################################################
%##########################################################

% \newpage

\section{Conclusion}
We have partly answered our simple question: 
How far do we get in language modeling 
when scaling LSTM to billions of parameters?
So far, we can answer: ``At least as far as current technologies 
like Transformers or State Space Models''.
We have enhanced LSTM to xLSTM by exponential gating with memory mixing
and a new memory structure. 
xLSTM models perform favorably on language modeling
when compared to state-of-the-art methods like Transformers and 
State Space Models.
The scaling laws indicate that larger xLSTM models will
be serious competitors to current Large Language Models 
that are built with the Transformer technology.
xLSTM has the potential to considerably impact other fields
like Reinforcement Learning, Time Series Prediction, 
or the modeling of physical systems.
% Max: changed "other deep learning fields" to "other fields"

%\begin{figure}[!b]
%\centering
%\includegraphics[angle=0,width=1.0\textwidth]{./figures/lstm_fig.pdf}
%\caption{LSTM memory cell. 
%Data flows ``feed-forward''
%without delay or ``recurrent'' with a one-step delay.
%``Input'' connections are from the
%external input to the LSTM, while ``recurrent'' connections take inputs
%from the LSTM network itself. Adopted from \citet{Greff:15}. 
%\label{fig:lstmCell}}
%\end{figure}

%\texttt{\{beck, poeppel, klambauer, brandstetter, hochreit \}@ml.jku.at}

\section*{Acknowledgements}
We thank Sebastian Lehner, Daniel Klotz, Thomas Adler, Matthias Dellago, Gerald Gutenbrunner, Fabian Paischer, Vihang Patil, Niklas Schmidinger, Benedikt Alkin, Kajetan Schweighofer, Anna Zimmel, Lukas Aichberger, Lukas Hauzenberger, Bernhard Schäfl and Johannes Lehner for helpful discussions and feedback.

\bibliography{bibs/xLSTM} 
\bibliographystyle{tmlr}
%\bibliographystyle{iclr2022_conference}

%##########################################################
%##########################################################
%##########################################################
%##########################################################
%##########################################################
%##########################################################
%##########################################################
%##########################################################
%##########################################################
%##########################################################
%##########################################################
%##########################################################
%##########################################################
%##########################################################

\newpage

\begin{appendix}
\resumetocwriting

%\addappheadtotoc

\tableofcontents

\newpage

\section{Extended Long Short-Term Memory}
\label{sec:appxLSTM}

\subsection{Vanilla Long Short-Term Memory Formulation: Vector Notation}
\label{sec:apporgLSTM}
The vanilla LSTM memory cell update rules~\citep{Greff:15} at time step $t$ extend the scalar cell state formulation to a vector of cell states:
\begin{align}
\cellstate{\Bc_t} \ &= \  \gates{\bff_t} \odot \cellstate{\Bc_{t-1}} \ + \ \gates{\bfi_t} \odot \cellstate{\Bz_t}
  &  & &\text{cell state} \\
\Bh_t  \ &= \ \gates{\bfo_t} \odot \tilde{\Bh}_t \ , 
  & \tilde{\Bh}_t \ &= \ \psi \left( \cellstate{\Bc_t}\right)
  &\text{hidden state} \\
\cellstate{\Bz_t} \ &= \ \varphi \left( \tilde{\Bz}_t \right) \ , 
  &\tilde{\Bz}_t \ &=  \ \BW_{\Bz} \ \Bx_t \ + \
  \BR_{\Bz}  \ \Bh_{t-1} \ + \  \Bb_{\Bz} \ \
  &\text{cell input} \\
\gates{\bfi_t} \ &= \ \sigma \left( \tilde{\bfi}_t  \right) \ , 
  &\tilde{\bfi}_t \ &= \ \BW_{\bfi} \ \Bx_t \ + \
  \BR_{\bfi}  \ \Bh_{t-1} \ + \  \Bb_{\bfi} \ \
  &\text{input gate} \\
\gates{\bff_t} \ &= \ \sigma \left(  \tilde{\bff}_t \right) \ , 
  &\tilde{\bff}_t \ &= \ \BW_{\bff} \ \Bx_t  \ + \
  \BR_{\bff}  \ \Bh_{t-1} \ + \  \Bb_{\bff} \ \
  &\text{forget gate} \\
\gates{\bfo_t} \ &= \ \sigma \left( \tilde{\bfo}_t \right) \ , 
  &\tilde{\bfo}_t  \ &= \ \BW_{\bfo} \ \Bx_t \ + \
  \BR_{\bfo}  \ \Bh_{t-1} \ + \  \Bb_{\bfo} \ \
  &\text{output gate} 
\end{align}

The matrices $\BW_{\Bz}$, $\BW_{\bfi}$,
$\BW_{\bff}$, and $\BW_{\bfo}$ correspond to the
input weights between inputs $\Bx_t$ and 
cell input, input gate, forget gate, and
output gate, respectively.
The matrices $\BR_{\Bz}$, $\BR_{\bfi}$,
$\BR_{\bff}$, and $\BR_{\bfo}$ correspond to 
the recurrent weights between hidden state $\Bh_{t-1}$
and cell input, input gate, forget gate, and
output gate, respectively.
$\Bb_{\Bz}$, $\Bb_{\bfi}$,
$\Bb_{\bff}$, and $\Bb_{\bfo}$ are the corresponding bias vectors.
$\varphi$ and $\psi$ are the cell input and hidden state 
activation functions (typically $\tanh$). 
$\psi$ is used to normalize or squash the cell state, which would be
unbounded otherwise.

\subsection{sLSTM}
\label{sec:appsLSTM}

Similar to the LSTM in Section~\ref{sec:apporgLSTM}, also the sLSTM can be vectorized to multiple cells:
\begin{align}
\cellstate{\Bc_t} \ &= \  \gates{\bff_t} \odot \cellstate{\Bc_{t-1}} \ + \ \gates{\bfi_t} \odot \cellstate{\Bz_t}
&  & &\text{cell state} \\
\cellstate{\Bn_t} \ &= \  \gates{\bff_t} \odot \cellstate{\Bn_{t-1}} \ + \ \gates{\bfi_t}
  &  & &\text{normalizer state} \\
\Bh_t  \ &= \ \gates{\bfo_t} \odot \tilde{\Bh}_t \ , 
  & \tilde{\Bh}_t \ &= \ \cellstate{\Bc_t}  \odot \cellstate{\Bn_t ^{-1}}
  &\text{hidden state} \\
\cellstate{\Bz_t} \ &= \ \varphi \left( \tilde{\Bz}_t \right) \ , 
  &\tilde{\Bz}_t \ &=  \ \BW_{\Bz} \ \Bx_t \ + \
  \BR_{\Bz}  \ \Bh_{t-1} \ + \  \Bb_{\Bz} \ \
  &\text{cell input} \\
\gates{\bfi_t} \ &= \ \expGate{\! \exp} \left( \tilde{\bfi}_t  \right) \ , 
  &\tilde{\bfi}_t \ &= \ \BW_{\bfi} \ \Bx_t \ + \
  \BR_{\bfi}  \ \Bh_{t-1} \ + \  \Bb_{\bfi} \ \
  &\text{input gate} \\
\gates{\bff_t} \ &= \ \expGate{\! \exp} \left( \tilde{\bff}_t \right) \ \text{OR} \ \sigma \left(  \tilde{\bff}_t \right) \ , 
  &\tilde{\bff}_t \ &= \ \BW_{\bff} \ \Bx_t  \ + \
  \BR_{\bff}  \ \Bh_{t-1} \ + \  \Bb_{\bff} \ \
  &\text{forget gate} \\
\gates{\bfo_t} \ &= \ \sigma \left( \tilde{\bfo}_t \right) \ , 
  &\tilde{\bfo}_t  \ &= \ \BW_{\bfo} \ \Bx_t \ + \
  \BR_{\bfo}  \ \Bh_{t-1} \ + \  \Bb_{\bfo} \ \
  &\text{output gate} 
\end{align}

Here, the cell input activation function $\varphi$ is $\tanh$, the hidden state activation function is the identity. $\varphi$ helps stabilizing the recurrence.

Considering external gradient contribution $\delta^\text{ext}_{\Bh_t}$ from subsequent layers and recurrent gradient contribution $\delta^{\BR}_{\Bh_t}$ from gradients from future states flowing over the cell interaction matrix $\BR$, we obtain the recursive backward pass of sLSTM, where $\delta_a$ indicates gradients with respect to parameter / internal variable $a$:

\begin{align}
	\delta_{\Bh_t} \ &= \ \delta^{ext}_{\Bh_t} \ + \ \delta^{\BR}_{\Bh_t}  &\\
	\delta_{\Bc_{t-1}} \ &= \ \bff_{t} \odot \delta_{\Bc_{t}} \ + \ \bfo_{t-1}  \odot {\Bn_{t-1}}^{-1} \odot \delta_{\Bh_{t-1}} &  \\
		\delta_{\Bn_{t-1}} \ &= \ \bff_{t} \odot \delta_{\Bn_{t}} \ - \ \bfo_{t-1} \odot \Bc_{t-1} \odot \Bn_{t-1}^{-2}  \odot \delta_{\Bh_{t-1}} & \\	
\delta_{\tilde{\bff}_{t}} \ &= \ \bff'_{t} \odot \Bc_{t-1} \odot \delta_{\Bc_{t}}  \ + \ \bff'_{t} \odot \Bn_{t-1} \odot \delta_{\Bn_{t}}  & \\
\delta_{\tilde{\bfi}_{t}} \ &= \ \bfi'_{t} \odot \Bz_{t} \odot \delta_{\Bc_{t}} \ + \  \bfi'_{t} \odot \delta_{\Bn_{t}} & \\
\delta_{\tilde{\bfz}_{t}} \ &= \ \bfi_{t} \odot \varphi'(\tilde{\bfz}_t) \ \odot \delta_{\Bc_{t}} & \\
\delta_{\tilde{\bfo}_{t}} \ &= \ \bfo'_{t} \odot \Bc_{t} \odot \Bn^{-1}_{t} \odot \delta_{\Bh_{t}}  & \\
\delta_{\bfx_t} \ &= \ \sum_{\bfg \in \{\bff, \bfi, \bfz, \bfo\}} \BW^{\top}_{\bfg} \delta_{\tilde{\bfg}_t} & \\
\delta^{\BR}_{\Bh_{t-1}} \ &= \ \sum_{\bfg \in \{\bff, \bfi, \bfz, \bfo\}} \BR^{\top}_{\bfg} \delta_{\tilde{\bfg}_{t}} &\\
\delta_{\BR_\bfg}^\top \ &= \ \sum_t \Bh_{t-1} \delta_{\tilde{\bfg}_{t}}^\top \ , & \bfg \in \{\bfi, \bff, \bfz, \bfo\}\\
\delta_{\BW_\bfg}^\top \ &= \ \sum_t \Bx_{t} \delta_{\tilde{\bfg}_{t}}^\top \ , &  \bfg \in \{\bfi, \bff, \bfz, \bfo\}
\end{align}

with the derivatives of the respective gate activation function $\bfi'_t = \exp' (\tilde{\bfi}_t ) = \exp (\tilde{\bfi}_t ) = \bfi_t $, $\bfo'_t = \sigma' (\tilde{\bfo}_t) $, and $\bff'_t = \sigma'(\tilde{\bff}_t) $ or $\bff'_t = \bff_t$ depending on the forget gate activation. $\varphi'(z)$ is the derivative of the cell input activation function $\varphi(z)$.

The matrices $\BR_{\Bz}$, $\BR_{\bfi}$, $\BR_{\bff}$, $\BR_{\bfo}$ are block-diagonal which is analogous to multiple heads in the mLSTM. This way, the parameters reduce to $d^2/(N_h)$, where $N_h$ is the number of heads, limiting the cell interactions to individual heads. This parameter efficient formulation of cell interactions together with the exponential gating is called the new memory mixing. Finally, to stabilize the backward pass, we clip the magnitude of $\delta^{\BR}_{\Bh_t}$ to $10$, as a means to prohibit exploding gradients for long context lengths.

\paragraph{Proof of Equivalence for sLSTM Stabilized Version.} The stabilization state $m$, see Equation~\eqref{eq:slstmstabil} in the main paper, has no gradient, and hence does not influence the other gradients. We go back to the scalar version (Equation~\ref{eq:slstmforward}) here for simplicity. We re-define $c^{(s)}_t$ and $n^{(s)}_t$ as stabilized cell and normalizer states:
	\begin{align}
		c_t \ &= \ c^{(s)}_t \exp \left(\cellstate{ m_t }\right)	 \\
		n_t \ &= \ n^{(s)}_t \exp \left(\cellstate{m_t} \right)	
	\end{align}
	Inserting Equation~\ref{eq:slstmstabil} into Equation~\ref{eq:slstmforward} yields:
	\begin{align}
		\tilde{h}^{(s)}_t \ &= \ c^{(s)}_t / n^{(s)}_t =
        \\
		&= \ \frac{ \exp \left( \log \left( \Rf_t \right) + \cellstate{m_{t-1}} - \cellstate{m_t} \right) c^{(s)}_{t-1} + \exp \left( \log \left ( \Ri_t \right) - \cellstate{m_t} \right) z_t }{\exp \left( \log \left( \Rf_t \right) + \cellstate{m_{t-1}} - \cellstate{m_t} \right) n^{(s)}_{t-1} + \exp \left( \log \left( \Ri_t \right) - \cellstate{m_t} \right) } 
  \\
		&= \ \frac{ \exp \left( \log \left( \Rf_t \right) + \cellstate{m_{t-1}}  \right) c^{(s)}_{t-1} + \exp \left( \log \left( \Ri_t \right) \right) z_t }{\exp \left( \log \left( \Rf_t \right) + \cellstate{m_{t-1}}  \right) n^{(s)}_{t-1} + \exp \left( \log \left( \Ri_t \right) \right) } 
  \\
		&= \ \frac{ \exp \left( \log \left( \Rf_t \right)  \right) c_{t-1} + \exp \left( \log \left( \Ri_t \right) \right) z_t }{\exp \left( \log \left( \Rf_t \right)  \right) n_{t-1} + \exp \left( \log \left( \Ri_t \right) \right) } 
  \\
		&= \ \frac{ \Rf_t c_{t-1} + \Ri_t z_t }{ \Rf_t  n_{t-1} +  \Ri_t } \ = \ c_t / n_t \ = \ \tilde{h}_t
	\end{align}

Therefore, since the loss solely depends on $h_t$, there's no dependency on $m_t$, and consequently, no gradient exists for this stabilization state.
Note that $m_t$ can be chosen arbitrarily. We choose $m_t = \max \left( \log \left( \bff_t \right) + m_{t-1}, \log \left( \bfi_t \right) \right) $, which stabilizes the exponential function. One can even find $m_t$, such that the normalizer state $n_t$ can be eliminated, but this version was experimentally found to be numerically unstable in the backward pass. 

\subsection{mLSTM}
\label{sec:appmLSTM}

Throughout this section, $\BOn \in \dR^{T}$ denotes a column vector of ones and $\BOn^\top \in \dR^{1 \times T}$ a row vector of ones, where $T$ is the dimension of this vector.

\paragraph{Recurrent mLSTM Backward Pass.}
The recurrent formulation of the mLSTM cell in Equation~\ref{eq:mlstm_recurrent_begin} yields the following backward pass recurrence, where $\delta_a$ indicates gradients with respect to parameter or internal variable $a$ and $\delta^\text{ext}_{\Bh_t}$ denotes gradients from subsequent layers:

	\begin{align}
    \label{eq:mlstm_recurrent_backward}
		\delta_{\tilde{\Bh}_t} &= \bfo_t \odot \delta_{\Bh_t}^{\text{ext}} & \\
		\delta_{\BC_{t-1}}^{\top} &= \Rf_{t} \delta_{\BC_{t}}^{\top} + \frac{  \Bq_{t-1} \delta_{\tilde{\Bh}_{t-1}}^\top }{ \max \left\{ \ABS{\Bn_{t-1}^\top \Bq_{t-1}}, 1 \right\}} & \\
		\delta_{\Bn_{t-1}} &= \Rf_{t} \delta_{\Bn_{t}} - \frac{\Bq_{t-1}^\top \BC_{t-1}^\top  \delta_{\tilde{\bfh}_{t-1}}}{ \max \left\{ \ABS{\Bn_{t-1}^\top \Bq_{t-1}}, 1 \right\}^2} \Omega \left( \Bn_{t-1}^\top \Bq_{t-1} \right) \Bq_{t-1} & \\
        \delta_{\Bv_t}^{\top} &= \Ri_t \Bk_{t}^{\top} \delta_{\BC_t}^{\top} & \\
		\delta_{\Bk_t}^\top &= \Ri_t \left( \Bv_{t}^\top \delta_{\BC_t} + \delta_{\Bn_t}^\top \right) & \\
		\delta_{\Bq_t} &= \frac{\BC_t^\top \delta_{\tilde{\Bh}_t}}{ \max \left\{ \ABS{\Bn_t^\top \Bq_t}, 1 \right\}} - \frac{\Bq_t^\top \BC_t^\top \delta_{\tilde{\Bh}_t}}{ \max \left\{ \ABS{\Bn_t^\top \Bq_t}, 1 \right\}^2} \Omega \left(\Bn_t^\top \Bq_t \right) \Bn_t & \\
		\delta_{\Bx_t} &= \sum_{g \in \{q, k, v\}} \BW^\top_g \delta_{\Bg_t} & \\
		\delta_{\BW_g}^\top &= \sum_t \Bx_t \delta_{\Bg_t}^\top \ , & g \in \{q, k, v\}\\
		\delta_{\Bb_g} &= \sum_t \delta_{\Bg_t} \ , & g \in \{q, k, v\} \\
        \delta_{\tilde{\Rf}_t} &= \left( \BOn^\top \left( \BC_{t-1} \odot \delta_{\BC_{t}} \right) \BOn + \BOn^\top \left( \Bn_{t-1} \odot \delta_{\Bn_{t}} \right) \right) \gamma \left(\tilde{\Rf}_t \right) \\
        \delta_{\tilde{\bfi}_t} &= \left( \BOn^\top \left( \left( \Bv_t \Bk^\top_t \right) \odot \delta_{\BC_{t}} \right) \BOn + \BOn^\top \left( \Bk_{t} \odot \delta_{\Bn_{t}} \right) \right) \exp \left( \tilde{\Ri}_t \right) \\
        \delta_{\tilde{\bfo}_t} &= \tilde{\Bh}_t \odot \sigma' \left( \tilde{\Ro}_t \right) \odot \delta_{\Bh_t}
	\end{align}
	
	and $\Omega \left( z \right) = \Theta \left( z - 1 \right) - \Theta \left( - z - 1 \right)$, $\Theta \left( z \right)$ being the Heaviside step function. $\gamma \left( z \right)$ is either $\sigma' \left( z \right)$ or $\exp \left( z \right) $, depending on the forget gate activation. 

\paragraph{Parallel mLSTM Forward Pass.}
\label{sec:appmLSTMparallel}
The mLSTM recurrence in Equations~(\ref{eq:mlstm_recurrent_begin}-\ref{eq:mlstm_recurrent_end}) can be reformulated in a parallel form, which is used to speed up training.
After training we can still use the recurrent formulation for fast text generation.
% While this is not important for decoding that can only happen recurrently, it is very important during training. 

% Max: Sorry, we cannot use "capital" T for sequence length.. This is too close to transposed.. I move everything back to capital S. I also have never seen a
% Max: Another suggestion would be "capital" N (used in bishop book and understanding deep learning book). But for now I will leave S.

Instead of processing each input $\Bx_t \in \dR^d$ at time step $t$ sequentially,
the parallel version processes all timesteps of a full sequence $\BX \in \dR^{T \times d}$ at once, where $T$ is the sequence length and $d$ is the head dimension. 
We present the forward pass of the mLSTM for a single head and drop the head dimension for simplicity. %GK{Dimension still $S$ -- switch 
%to $T$!}

% Max: Should we add the weights and biases + pre-activation computation so that the forward pass starts with $\BX$? For now I left this out. Can change this if requested.

Let $\tilde{\bm{\Rf}} \in \dR^T$ be the forget gate pre-activations and $\tilde{\bm{\Ri}} \in \dR^T$ be the input gate pre-activations for a full sequence.
We construct the forget gate activation matrix $\bm{\rF} \in \dR^{T \times T}$ by
\begin{equation}
    \bm{\rF}_{ij} = \begin{cases}
        0                                                  & \text{for} \; i < j \\ %j > i \\
        1                                                  & \text{for} \; i = j \\ %j = i \\
        \prod_{k=j+1}^{i} \sigma\left(\tilde{\Rf}_k\right) & \text{for} \; i > j %j < i
    \end{cases} \ ,
\end{equation}
and the input gate pre-activation matrix $\tilde{\bm{\rI}} \in \dR^{T \times T}$ by
\begin{equation}
    \tilde{\bm{\rI}}_{ij} = \begin{cases}
        0     & \text{for} \; i < j \\ %j > i    \\
        \tilde{\Ri}_j & \text{for} \; i \geq j %j \leq i \\
    \end{cases} \ .
\end{equation}

By applying the elementwise exponential input gate activation function naively, we obtain the unstabilized gate activation matrix $\bm{\rD} \in \dR^{T \times T}$ as
\begin{equation}
    \bm{\rD} = \bm{\rF} \odot \exp(\tilde{\bm{\rI}}) \ .
\end{equation}

In order to avoid overflow due to the exponential function we apply the same stabilization as in the recurrent sLSTM, see Equation~\ref{eq:slstmstabil}.
In the parallel formulation of the mLSTM we get a numerically stable gate activation matrix $\bm{\rD}' \in \dR^{T \times T}$
by taking the logarithm of $\bm{\rD}$ element-wise and subtracting the row-wise maximum value of $\bm{\rD}$ from each element:
\begin{align}
    \widetilde{\bm{\rD}} & = \log \bm{\rD} = \log \left(\bm{\rF} \odot \exp(\tilde{\bm{\rI}})\right) = \log \bm{\rF} + \tilde{\bm{\rI}} \\
    \bm{\rD}'            & = \exp(\widetilde{\bm{\rD}} - \max \widetilde{\bm{\rD}})
\end{align}

Given the queries, keys and values $\BQ, \BK, \BV \in \dR^{T \times d}$, for a full sequence we can compute all hidden pre-activation states $\widetilde{\bm{\rH}} \in \dR^{T \times d}$
in parallel for the un-stabilized version by
\begin{equation}
    \widetilde{\bm{\rH}} = \BC \, \BV \ , \quad \text{with} \, \, \, \BC = \frac{\widetilde{\BC}}{\max \left( | \sum_{j=1}^{T} \widetilde{\BC}_{ij}|, 1\right)} \ , \quad 
    \text{and} \, \, \, \widetilde{\BC} = \frac{\BQ \BK^\top }{\sqrt{d}} \odot \bm{\rD} \ .
\end{equation}
Note that we extract the $\frac{1}{\sqrt{d}}$ factor for $\BK$ explicitly here and further on. For the stabilized version this yields
\begin{equation}
    \widetilde{\bm{\rH}} = \BC \, \BV \ , \quad \text{with} \, \, \, \BC = \frac{\widetilde{\BC}'}{\max \left( | \sum_{j=1}^{T} \widetilde{\BC}'_{ij}|, \exp(-\max \widetilde{\bm{\rD}})\right)} \ , \quad 
    \text{and} \, \, \, \widetilde{\BC}' = \frac{\BQ \BK^\top}{\sqrt{d}} \odot \bm{\rD}' \ , 
\end{equation}
where for both versions the hidden pre-activation states $\widetilde{\bm{\rH}}$ are identical.

With the output gate pre-activations $\widetilde{\bm{\rO}} \in \dR^{T \times d}$ we can compute the hidden states $\BH \in \dR^{T \times d}$ for all timesteps 
by applying the output gate in parallel for each timestep element-wise:
\begin{equation}
    \bm{\rH} = \sigma(\widetilde{\bm{\rO}}) \odot \widetilde{\bm{\rH}} \ .
\end{equation}

This gives the parallel forward pass of the mLSTM for a full input sequence $\BX \in \dR^{T \times d}$.

\paragraph{Parallel mLSTM Backward Pass.}

We present the backward pass of the mLSTM for the stabilized version only.
For completeness we summarize the forward pass in the stabilized version before we present the backward pass. 

Given the forget gate matrix $\bm{\rF} \in \dR^{T \times T}$, 
the logarithm of the forget gate matrix $\overline{\bm{\rF}} = \log \bm{\rF} \in \dR^{T \times T}$, 
and the input gate matrix $\bm{\rI} \in \dR^{T \times T}$ as introduced above,
together with the queries, keys and values $\BQ, \BK, \BV \in \dR^{T \times d}$, we can write the forward pass of the mLSTM 
in the stabilized version as:
\begin{alignat}{2}
    \widetilde{\bm{\rD}} &= \overline{\bm{\rF}} + \tilde{\bm{\rI}} \\
    \Bm &= \max_j \widetilde{\bm{\rD}}_{ij} \ , \quad &\text{row-wise maximum} \\
    \bm{\rD}' &= \exp(\widetilde{\bm{\rD}} - \Bm \, \BOn^\top) \\
    \widetilde{\BC}' &= \frac{\BQ \BK^\top}{\sqrt{d}} \odot \bm{\rD}' \\
    \Bb &= \sum_{j=1}^{T} \widetilde{\BC}'_{ij} = \widetilde{\BC}' \ \BOn \ , \quad &\text{row-wise sum} \\ 
    \Bn &= \max \left( |\Bb|, \exp(-\Bm) \right) \\
    \BC &= \widetilde{\BC}' \odot \left( \Bn^{-1} \ \BOn^\top \right) \\
    \widetilde{\bm{\rH}} &= \BC \ \BV
\end{alignat}

With this forward pass we can compute the gradients $\delta_a$ for all intermediate and input variables to the mLSTM forward pass 
in the backward pass. We denote the gradient with respect to variable $a$ as $\delta_a$.

Given the output gradient $\delta_{\widetilde{\bm{\rH}}} \in \dR^{T \times d}$ we can compute the backward pass for the intermediate gradients as:
\begin{alignat}{2}
    \delta_{\BC}^\top &= \BV \delta_{\widetilde{\bm{\rH}}}^\top \\
    \delta_{\Bn} &=-\left(\widetilde{\BC}' \odot \left(\Bn^{-2} \ \BOn^\top \right)\odot \delta_{\BC} \right) \ \BOn \\
                 &= - \left( \left( \widetilde{\BC}' \odot \delta_{\BC} \right) \BOn \right) \odot \Bn^{-2} \\
    \delta_{\Bb} &= \sign \left( \Bn \right) \odot \delta_{\Bn} \odot \begin{cases}
        1 & \text{if} \; |\Bb| > \exp(-\bm{\Rm}) \\
        0 & \text{otherwise}
    \end{cases} \\
    \delta_{\widetilde{\BC}', \BC} &= \left(\Bn^{-1} \ \BOn^\top \right) \odot \delta_{\BC} \ , \quad &\text{column-wise broadcast} \\
    \delta_{\widetilde{\BC}', \Bb}^\top &= \BOn \ \delta_{\Bb}^\top \ , \quad &\text{column-wise broadcast} \\
    \delta_{\widetilde{\BC}'} &= \delta_{\widetilde{\BC}', \BC} + \delta_{\widetilde{\BC}', \BB} \\
    \delta_{\bm{\rD}'} &= \frac{\BQ \BK^\top}{\sqrt{d}} \odot \delta_{\widetilde{\BC}'} \\
    \delta_{\widetilde{\bm{\rD}}} &= \exp(\widetilde{\bm{\rD}} - \Bm) \odot \delta_{\bm{\rD}'} = \bm{\rD}' \odot \delta_{\bm{\rD}'}
\end{alignat}
We do not compute the gradients for $\Bm$ as they cancel out (see the proof in the recurrent sLSTM).

With these intermediate gradients the gradients for the logarithmic forget gate matrix $\delta_{\overline{\bm{\rF}}} \in \dR^{T \times T}$, the input gate matrix $\delta_{\bm{\rI}} \in \dR^{T \times T}$, 
and the queries, keys and values $\delta_{\BQ}, \delta_{\BK}, \delta_{\BV} \in \dR^{T \times d}$ are given by
\begin{alignat}{1}
    \delta_{\overline{\bm{\rF}}} &= \delta_{\widetilde{\bm{\rD}}} \\
    \delta_{\bm{\rI}} &= \delta_{\widetilde{\bm{\rD}}} \\
    \delta_{\BQ} &= \left( \bm{\rD}' \odot \delta_{\widetilde{\BC}'} \right) \frac{\BK}{\sqrt{d}} \\ 
    \delta_{\BK} &= \left( \bm{\rD}' \odot \delta_{\widetilde{\BC}'} \right)^\top \frac{\BQ}{\sqrt{d}} \\
    \delta_{\BV} &= \BC^\top \delta_{\widetilde{\bm{\rH}}}
\end{alignat}

Having computed the gradients for the logarithmic forget gate matrix $\delta_{\overline{\bm{\rF}}}$, 
we can compute the gradients for the forget gate pre-activations $\delta_{\tilde{\bm{\Rf}}} = \left[\delta_{\tilde{\Rf}_1}, \delta_{\tilde{\Rf}_2}, ..., \delta_{\tilde{\Rf}_T} \right]^\top \in \dR^T$.

Recall the logarithmic forget gate matrix $\overline{\bm{\rF}} = \log \bm{\rF}$ is computed by
\begin{equation}
    \overline{\bm{\rF}}_{ij} = \log \bm{\rF}_{ij} = \begin{cases}
        -\infty                                                  & \text{for} \; i < j \\ % j > i \\
        0 & \text{for} \;  i = j \\ % j = i \\
        \sum_{k=j+1}^{i} \underbrace{\log \sigma\left(\tilde{\Rf}_k\right)}_{=: \overline{\Rf}_k} =  \sum_{k=j+1}^{i} \overline{\Rf}_k & \text{for} \; i > j % j < i
    \end{cases} \ .
\end{equation}

With the substitution $\overline{\bm{\Rf}} = \log \sigma(\tilde{\bm{\Rf}})$ we compute the gradients 
for the logarithmic forget gate activations $\delta_{\overline{\bm{\Rf}}} = \left[\delta_{\overline{\Rf}_1}, \delta_{\overline{\Rf}_2}, ..., \delta_{\overline{\Rf}_T} \right]^\top \in \dR^T$ 
as 
\begin{align}
    \delta_{\overline{\Rf}_k} &= \sum_{j=1}^{k-1} \sum_{i=k}^{T} \left(\delta_{\overline{\bm{\rF}}}\right)_{ij} \ , \\
    \delta_{\tilde{\Rf}_k} &= \sigma(- \ \tilde{\Rf}_k) \cdot \delta_{\overline{\Rf}_k} \ ,
\end{align}

\raggedbottom

where the last equation makes use of the following:
\begin{align}
    \begin{split}
    \frac{\Rd}{\Rd x}\left(\log \sigma(x)\right) &= - \left(1 + \exp(-x)\right)^{-1} \cdot \exp(-x) \cdot (-1) \\
    &= \frac{\exp(-x)}{1+\exp(-x)} = \frac{1}{1+\exp(x)} \\
    &=\sigma(-x)
    \end{split}
\end{align}

Finally, we compute the input gate pre-activations' gradients $\delta_{\tilde{\bm{\Ri}}} = \left[\delta_{\tilde{\Ri}_1}, \delta_{\tilde{\Ri}_2}, ..., \delta_{\tilde{\Ri}_S} \right]^\top \in \dR^T$ 
as the column-wise sum over the rows of the input gate matrix $\delta_{\bm{\rI}}$:
\begin{align} 
    \delta_{\tilde{\Ri}_k} &= \sum_{i=k}^{T} \left(\delta_{\bm{\rI}}\right)_{ik}
\end{align}

This completes the backward pass of the parallel mLSTM for a full input sequence $\BX \in \dR^{T \times d}$.

\nopagebreak
% \vspace*{-8cm}
% \clearpage
% \newpage
% \paragraph{Combination of parallel and recurrent formulation}
% \MB{I would not add this! We did not implement it and made no experiments.}
% \KP{You are right. It is more of an outlook of what is possible - it's not needed.} 
% \MB{Good point this is how we could end this section! Outlook for further efficiency gains is to combine both formulations similar to GLA \& Retention!}
% The speed of the parallel formulation and the memory efficiency of the recurrent formulation can be combined in a chunked version similar to \citet{Yang:23arxiv}. The basic idea is to make maximal use of memory and parallel compute via the parallel formulation for one chunk and to store the final state as a recurrent starting point for the next chunk. This can be balanced for optimal use of the computational resources. 

\subsection{Detailed Block Structure}
\begin{figure}[H]
\centering
\includegraphics[width=0.95\textwidth]{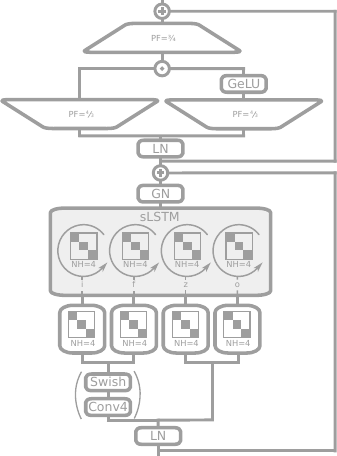}
\caption{Schematic representation of an sLSTM Block -- post up-projection: Embedded in a pre-LayerNorm residual structure, the input is optionally passed through a causal convolution of window size $4$ that includes a Swish activation for input and forget gates. Then, for all input, forget and output gates $\Ri$, $\Rf$, $\Ro$, and the cell update $\Rz$ the input is fed through a block-diagonal linear layer with four diagonal blocks or ``heads''. These diagonal blocks coincide with the recurrent gate pre-activations from the last hidden state, which corresponds to an sLSTM with four heads depicted with the circular arrows. The resulting hidden state goes through a GroupNorm layer~\citep{Wu:2018} -- a head-wise LayerNorm for each of the four heads. Finally, the output is up- and down-projected using a gated MLP, with GeLU activation function and projection factor $4 / 3$ to match parameters.}
\label{fig:appsLSTM_detailed}
\end{figure}

\begin{figure}[H]
\centering
\includegraphics[width=0.95\textwidth]{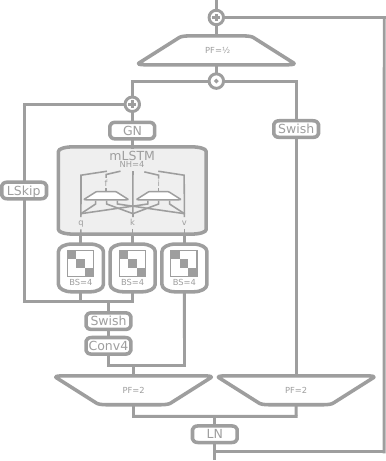}
\caption{Schematic representation of an mLSTM block -- pre up-projection: Embedded in a pre-LayerNorm residual structure, the input is up-projected first with projection factor 2, once for an externalized output gate and once as input for the mLSTM cells. 
The mLSTM cell input is dimension-wise causally convoluted (kernel size 4), before entering a learnable skip connection. 
We obtain input $q$ and $k$ via block-diagonal projection matrices of block size 4. 
The values $v$ are fed directly, skipping the convolution part. 
After the mLSTM sequence mixing, outputs are normalized via GroupNorm~\citep{Wu:2018} -- a head-wise layer norm for each of the four heads. 
Finally, the learnable skip input is added and the result is gated component-wise with the external output gate. 
The output is down-projected.}
\label{fig:appmLSTM_detailed}
\end{figure}

\newpage
\section{Experiments}
\label{sec:appExp}

\paragraph{Training Setup.} For all experiments, we use Python \footnote{\url{https://python.org}} 3.11 with PyTorch 2.2.0 \footnote{\url{https://pytorch.org}}, and CUDA 12.1 \footnote{\url{https://docs.nvidia.com/cuda/archive/12.1.0/}} on NVIDIA A100 GPUs. 
We developed and trained all our models and baselines over the course of three months on a cluster
with 128 nodes of eight NVIDIA A100 GPUs each. More than 95\% of this compute were used for the Language Modeling experiments in Sections~\ref{sec:ExpComparison} and~\ref{sec:ExpLanguage}.
%Markus: Consistently use $1e^{xx}$ or 0.0xxx and xxx000 or xxxk in all experiments
\paragraph{Nearest Neighbor Search Task.} 
\label{sec:appNearestNeighborSearch} 
For this auxiliary task, we use randomly sampled feature vectors of dimension 2 and unit norm. 
The attached value is a uniformly distributed random number from $[0, 1]$, leading to inputs vectors of dimension 3. 
The first feature vector serves as search key, with the first value being ignored. 
Then the model has to predict the value of the nearest neighbor so far in the sequence. 
We train on 8192 sequences of context length up to 64 (uniformly sampled) and validate on 8192 different samples. 
All models have two blocks and embedding dimension 128. 
We use a dropout of 0.1, 10\% linear warm-up steps and cosine decay to 1e-7 for 100k total training steps. 
We sweep over learning rates 1e-4, 1e-3, 1e-2, 1e-1 and 5 seeds each. 
The reported values in Figure~\ref{fig:lstmProblems} are mean values for the best learning rate and 99\% confidence intervals. 
Note that LSTM requires very high learning rates, whereas Transformers (Llama) perform best at the smallest learning rate. 
The xLSTM[0:1] reaches similar performance across all learning rates.

\paragraph{Wikitext-103 Rare Token Prediction.}
\label{sec:appWiki103} 
For this exemplary experiment on rare token prediction, we trained 125M-sized models on Wikitext-103~\citep{Merity:17}. 
All models have an embedding dimension of 768 in a post up-projection structure of 12 residual blocks. 
The Transformer model (Llama) uses Multi-Head Attention, for what is called LSTM the Multi-Head Attention is replaced by an LSTM 
%\MB{"LSTM uses an LSTM network?" what do you mean?} 
and the xLSTM[1:0] contains mLSTM layers with matrix memory. 
Models were trained with maximum learning rate 1e-3, 4k steps linear warm-up and cosine decay for in total 50k steps, using a batch size of 256 and context length of 512. 
We use the validation perplexity as a stopping criterion and evaluate on the test set.

\subsection{Synthetic Tasks and Long Range Arena}
\label{sec:appExpSynthetic}

\subsubsection{Test of xLSTM's Exponential Gating with Memory Mixing.}\label{sec:appExpSynthetic-formalLang}
We evaluate xLSTM on a suite of formal language tasks to test its exponential gating and memory mixing mechanism.

Formal languages provide a framework to probe the generalization capabilities of models.
They allow to specifically test different expressivity levels, e.g.\ along the Chomsky hierarchy.
Typical language model architectures do not necessarily fit perfectly in these hierarchies~\citep{Deletang:23} --- nevertheless these languages allow to illustrate differences in generalization expressivity between different architectures.
Our evaluation tasks are heavily based on the work of \citet{Deletang:23}.

\paragraph{Experiment Setup.}
The different formal language tasks in the experiment (see individual tasks description below) encompass different levels of the Chomsky hierarchy as well as additional counting and memory-focused tasks.
We use different lengths per sample, which allows us to validate in a length extrapolation setting. 
We train on a varying task length up to 40. The evaluation is done for task lengths between 40 and 256 as we are only interested in the ``task generalization capabilities`` of the models.

In all experiments, we use two blocks (or layers for the pure LSTM) for all models. 
We compare Llama, Mamba, Retention, Hyena, RWKV-4, RWKV-5, RWKV-6, LSTM, xLSTM[0:1], xLSTM[1:0] and xLSTM[1:1].
The sLSTM block is used without a convolution and with normal weight initialization.
LSTM (Block) refers to an architecture where a vanilla LSTM is used instead of self-attention inside a Transformer block.

All models are trained with 3 different learning rates (1e-2, 1e-3, 1e-4), each with two seeds. Batch size is 256 --- cosine annealing (min lr: 1e-5) with 10\% warm-up steps is applied.
We use AdamW~\citep{Loshchilov:19} ($\beta_1 = 0.9$, $\beta_2 = 0.99$) and a weight decay of 0.1 for training.
In each experiment we train for 100k steps --- the samples are generated randomly, however, all experiments are trained and evaluated on the same samples.

\begin{figure}
\centering
%\begin{wrapfigure}{r}{0.5\textwidth}
\includegraphics[width=0.65\textwidth]{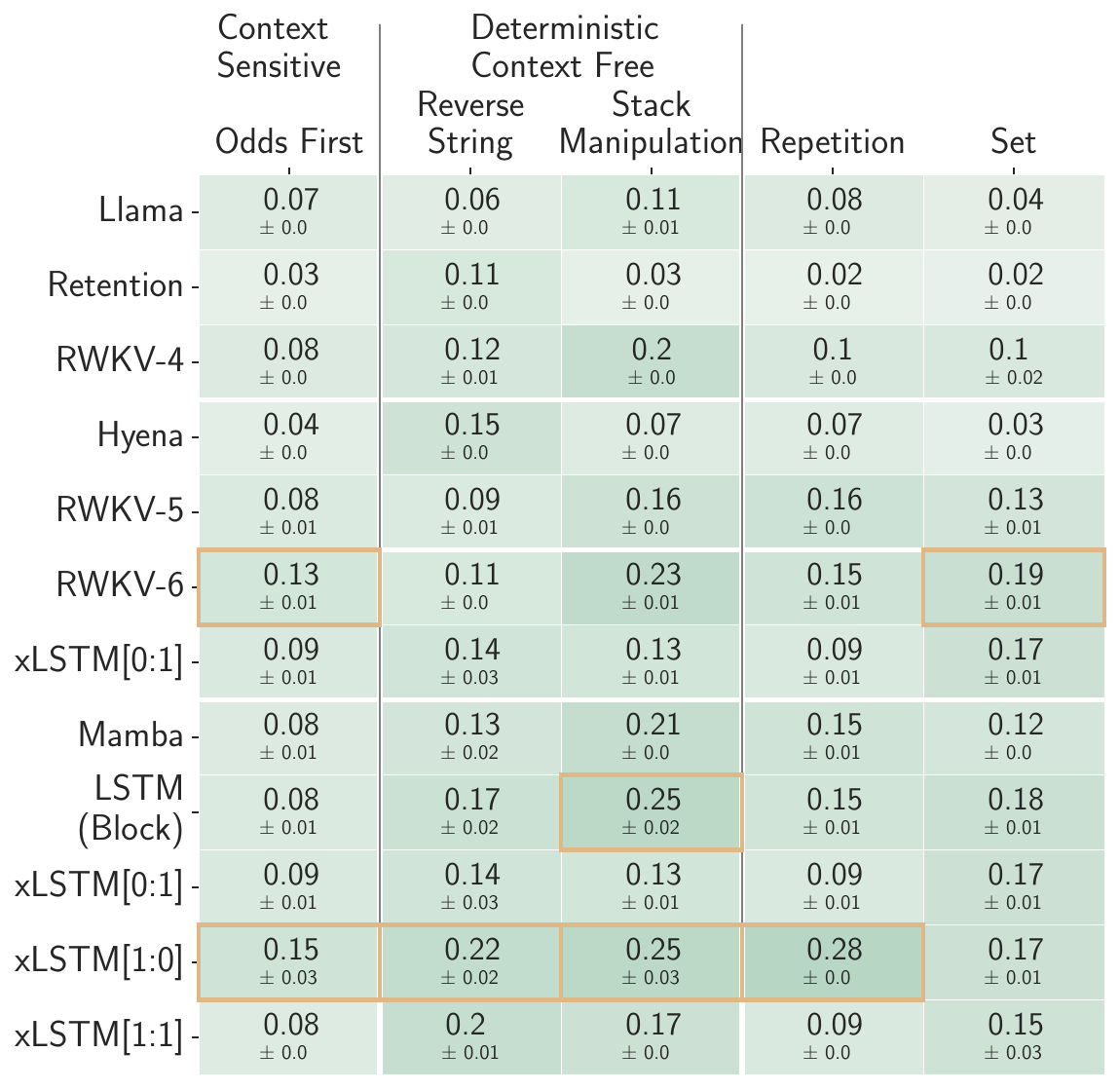}
\caption{Supplementary results given by scaled accuracy of different models at 
   solving formal language tasks. 
   Tasks are grouped by 
   the Chomsky hierarchy. \label{fig:formal-appendix}}
%\end{wrapfigure}
\end{figure}

\paragraph{Additional Formal Language Results.} Figure~\ref{fig:formal-appendix} showcases supplementary results
on formal language task, detailing tasks where no model attained a minimum scaled accuracy of 0.3. Although no model achieves proper extrapolation of the task to a larger context length, xLSTM performs best among the evaluated models.  

\paragraph{Individual Task Description.}
The majority of tasks are based on \citet{Deletang:23}.
We provide the vocabulary size $|V|$ and the random accuracy $s_{rand}$ (for accuracy scaling), used in the evaluation.
As we evaluate different task lengths each task has a padding token which is used to pad the sequence to the given context length.
In Listing~\ref{task:majority} there is an example for each task.

\begin{itemize}
    \item\textbf{Bucket Sort}
    Given a string of tokens of a sorted alphabet, compute the sorted string. \\
    $|V|=11 \quad s_{\text{rand}}=\frac{1}{|V| - 1}$

    \item\textbf{Cycle Nav}
    Given a string of ``movement tokens'' ($+1$, $-1$, STAY) compute the end position of the agent with start position 0. The position must be computed modulo the maximum position. \\
    $|V|=9 \quad s_{\text{rand}}=\frac{1}{|V|-4}$

    \item\textbf{Even Pairs}
    Given a binary string of \textit{a} and \textit{b} tokens, compute whether the number of \textit{ab} and \textit{ba} is even.
    This task can be solved by checking if the first and last token of the string are equal. \\
    $|V|=3 \quad s_{\text{rand}}=0.5$

    \item\textbf{Majority}
    Given a string of tokens, compute the token that occurred most often in the sequence.\\
    $|V|=64 \quad s_{\text{rand}}=\frac{1}{|V| - 1}$
    
    \item\textbf{Majority Count}
    Given a string of tokens of an ordered alphabet. Compute the count of the token that occurred most often in the sequence. If the count exceeds the vocab size, the highest vocab token should be outputted.\\
    $|V|=64 \quad s_{\text{rand}}=\frac{1}{|V| - 1}$

    \item\textbf{Missing Duplicate}
    Given a string of tokens. The string is repeated but one of the tokens is masked in the repetition. Output the token that is masked. \\
    $|V|=11 \quad s_{\text{rand}}=\frac{1}{|V| - 2}$

    \item\textbf{Mod Arithmetic (w/o Brackets)}
    Calculate the result --- modulo the max number --- of the arithmetic operations in the context. The maximum number is the vocabulary size minus the number of special tokens (+,-,*,=, [PAD]). \\
    $|V|=10 \quad s_{\text{rand}}=\frac{1}{|V|-5}$

    \item\textbf{Mod Arithmetic (w Brackets)}
    Calculate the result --- modulo the maximum number --- of the arithmetic operations in the context. The maximum number is vocabulary size minus the number of special tokens (+,-,*,=,(,), [PAD]). \\
    $|V|=12 \quad s_{\text{rand}}=\frac{1}{|V|-7}$

    \item\textbf{Odds First}
    An string of tokens $t_1, t_2, t_3, ...t_n$ is given. Output all tokens with and odd index ($t_1,t_3,...$) then the token with an even index ($t_2$, $t_4$,..) . Apart from that keep the ordering of the initial string. \\
    $|V|=12 \quad s_{\text{rand}}=\frac{1}{|V| - 2}$
    
    \item\textbf{Parity}
    Given a binary string of \textit{a} and \textit{b} tokens, compute if the number of \textit{b}`s is even. If the number is even output \textit{a} otherwise \textit{b}. This is equivalent to sequentially calculating the half-adder sum. \\
    $|V|=3 \quad s_{\text{rand}}=0.5$    

    \item\textbf{Repetition}
    Given a string of tokens --- repeat it. \\
    $|V|=12 \quad s_{\text{rand}}=\frac{1}{|V| - 2}$

    \item\textbf{Reverse String}
    Given a string of tokens --- repeat it in reverse order. \\
    $|V|=12 \quad s_{\text{rand}}=\frac{1}{|V| - 2}$

    \item\textbf{Stack Manipulation}
    An initial stack content is given, followed by a sequence of push and pop operations. Compute the stack content after the operations \\
    $|V|=11 \quad s_{\text{rand}}=\frac{1}{\lfloor\frac{|V| - 3}{2}\rfloor}$

    \item\textbf{Set}
    Given a string of tokens, compute the ordered set of the tokens. Keep the ordering so that tokens that occurred first are also outputted first. \\
    $|V|=128 \quad s_{\text{rand}}=\frac{1}{|V| - 2}$

    \item\textbf{Solve Equation}
    Given is an equation with the operators \{+,-,*,=,(,)\}, number, and an unknown variable x. Compute the value of the variable modulo the max number. The maximum number is vocabulary size minus the number of special tokens (+,-,*,=,(,), [PAD], [ACT]). \\
    $|V|=14 \quad s_{\text{rand}}=\frac{1}{|V|-9}$
\end{itemize}

\clearpage

\begin{lstlisting}[frame=single,escapeinside={<@}{@>},caption={Examples of the formal language tasks. Red tokens are evaluated for loss and accuracy metrics, but are padded for the input. The tokens are illustrated in a way that allows easy semantic interpretation for the given task --- hence, some tokens are represented by multiple characters.},label=task:majority,captionpos=b]
Bucket Sort
  Sequence: 1 4 8 6 1 <@\textcolor{red}{1}@> <@\textcolor{red}{1}@> <@\textcolor{red}{4}@> <@\textcolor{red}{6}@> <@\textcolor{red}{8}@>
Cycle Nav
  Sequence: STAY +1 -1 +1 STAY +1 +1 +1 -1 <@\textcolor{red}{P3}@>
Even Pairs
  Sequence: a b b a a b a b a <@\textcolor{red}{a}@>
Majority
  Sequence: 1 7 6 4 3 8 1 7 2 <@\textcolor{red}{1}@>
Majority Count
  Sequence: 1 7 6 4 4 8 1 7 2 <@\textcolor{red}{2}@>
Missing Duplicate
  Sequence: 4 8 6 2 5 4 8 6 2 [MIS] <@\textcolor{red}{5}@>
Mod Arithmetic (w/o Braces)
  Sequence: 0 - 4 + 0 - 2 = <@\textcolor{red}{4}@> [PAD]
Mod Arithmetic (w Braces)
  Sequence: ( ( ( 2 ) * - 2 ) - ( - 4 - 2 ) ) = <@\textcolor{red}{2}@>
Odds First
  Sequence: 2 7 3 2 6 9 [ACT] <@\textcolor{red}{2}@> <@\textcolor{red}{3}@> <@\textcolor{red}{6}@> <@\textcolor{red}{7}@> <@\textcolor{red}{2}@> <@\textcolor{red}{9}@>
Parity:
  Sequence: a b b a a b a <@\textcolor{red}{b}@>
Repetition
  Sequence: 2 4 8 6 2 [ACT] <@\textcolor{red}{2}@> <@\textcolor{red}{4}@> <@\textcolor{red}{8}@> <@\textcolor{red}{6}@> <@\textcolor{red}{2}@>
Reverse String
  Sequence: 2 4 8 6 2 [ACT] <@\textcolor{red}{2}@> <@\textcolor{red}{6}@> <@\textcolor{red}{8}@> <@\textcolor{red}{4}@> <@\textcolor{red}{2}@>
Stack Manipulation
  Sequence: ST1 ST1 ST3 POP POP PS3 PS3 [ACT] <@\textcolor{red}{ST1}@> <@\textcolor{red}{ST3}@> <@\textcolor{red}{ST3}@>   
Set
  Sequence: 8 6 6 3 5 4 5 3 [ACT] <@\textcolor{red}{8}@> <@\textcolor{red}{6}@> <@\textcolor{red}{3}@> <@\textcolor{red}{5}@> <@\textcolor{red}{4}@>
Solve Equation:
  Sequence: ( ( ( 2 + 0 ) + - x ) - ( 1 ) ) = 2 [ACT] <@\textcolor{red}{2}@>
\end{lstlisting}

\subsubsection{Test of xLSTM's Memory Capacities on Associative Recall Tasks.}\label{sec:appExpSynthetic-mqar}
\begin{figure}[h]
\includegraphics[width=\textwidth]{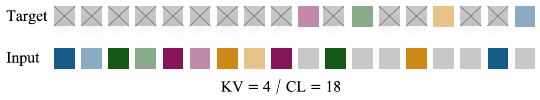}
\caption{Illustration of the MQAR task. Color pairs represent key-value pairs (keys have darker shade). The first part of the sequence defines the key-value pairs for the respective sample. After that, the keys appear randomly according to a power law distribution \protect\footnotemark. Grey tokens in the input sequence represent a zero token. The ``target'' sequence contains the value after the respective key appearance --- the rest of the tokens are ignored for the accuracy and loss calculation. The model must predict the value tokens given the respective key.}
\label{fig:mqr}
\end{figure}

We test the memory capacity of xLSTM with the Multi-Query Associative Recall task proposed by \citet{Arora:23arxiv}.
Figure~\ref{fig:mqr} illustrates the basic task setup.

\paragraph{Why Multi-Query Associative Recall for Memory Tests of LLM Architectures.}
Associative Recall (AR), the ability to retrieve a specific value (information) associated with a given key (information), constitutes a key capability for LLM to perform well~\citep{Poli:24arxiv, Arora:23arxiv, Olsson:22short}.
Especially its quality of in-context learning seems to be strongly connected to this capability~\citep{Olsson:22short}.
\citet{Arora:23arxiv} attribute performance gaps between early non-Transformer and Transformer language models specifically to performance gaps in associative recall.
They argue that prior AR evaluations fall short of capturing these differences and propose MQAR, 
which can show the AR performance differences that translate to performance differences in language modeling performance.
Hence, MQAR is especially suitable to analyze the memory capacity of LLM. 
Transformer (e.g.\ Llama) models can be seen as the gold standard for this task as their memory 
is exponential in the coding dimension~\citep{Ramsauer:21}.
% Auer - Sidenote, probably note relevant in that details
%\citet{Arora:23arxiv} argue that prior AR evaluations fall short in their experiment setup as they (1) only use one query-key pair per sample, (2) at a fixed position in the sequence and (3) only operate on a small vocabulary size (smaller than model/embedding dimension).
%They evaluated the performance gap between Transformers and (early) Transformer alternatives and showed that a large part of the gap can be attributed to shortfalls in AR.
%The Multi-Query Associative Recall (MQAR) task tackles these limitations of previous AR benchmark tasks.

\paragraph{Experiment Setup.}
There are two relevant variables that determine 
different experimental setups. 
(1) \textbf{Context Length (CL)}: Length of the sequence of 
one sample --- this influences the distances between the key--value 
definition and the recall. 
(2) \textbf{Number Key-Value Pairs (KV)}: Influences how many key-value 
pairs the model needs to keep track of. The vocabulary size is always 8192. 

In all experiments, we use two blocks (or layers for the pure LSTM) for all models.
LSTM (Block) model refers to an architecture 
where a vanilla LSTM is used instead of self-attention inside a Transformer block.

For each task setup, we train each model with 4 different learning rates (batch size > 24: \{1e-2, 2.15e-3, 4.6e-4, 1e-4\}, batch size 24: \{1e-3, 2.2e-4, 5e-5, 1e-5\}). The batch size (BS) changes depending 
on the context length (CL) (CL=64/128: BS=512; 
CL=256: BS=256; CL=756: BS=128; 
CL=1024: BS=96; CL=2048: BS=24). 
We vary the embedding dimension \textbf{(Model Dim)} between different experiments 
-- different numbers of heads are used accordingly.
For each experiment, we generate 100,000 training samples (validation: 3,000 samples) 
and train for 64 epochs. 
We apply cosine annealing (min lr: 1e-4 and 1e-5) with 10\% warm-up steps.
We use AdamW~\citep{Loshchilov:19} and a weight decay of 0.1 for training.

We conduct three different experiments:

\begin{itemize}
    \item\textbf{MQAR-Experiment 1} evaluates, in the same fashion as \citet{Arora:23arxiv}, a variety of models (Llama, Mamba, Mamba (noWT) - i.e.\ without weight tying, Retention, Hyena, H3, RWKV-4, RWKV-5, RWKV-6, LSTM, LSTM (Block), xLSTM[0:1], xLSTM[1:0] and xLSTM[1:1]) on increasing task difficulty by increasing the context length and number of key-value pairs simultaneously. We benchmark three parameter settings: CL,KV=\{(64,4),(128,8),(256,16)\}.
    
    \item\textbf{MQAR-Experiment 2} increases the task difficulty notably and goes beyond 
    previous evaluations on this task. 
    We individually scale the context length (CL=\{756, 1024, 2048\})
    and the key-value pairs (KV=\{48, 96, 256\}) and evaluate all combinations.
    This experiment especially probes the memory capacity because the number of key-value pairs is high.
    To reduce the computational burden we only evaluate models that perform flawlessly in Experiment 1 --- 
    additionally we evaluate Transformer only in the hardest setting (CL=2048) as sanity check, 
    because no performance decrease is expected.
    
    \item\textbf{MQAR-Experiment 3} analyzes whether the AR capability learned 
    on a certain context length extrapolates to bigger context lengths. 
    For each KV setting of Experiment 2, we use the models 
    (we select the 3 biggest model dimensions) trained on CL=2048 and
    evaluate bigger context lengths (CL=\{4096, 6144, 8192\}).
\end{itemize}

\paragraph{Extended Results.}

\begin{figure}[htp]
    \centering
    \includegraphics[width=\textwidth]{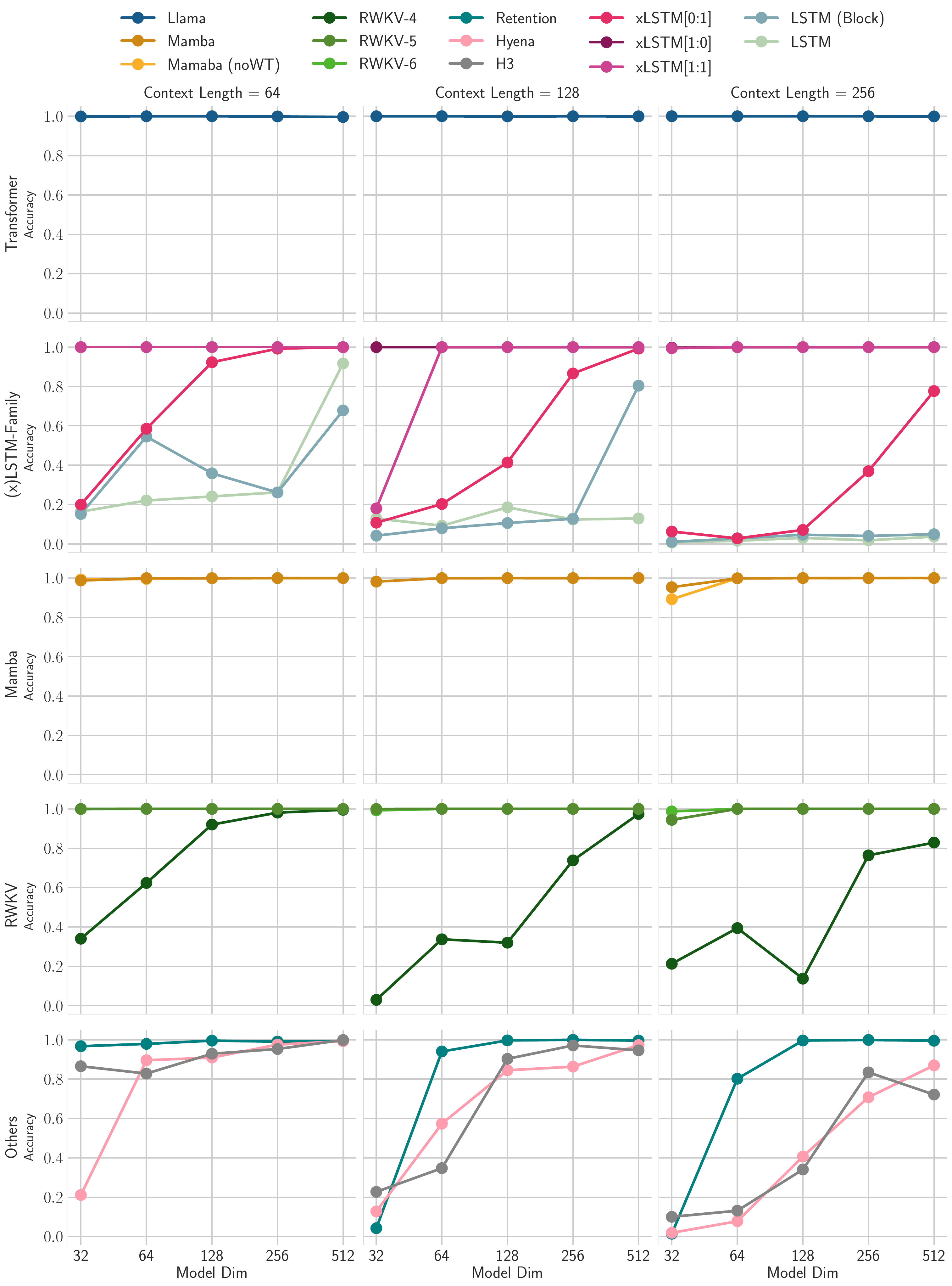}
    \caption{Result of MQAR-Experiment 1. The columns show different task settings (context length and key-value pairs). The rows group related models for better clarity. The $x$-axis gives the model size and 
    the $y$-axis the validation accuracy.}
    \label{fig:mqar-app-small}
\end{figure}

The result of Experiment 1 can be found in Figure~\ref{fig:mqar-app-small}. 
In accordance to the results of \citet{Arora:23arxiv} H3, Hyena, RWKV-4 
fail to solve the task with a smaller model dimension. 
In contrast, xLSTM[1:1], xLSTM[1:0], Mamba, RWKV-5 and RWKV-6 are able 
to solve these settings for all model dimensions.
The comparison of xLSTM[0:1] with both original LSTM variants indicates 
that the exponential gating mechanism improves the AR capabilities of the model.
However, both fall short because of the reduced memory capacity compared to xLSTM[1:1] and xLSTM[1:0].

\begin{figure}[htp]
    \centering
    \includegraphics[width=\textwidth]{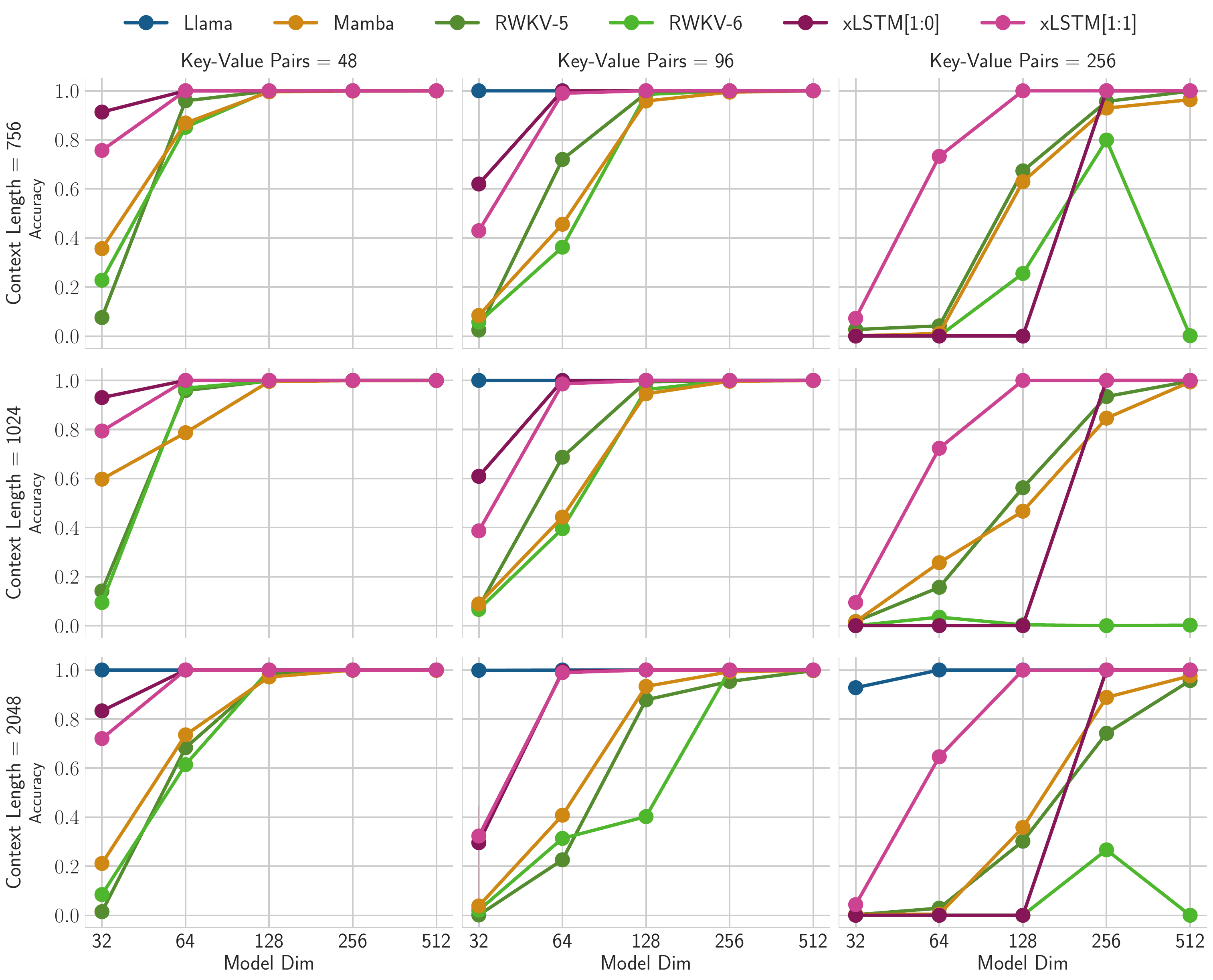}
    \caption{Result of MQAR-Experiment 2. The columns and rows correspond to different numbers of key-value pairs and the context length respectivly. The $x$-axis gives the model size and 
    the $y$-axis the validation accuracy.}
    \label{fig:mqar-app-big}
\end{figure}

The results of Experiment 2 are presented in Figure~\ref{fig:mqar-app-big}.
Scaling the context length has a low impact on the performance of the models. 
However, while xLSTM[1:1] and xLSTM[1:0] show no clear decay, 
both RWKV variants slightly, but consistently lose performance with increasing context lengths.
The varying number of key-value pairs,
which mainly probes the memory capacity of the non-Transformer models, 
has a more notable impact across all models.
RWKV-5 seems to outperform RWKV-6. The latter fails to learn the task at all in some KV=256 settings.
Overall xLSTM[1:1] is the best-performing non-Transformer 
model --- suggesting that it provides enhanced memory capacity, also in long contexts.

\begin{figure}[htp]
    \centering
    \includegraphics[width=\textwidth]{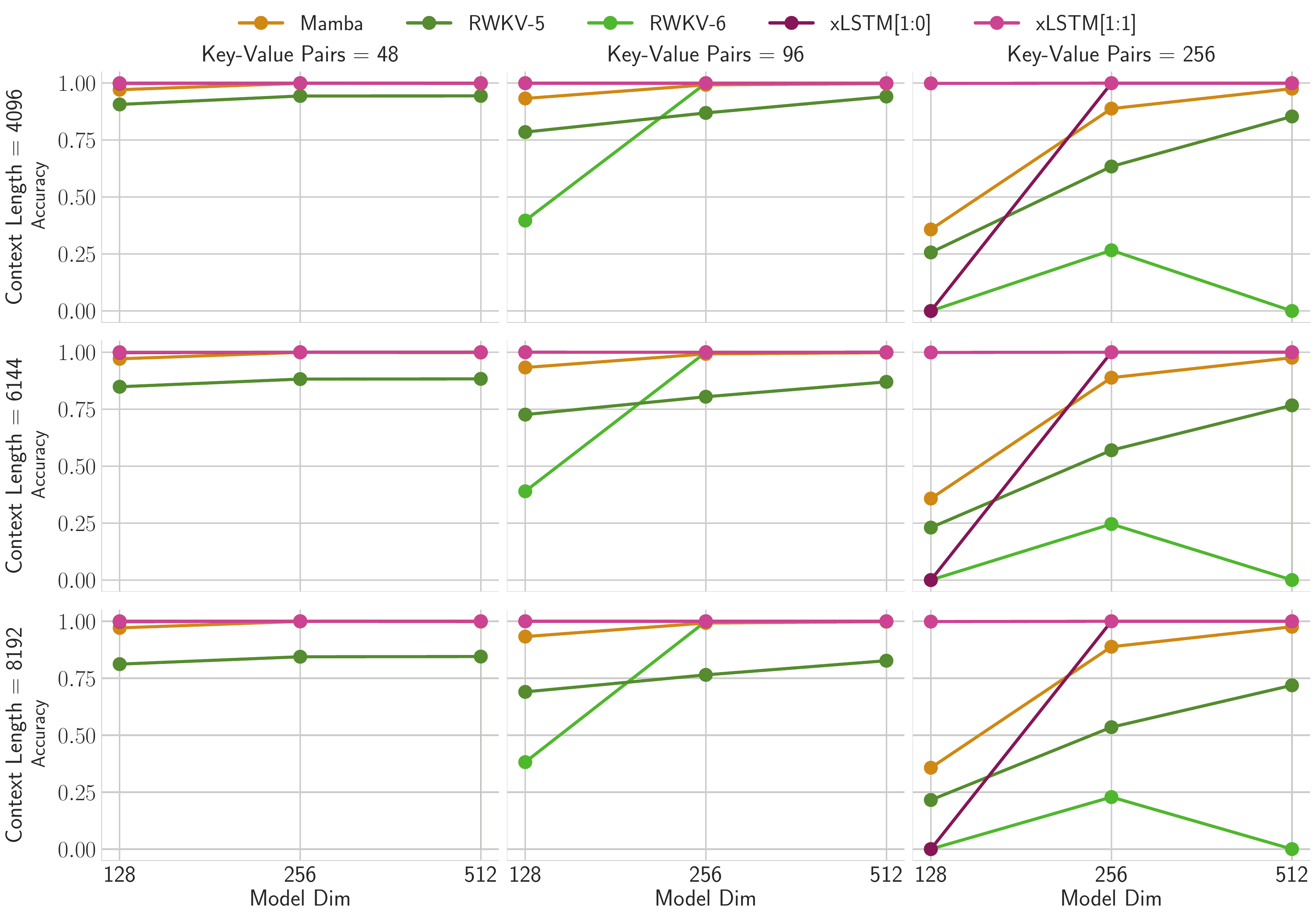}
    \caption{Result of MQAR-Experiment 3 (Extrapolation). All evaluated models were trained on context length 2048 and the number of key-value pairs given by the columns of the plot. The rows show the different context lengths used in the evaluation. The $x$-axis gives the model size and the $y$-axis the validation accuracy. }
    \label{fig:mqar-app-extrapolate}
\end{figure}

Figure~\ref{fig:mqar-app-extrapolate} shows the extrapolation results from Experiment 3.
For xLSTM[1:1], xLSTM[1:0], and Mamba the model performance 
does not change in the extrapolation setting. The RWKV models (especially RWKV5) degrade 
slightly with increasing context length.
xLSTM[1:1] performs best, as it maintains its superior performance of Experiment 2.

\footnotetext{The keys are distributed on the ``evaluation part'' of the sequence given a power-law distribution. This is motivated by similar structures in natural language text.}

\newpage

\subsubsection{Test of xLSTM's Long Range Capabilities on the Long Range Arena.}\label{sec:appExpSynthetic-lra}
We assess the performance of xLSTM across tasks in the Long Range Arena
benchmark~\citep{Tay:21}, examining its ability to effectively handle
longer context lengths and diverse data types.

Our experiments on Long Range Arena benchmark are composed of five tasks:
\begin{itemize}
\item \textbf{Retrieval}: The task is to predict if two documents have a citation link. 
The dataset of text documents is derived from the ACL Anthology Network~\citep{Radev:09}.
\item \textbf{ListOps}: This is a set of modular arithmetic tasks including brackets and lists of numbers, using the operations \texttt{MIN}, \texttt{MAX}, \texttt{MEDIAN} and \texttt{SUMMOD} (modular sum).
A particular example is: \texttt{[MAX 4 3 [MIN 2 3 ] 1 0 [MEDIAN 1 5 8 9, 2]]} $\xrightarrow{}$ \texttt{5}
\item \textbf{Image}: This task is based on a version of the CIFAR dataset~\citep{Krizhevsky:09}, 
where images are transformed to a sequence of pixels and this sequence has to be classified into the usual CIFAR classes. 
We test both a gray-scale (G-Image) and RGB (RGB-Image) version of this dataset, 
as \citet{Orvieto:23} uses colored images contrary to the standard setup.
\item \textbf{Pathfinder}: The input for this task is a 32x32 gray-scale image, given as pixel sequence,
with two dots and several curved lines on it. 
The task is to predict if the two dots are connected by any of the lines~\citep{Linsley:18}.
% \item \textbf{Pathfinder-X} Is a longer version of Pathfinder task, where images have dimension of 128x128.
% \item Byte-level Text classification task utilizes IMDb reviews data \citep{Maas:11} to evaluate models' performance in classifying text at the character level, simulating longer input sequences for document processing.
\end{itemize}
We omit the \textbf{Text} classification task~\citep{Maas:11}, 
as the language modeling experiments already test this kind of data, 
and the \textbf{Pathfinder-X} version of \textbf{Pathfinder}.

\paragraph{Experiment Setup.}

The architectures that are tested in this experiment 
comprise Llama, Mamba, LSTM, RWKV-4, and xLSTM. 
% GUENTER: wasnt there GPT AND LLAMA? Yes, as Andi only did experiments on Lllama, and reported this as Transformer (consistent with the toy tasks from the intro), we now use Transformer here with the LLama specification. The results are not too different.
% Korbinian: After discussions we now call all Llama models Llama, and GPTs GPT 
LSTM (Block) refers to an architecture where a vanilla LSTM is used 
inside a post up-projection block (like Transformer with attention replaced by LSTM).
For xLSTM we choose the best performing of xLSTM[0:1] or xLSTM[1:0] on the validation set, specifically the former for the Image tasks and the latter for all other ones.

We use the hyperparameter settings of the S5 model~\citep{Smith:22} and 
Linear Recurrent Unit model~\citep{Orvieto:23}, with additional hyperparamter search on learning rates and schedulers for all models. 
We use two different schedulers: Linear Warm-up Cosine Annealing and Linear Warm-up Cosine Annealing with Restarts.
Both learning rate schedulers were evaluated with learning rates of 1e-3, 6e-4 and 1e-4. For the second scheduler, the number of restarts ($R$) is set to 3. 
The model hyperparameters for each dataset are displayed in Table~\ref{tab:lra_hyperparameters}.

\begin{table}
    %\begin{adjustbox}{width=1.0\textwidth}
    \centering
        \begin{tabular}[htp]{lcccc}
    \toprule
         \thead{Task} & \thead{\#Blocks} & \thead{Embedding\\Dim} & \thead{Batch \\Size} &  \thead{Training \\ Steps} \\
    \midrule
        Retrieval & 6 & 128 & 64 & 100k \\
        ListOps & 8 & 128 & 32 & 80k \\ 
        Pathfinder & 6 & 192 & 64 & 500k \\ 
        G-Image & 6 & 512 &  64 & 180k \\ 
         RGB-Image & 6 & 512 & 64 & 180k \\ 
    \bottomrule
    \end{tabular}

    % \end{adjustbox}
    \vspace{0.1cm}
    
    \caption{Long Range Arena model hyperparameters. These are the model hyperparameters used in each of the Long Range Arena tasks. For each model we used the best learning rate and the better of the two learning rate schedulers.
    }
    \vspace{0.1cm}
    
    \label{tab:lra_hyperparameters}
\end{table}
% KP: moved to main hyperparam figure
% \input{tables/lra_xlstm_hyperparameters}

\paragraph{Results.}
Table~\ref{tab:lra_test_accuracy} shows the result of experiments on the Long Range Arena benchmark. 
xLSTM demonstrates consistent strong performance on all of the tasks, suggesting that the proposed architecture is remarkably efficient in handling different aspects of long context problems.
\begin{table}
    \centering
    \begin{adjustbox}{width=1.0\textwidth}
    \begin{tabular}{lcccccc}
    \toprule
    & \thead{Retrieval \\ acc $\uparrow$} & \thead{ListOps \\ acc $\uparrow$} & \thead{Pathfinder \\ acc $\uparrow$} & \thead{G-Image \\ acc $\uparrow$} & \thead{RGB-Image \\ acc $\uparrow$} & \thead{Ranking \\ acc $\uparrow$} \\ \midrule
Random Baseline & 0.500 & 0.100 & 0.500 & 0.100 & 0.100 &  \\ \midrule
%GPT & 0.821 & 0.390 & 0.869 & 0.541 & 0.634 & 5.4 \\ 
Llama & 0.845 & 0.379 & 0.887 & 0.541 & 0.629 & 5.2\\ %\midrulePathfinder
Mamba & \underline{0.902} & 0.325 & \textbf{0.992} & 0.689 & \textbf{0.765} & 2.2 \\ %\midrule
RWKV-4 & 0.898 & 0.389 & 0.914 & \underline{0.691} & 0.757 & 3.0\\ 
LSTM & X & 0.275 & X & 0.675 & 0.718 & 5.4 \\ 
LSTM (Block) & 0.880 & \textbf{0.495} & X & 0.690 & 0.756 & 3.4 \\ \midrule
xLSTM & \textbf{0.906} & \underline{0.411} & \underline{0.919} & \textbf{0.695} & \underline{0.761} & 1.6 \\ 
\bottomrule
    \end{tabular}
    
    \end{adjustbox}
    \vspace{0.1cm}
    
    \caption{Long Range Arena test accuracy. Bold highlights the best performing model, underlined the second best. X denotes models that fail to outperform random baselines. xLSTM is the best of xLSTM[1:0], xLSTM[0:1] based on validation dataset accuracy.
    }
    \vspace{0.1cm}
    
    \label{tab:lra_test_accuracy}
    
\end{table}

% \KP{Selected best xLSTM architecture for each task.}
\newpage

\subsection{Method Comparison and Ablation Study on SlimPajama (15B)}

\paragraph{General Training Procedure.}
\label{sec:appGenTrainProc}
We tokenize our datasets using the HuggingFace GPT-2 tokenizer \citep{Radford:19, Brown:20short} \footnote{\url{https://huggingface.co/docs/transformers/en/model\_doc/gpt2}} and use this tokenizer for all models in this paper.
In general, we try to follow \citet{Brown:20short} for the general training setup, i.e. we choose context length 2048 and batch sizes 256 or 512 for our models.
We use the AdamW~\citep{Loshchilov:19} optimizer with beta parameters ($\beta_1$, $\beta_2$)=(0.9, 0.95) and an epsilon parameter of 1e-5.
As learning rate scheduler we use a linear warm-up with 750 steps and cosine decay to 10\% of the peak learning rate. 
We apply a weight decay of 0.1 to all our models and always exclude the token embedding matrix from weight decay.
If not specified otherwise, we do not tie the weights of the token embedding and the language model head.
For parallelization, we use PyTorch FSDP in \texttt{SHARD\_GRAD\_OP} mode with mixed precision in \texttt{bfloat16}, where applicable. 
For small models we use \texttt{NO\_SHARD}.
We keep the weights in \texttt{float32} and reduce the gradients across GPUs in \texttt{float32}.
We use \texttt{torch.compile} to speed up models, except for Transformer models as their training curves did not match the non-compiled versions.
For xLSTM[7:1], we use positions [3, 5, 7, 40, 42, 44] for sLSTM-based blocks, except for the 125M size, where we use [3, 20] (this is actually a [11:1] ratio). 
We do not use any positional encoding for our xLSTM models.

\begin{table}[htbp]
    \centering
    \begin{adjustbox}{width=\textwidth}    
    \begin{tabular}{llrrrrrr}
    \toprule
                                                      & Model  & \thead{EmbeddingDim} & \thead{\#Blocks} & \thead{\#Heads/HeadDim} & \thead{\#Params                     \\ M} & \thead{Peak LR \\ (15B)} & \thead{Peak LR \\ (300B)}  \\
    \midrule
    \multirow{5}{*}{{\rotatebox[origin=c]{90}{125M}}} 
    & RWKV-5 & 768                  & 12               & -                       & 176.5           & 3e-3    & -       \\
    & RWKV-6 & 768                  & 12               & -                       & 173.6           & 3e-3    & -       \\
    & HGRN2  & 768                  & 12               & -                       & 162.2           & 3e-3    & -       \\
    & RWKV-4 & 768                  & 12               & -                       & 169.4           & 3e-3    & 6e-4    \\
    & Llama  & 768                  & 12               & 12 / 64                 & 162.2           & 3e-3    & 3e-3    \\
    & Mamba  & 768                  & 24               & -                       & 167.8           & 3e-3    & 3e-3    \\
    & xLSTM  & 768                  & 24               & 4 / 384                 & 163.8           & 3e-3    & 1.5e-3  \\
    \midrule
    \multirow{5}{*}{{\rotatebox[origin=c]{90}{350M}}} 
    & RKWV-5 & 1024                 & 24               & -                       & 455.7           & 1e-3    & -       \\
    & RWKV-6 & 1024                 & 24               & -                       & 441.6           & 1e-3    & -       \\
    & HGRN2  & 1024                 & 24               & -                       & 411.4           & 1e-3    & -       \\
    & RWKV-4 & 1024                 & 24               & -                       & 430.5           & 1e-3    & 4e-4    \\
    & Llama  & 1024                 & 24               & 16 / 64                 & 406.6           & 1.5e-3  & 1.5e-3  \\
    & Mamba  & 1024                 & 48               & -                       & 423.1           & 1.5e-3  & 1.5e-3  \\
    & xLSTM  & 1024                 & 48               & 4 / 512                 & 409.3           & 1e-3    & 7.5e-4  \\
    \midrule
    \multirow{5}{*}{{\rotatebox[origin=c]{90}{760M}}} 
    & RWKV-5 & 1536                 & 24               & -                       & 947.8           & 9e-4    & -       \\
    & RWKV-6 & 1536                 & 24               & -                       & 907.7           & 9e-4    & -       \\
    & HGRN2  & 1536                 & 24               & -                       & 834.2           & 9e-4    & -       \\
    & RWKV-4 & 1536                 & 24               & -                       & 891.0           & 2e-3    & 2.5e-4  \\
    & Llama  & 1536                 & 24               & 16 / 96                 & 834.1           & 1.25e-3 & 1.25e-3 \\
    & Mamba  & 1536                 & 48               & -                       & 870.5           & 1.25e-3 & 1.25e-3 \\
    & xLSTM  & 1536                 & 48               & 4 / 768                 & 840.4           & 9e-4    & 6.25e-4 \\
    \midrule
    \multirow{5}{*}{{\rotatebox[origin=c]{90}{1.3B}}} 
    & RWKV-5 & 2048                 & 24               & -                       & 1616.0          & 9e-4    & -       \\
    & RWKV-6 & 2048                 & 24               & -                       & 1537.5          & 9e-4    & -       \\
    & HGRN2  & 2048                 & 24               & -                       & 1439.4          & 9e-4    & -       \\
    & RWKV-4 & 2048                 & 24               & -                       & 1515.2          & 1e-3    & 2e-4    \\
    & Llama  & 2048                 & 24               & 16 / 128                 & 1420.4          & 1e-3    & 1e-3    \\
    & Mamba  & 2048                 & 48               & -                       & 1475.3          & 1e-3    & 1e-3    \\
    & xLSTM  & 2048                 & 48               & 4 / 1024                & 1422.6          & 9e-4    & 5e-4    \\
    \midrule
    \multirow{5}{*}{{\rotatebox[origin=c]{90}{2.7B}}}
    & RWKV-5 & 2048                 & 24               & -                       & 3194.7          & 8e-4    & -       \\                  
    & RWKV-6 & 2048                 & 24               & -                       & 3021.9          & 8e-4    & -       \\
    & HGRN2  & 2048                 & 24               & -                       & 2795.4          & 8e-4    & -       \\                      
    & RWKV-4 & 2560                 & 32               & -                       & 2984.8          & 8e-4    & -       \\
    & Llama  & 2560                 & 32               & 32 / 80                 & 2779.5          & 8e-4    & -       \\
    & Mamba  & 2560                 & 64               & -                       & 2897.2          & 8e-4    & -       \\
    & xLSTM  & 2560                 & 64               & 4 / 1280                & 2788.3          & 8e-4    & -       \\
    \bottomrule
\end{tabular}
    \end{adjustbox}
    \vspace{0.1cm}
    \caption{Peak learning rates and model dimensions for scaling law plots.}
    \label{tab:spaj_peaklr}
\end{table}

\paragraph{Details on Comparison to Other Methods.}
\label{sec:appExpComparison}
For the model comparison on 15B training tokens of SlimPajama we train all models with context length 2048 and batch size 256.
We use a peak learning rate of 1e-3 for all models for comparability. 
The learning rate decays over 30k training steps. 
The models are compared after one epoch at training step 28170. 
As model implementations we use the original repositories' code for Mamba~\citep{Gu:24arxiv} \footnote{\url{https://github.com/state-spaces/mamba}}, RWKV-5, RWKV-6~\citep{Peng:24arxivshort} \footnote{\url{https://github.com/BlinkDL/RWKV-LM/}}. 
For RWKV-4 we use a cleaned and validated re-implementation based on the original repo and kernels~\citep{Peng:23arxivshort}. 
In our RWKV-4 implementation we enable weight decay on all parameters except biases, the token embedding weight and all LayerNorm weights.
For HGRN~\citep{Qin:23}, GLA~\citep{Yang:23arxiv}, HGRN2~\citep{Qin:24arxiv} we use the a re-implementation \texttt{flash-linear-attention}~\citep{Yang:24} by the authors of GLA~\citep{Yang:23arxiv,Yang:24} \footnote{\url{https://github.com/sustcsonglin/flash-linear-attention}}. 
For GPT-3 and Llama-like Transformers, we use our own implementations based on PyTorch. Note that for all xLSTMs, Transformers, Mamba and RWKV-4, we use Mixed Precision training with \texttt{bfloat16} and weights in \texttt{float32} precision.
Following the general training procedure we use \texttt{torch.compile} for all models, except for Transformers and models using the \texttt{flash-linear-attention} library because of compilation problems.

As RWKV-6 performs worse than RWKV-5, we also train a model with peak learning rate 4e-4, as reported in the original repository for 350M parameter models~\footnote{\url{https://github.com/BlinkDL/RWKV-LM/blob/64b7fe4c66fcf7da37019630268075b0558f6dc5/RWKV-v5/train.py\#L44}}. This model reaches a perplexity of 16.38, worse than the 15.03 for the standard peak learning rate 1e-3 as reported in Table~\ref{tab:spaj15b_model_results}. Similarly, we tested the repository learning rates for other model sizes and all performed worse than the ones we also use for xLSTM (see Table~\ref{tab:spaj_peaklr}).

\paragraph{Details on Training Precision for Baselines.}
For models from \texttt{flash-linear-attention} and
RWKV-5/6 models we found that PyTorch automatic mixed precision training did not work, but
casting the model weights to float32 initially with FSDP parameter precision bfloat16 led to a
working configuration. In this setting, models perform better than in full bfloat16 training, where the weights are casted to bfloat16 initially as well. Full float32 training did not work because of the custom kernels that require bfloat16.

\paragraph{General Details on Ablation Studies.} 
\label{sec:appAblDetails}
We follow our general training procedure and train all models with context length 2048, batch size 256 and peak learning rate 1e-3. 
We report perplexity values on the validation set.

\pagebreak
 \paragraph{Additional Ablation Study on Matrix Memory.} As default block configuration we use the mLSTM in the pre up-projection block (see Figure~\ref{fig:appmLSTM_detailed}) and the sLSTM in the post up-projection block (see Figure~\ref{fig:appsLSTM_detailed}). 
In this experiment we study combination of mLSTM with different block variants using the xLSTM[1:0] architecture. 
We compare the mLSTM in a post up-projection block (see Figure~\ref{fig:Backbones}~and~\ref{fig:appsLSTM_detailed}) with $\text{ReLU}^2$ activation function and non-gated feed-forward network to mLSTM in a pre up-projection block with and without a dimension-wise causal convolution.
Table~\ref{tab:abl_matrixmemory} shows that the matrix memory benefits from the pre up-projection block structure, and that the convolution within this block is important.

\begin{table}[htbp]
    \centering
    \begin{adjustbox}{width=\textwidth}    
    \begin{tabular}{llcccr}
    \toprule
    Model                       & Details                                         & \#Blocks & \thead{Embedding                 \\ Dim}& \thead{\#Params         \\ M} & \thead{SlimPajama \\ (15B) ppl $\downarrow$}  \\
    \midrule
    \multirow{3}{*}{xLSTM[1:0]} & Post Up-Projection Block (ReLU2) & 24       & 1024             & 430.4 & 13.90 \\ % (1)
                                & Pre Up-Projection Block, No Convolution         & 48       & 1024             & 408.8 & 15.41 \\ % (2)
                                & Pre Up-Projection Block, With Convolution       & 48       & 1024             & 409.3 & \first{13.43} \\ % (3) % this is the reevaluation result
    \bottomrule
\end{tabular}

% % mLSTM
% (2)
% spaj15b-xlstmblock2-350M_vlstmv2_abl_noconvnocact_0-B48E1024H4gbs256--sn-350M-ns-30000-l-0.001-wd-0.1-gas-1-rd-float32-pd-bfloat16-bd-bfloat16-utc-1-nb-48-ed-1024-nh-4-nn-8-s-NO_SHARD-bs-4-vbsf-6-seed-0--240415_075401 350M_vlstmv2_conv val/._Perplexity	15.407269477844238
% (1)
% spaj15b-vlstm-350M_vlstmv1_fprodv0_0-B24E1024H4gbs256--sn-350M-ns-30000-l-0.001-wd-0.1-gas-1-rd-float32-pd-bfloat16-bd-bfloat16-utc-1-nb-24-ed-1024-nh-4-nn-4-s-NO_SHARD-bs-8-vbsf-3-seed-0--240315_164124 350M_vlstmv1 val/._Perplexity	13.900581359863281
% (3)
% spaj15b-xlstmblock2-350M_vlstmv2_fprodv0_0-B48E1024H4gbs256--sn-350M-ns-30000-l-0.001-wd-0.1-gas-1-rd-float32-pd-bfloat16-bd-bfloat16-utc-1-nb-48-ed-1024-nh-4-nn-8-s-NO_SHARD-bs-4-vbsf-6-seed-0--240315_163952 350M_vlstmv2 val/._Perplexity	13.446783065795898
    \end{adjustbox}
    \vspace{0.1cm}
    \caption{Matrix Memory variants. We study different configurations for the matrix memory. Matrix memory in the pre up-projection block performs best and gives xLSTM[1:0]. Notably, it seems that the dimension-wise causal convolution within the pre up-projection block is important.}
    \label{tab:abl_matrixmemory}
\end{table}

\paragraph{Details on new xLSTM Components Ablation Study.} In Table~\ref{tab:ablstudies} (top), we show our modifications to the vanilla LSTM that transform the vanilla LSTM into the xLSTM. 
We start with a large default PyTorch LSTM with 24 layers and 1536 hidden size. Due to a lack of skip-connections and LayerNorms, vanilla LSTMs of this size are not trainable. 
We then add skip-connections and pre-LayerNorms before each LSTM layer corresponding to a residual architecture. 
This enables training for LSTMs at this scale. 
Replacing every second LSTM layer by a non-gated feed-forward network with GeLU activation function (similar to \citeauthor{Vaswani:17}), which corresponds to the post up-projection backbone (see Figure~\ref{fig:Backbones}) further boosts performance. 
Adding Exponential Gating to this architecture yields the sLSTM as depicted in Figure~\ref{fig:appsLSTM_detailed}, with another large performance improvement.
Finally, adding the best Matrix Memory variant found in Table~\ref{tab:abl_matrixmemory} by replacing some sLSTM blocks with the mLSTM (see Figure~\ref{fig:appmLSTM_detailed}) gives xLSTM[7:1] with the best performance.

\paragraph{Details on Gating Technique Ablation Study.} In Table~\ref{tab:ablstudies} (bottom), we investigate the effect of trainable and input-dependent gates for mLSTM. The results show that, in contrast to other methods~\citep{Katharopoulos:20, Sun:23arxiv,Qin:23, Katsch:23, Yang:23arxiv, Qin:24arxiv, Peng:24arxivshort}, having the gates both learnable and input dependent gives the best results.
% KP: Go over that again

%Architecture Modifications from LSTM to xLSTM. We show the design decisions extending the LSTM to become an xLSTM. 
%We start with a large default PyTorch LSTM. Due to a lack of skip-connections and layernorms an LSTM of this size is not trainable.
%Adding skip-connections and pre-layernorms fixes this. Additionally placing MLPs in between the LSTM layers further improves performance.
%By adding exponential gating we outperform the sigmoid gating LSTM by a large margin. 
%Finally, adding the best matrix memory from Table~\ref{tab:spaj15b_abl_matrixmemory} yields the xLSTM[7:1].
%Note that for xLSTM[0:1] the Pre Up-Projection Block performs better, while for xLSTM[7:1] the Post Up-Projection block yields better performance. 
%We hypothesize that this is due to the fact that the Pre Up-Projection Block yields a larger memory state vector and consequently a higher memory capacity. 
%For xLSTM[7:1] the memory is dominated by the mLSTMs matrix memory, hence a smaller memory state vector in the sLSTM is sufficient. \MB{Use reevaluation ppl numbers (will not change the results, just numerical).} 

\paragraph{Details on Scaling Experiments.}
We follow our general training procedure (see paragraph above) 
and train all models, including the 1.3B and 2.7B model sizes,
with context length 2048 and batch size 256. 
We use the peak learning rates from Table~\ref{tab:spaj_peaklr}.
For Llama and Mamba we use the learning rates reported by \citet{Gu:24arxiv}.

\subsection{xLSTM Large Language Models -- SlimPajama300B}
\label{sec:appExpLanguage}

\paragraph{General Training Procedure.}
We use the same general training procedure as in Section~\ref{sec:appGenTrainProc} with peak learning rates from Table~\ref{tab:spaj_peaklr}. 
For Llama and Mamba we use the learning rates reported by \citet{Gu:24arxiv}.
All models are trained with context length 2048. 
The 125M, 350M and 760M models are trained with batch size 256 for 600k training steps, whereas the 1.3B models are trained with batch size 512 for 300k training steps. 
We keep the same learning rate scheduler across all models.
% GUENTER: i think it's too much to introduce
% "General Training Procedure"
% Max: okay, makes sense

\paragraph{Details on Downstream Evaluation.} 
We use the LM Evaluation Harness from EleutherAI~\citep{Sutawika:23short} for evaluating the following tasks that measure common sense reasoning: 
LAMBADA (OpenAI version in LM Evaluation Harness)~\citep{Paperno:16}, HellaSwag~\citep{Zellers:19}, PIQA~\citep{Bisk:20}, ARC-challenge, ARC-easy~\citep{Clark:18arxiv}, WinoGrande~\citep{Sakaguchi:21}.
This selection of downstream tasks is also used in
previous work by~\citet{Gu:24arxiv}.

Following \citet{Gu:24arxiv}, we report accuracy for LAMADA, WinoGrande, PIQA, and ARC-easy, and accuracy normalized by sequence length for HellaSwag and ARC-challenge.

We evaluate all models in full \texttt{float32}, full \texttt{bfloat16} and \texttt{bfloat16} Mixed Precision with weights in \texttt{float32}. 
For each model we select the best value respectively.

\paragraph{Details on PALOMA.}
We use 16 out of the 18 data sources of the PALOMA dataset~\citep{Magnusson:23arxivshort}. 
We use C4~\citep{Raffel:19arxiv}, 
MC4-EN~\citep{Xue:21}, 
Wikitext-103~\citep{Merity:17}, 
PennTreebank~\citep{Vadas:11}, 
RedPajama~\citep{Together:23}, 
Falcon Refinedweb (Refined Web)~\citep{Penedo:23arxiv}, 
Dolma v1.5~\citep{Soldaini:24arxivshort}, 
M2D2 S2ORC, 
M2D2 Wikipedia~\citep{Reid:22}, 
C4-100-Domains (C4 Domains)~\citep{Chronopoulou:22}, 
Dolma-100-Subreddits (Dolma Subreddits)~\citep{Soldaini:24arxivshort}, 
Dolma-100-Programming Languages (Dolma Coding)~\citep{Soldaini:24arxivshort, Kocetkov:22arxiv}, 
TwitterAAE~\citep{Blodgett:16, Liang:23short}, 
Manosphere Corpus~\citep{Ribeiro:21}, 
GAB Corpus~\citep{Zannettou:18}, 
4CHAN Corpus~\citep{Papasavva:20}. 
We leave out ThePile~\citep{Gao:21arxiv} and ICE~\citep{Greenbaum:96} as they are not part of Paloma's Huggingface dataset repository\footnote{\url{https://huggingface.co/datasets/allenai/paloma}}.
A detailed description of these datasets can be found in \citet[Table~2]{Magnusson:23arxivshort}. 
All models are evaluated in \texttt{bfloat16} Mixed Precision. 

Results on the data sources % Max: the term "data source" is used in PALOMA paper. 
TwitterAAE, Manosphere, GAB and 4CHAN are reported in 
Table~\ref{tab:ppleval_app} and for each individual dataset
the results are given in Section~\ref{sec:paloma}.

\begin{table}[htbp]
    \centering
    \begin{adjustbox}{width=0.6\textwidth}
        \begin{tabular}{llcrrrr}
    \toprule
                                                      & Model      & \thead{\#Params                                  \\M} & \thead{Twitter\\AAE} & Manosphere & 4CHAN & GAB   \\
    \midrule
    \multirow{5}{*}{{\rotatebox[origin=c]{90}{125M}}} & RWKV-4     & 169.4           & 265.80 & 39.31 & 18.48 & 53.89 \\
                                                      & Llama      & 162.2           & 277.93 & 32.98 & 14.03 & 56.45 \\
                                                      & Mamba      & 167.8           & 258.17 & 32.14 & 14.01 & 51.58 \\
                                                      & xLSTM[1:0] & 163.8           & 244.53 & 31.45 & 13.27 & 51.00 \\
                                                      & xLSTM[7:1] & 163.7           & 248.51 & 30.90 & 13.45 & 50.25 \\
    \midrule
    \multirow{5}{*}{{\rotatebox[origin=c]{90}{350M}}} & RWKV-4     & 430.5           & 216.17 & 30.25 & 13.82 & 42.25 \\
                                                      & Llama      & 406.6           & 231.09 & 25.90 & 11.49 & 43.04 \\
                                                      & Mamba      & 423.1           & 202.88 & 25.24 & 11.60 & 40.78 \\
    %   & Mamba      & 371.5           & 201.06 & 25.31 & 11.5  & 40.28 \\
                                                      & xLSTM[1:0] & 409.3           & 200.61 & 24.58 & 11.20 & 39.83 \\
                                                      & xLSTM[7:1] & 408.4           & 206.25 & 24.73 & 11.31 & 39.86 \\
    \midrule
    \multirow{5}{*}{{\rotatebox[origin=c]{90}{760M}}} & RWKV-4     & 891.0           & 195.27 & 24.66 & 12.00 & 35.73 \\
                                                      & Llama      & 834.1           & 205.50 & 22.69 & 10.40 & 37.68 \\
                                                      & Mamba      & 793.2           & 182.74 & 22.58 & 10.47 & 36.25 \\
                                                      & xLSTM[1:0] & 840.4           & 179.74 & 21.66 & 10.11 & 35.33 \\
                                                      & xLSTM[7:1] & 839.7           & 180.19 & 21.78 & 10.22 & 34.89 \\
    \midrule
    \multirow{5}{*}{{\rotatebox[origin=c]{90}{1.3B}}} & RWKV-4     & 1515.2          & 174.87 & 23.51 & 11.34 & 33.18 \\
                                                      & Llama      & 1420.4          & 192.52 & 20.67 & 9.67  & 34.84 \\
                                                      & Mamba      & 1475.3          & 171.38 & 20.37 & 9.80  & 32.01 \\
                                                      & xLSTM[1:0] & 1422.6          & 166.16 & 19.94 & 9.64  & 31.90 \\
                                                      & xLSTM[7:1] & 1420.1          & 171.36 & 20.28 & 9.64  & 32.17 \\
    \bottomrule
\end{tabular}

    \end{adjustbox}
    \vspace{0.1cm}
    \caption{Perplexity values per domain.
    \label{tab:ppleval_app}
}  
\end{table}

In order to evaluate the perplexity values on each data source,
we split the text documents into sequences of length 2048, which corresponds to the pre-training context length of all models.
For documents longer than 2048 tokens we split each document into non-overlapping input sequences. 
In this case for the last input sequence, we follow the LM Evaluation Harness and fill up the full 2048 token context window with previous tokens, but compute the perplexity only on the remaining tokens.

We compute the token perplexities per data source in Table~\ref{tab:ppleval} as the exponential of the negative loglikelihoods per domain weighted by the number of tokens per domain in that data source as it is defined in \citet[Equation~1]{Magnusson:23arxivshort}

\clearpage

\section{Detailed Results on PALOMA Language Model Evaluation}
\label{sec:paloma}

We report the perplexity values on each of the 571 subdomains of PALOMA in Table~\ref{tab:spaj300B_ppl_evaluation_detail}.
Note that the aggregated perplexity values in Table~\ref{tab:ppleval} are not macro averages of the values shown in Table~\ref{tab:spaj300B_ppl_evaluation_detail}.

{\footnotesize
	\newlength{\mywidth}
	\setlength{\mywidth}{1.7cm}
	\addtolength{\mywidth}{-\tabcolsep}
	\setlength{\tabcolsep}{0pt}
\begin{longtable}{m{6cm}m{1.7cm}m{1.7cm}m{1.7cm}m{1.7cm}m{1.7cm}}
 \caption{PPL Evaluations: For the 1.3B sized models trained on 300B SlimPajama tokens, these are the detailed evaluation results on the respective validation datasets.
%\vspace{0.1cm}
\label{tab:spaj300B_ppl_evaluation_detail}}\\

	Dataset & \makebox[\mywidth][c]{Llama} & \makebox[\mywidth][c]{Mamba} & \makebox[\mywidth][c]{RWKV-4} & \makebox[\mywidth][c]{xLSTM[7:1]} & \makebox[\mywidth][c]{xLSTM[1:0]} \\ \endfirsthead
	Dataset & \makebox[\mywidth][c]{Llama} & \makebox[\mywidth][c]{Mamba} & \makebox[\mywidth][c]{RWKV-4} & \makebox[\mywidth][c]{xLSTM[7:1]} & \makebox[\mywidth][c]{xLSTM[1:0]} \\ \endhead
	\hline
	\#Params (M) & \makebox[\mywidth][c]{1420} & \makebox[\mywidth][c]{1475} & \makebox[\mywidth][c]{1515} & \makebox[\mywidth][c]{1420} & \makebox[\mywidth][c]{1423} \\
	4chan\_meta\_sep\_val-00000000  & \colorbox[HTML]{91d183}{\makebox[\mywidth][c]{9.58}} & \colorbox[HTML]{c9e99b}{\makebox[\mywidth][c]{9.72}} & \colorbox[HTML]{ffffe5}{\makebox[\mywidth][c]{11.37}} & \colorbox[HTML]{77c578}{\makebox[\mywidth][c]{9.53}} & \colorbox[HTML]{84cb7d}{\makebox[\mywidth][c]{9.55}}\\
	4chan\_meta\_sep\_val-00000001  & \colorbox[HTML]{a0d789}{\makebox[\mywidth][c]{9.95}} & \colorbox[HTML]{ceeb9e}{\makebox[\mywidth][c]{10.06}} & \colorbox[HTML]{ffffe5}{\makebox[\mywidth][c]{11.57}} & \colorbox[HTML]{8bce80}{\makebox[\mywidth][c]{9.91}} & \colorbox[HTML]{77c578}{\makebox[\mywidth][c]{9.88}}\\
	4chan\_meta\_sep\_val-00000002  & \colorbox[HTML]{8ccf81}{\makebox[\mywidth][c]{9.42}} & \colorbox[HTML]{c3e698}{\makebox[\mywidth][c]{9.53}} & \colorbox[HTML]{ffffe5}{\makebox[\mywidth][c]{11.00}} & \colorbox[HTML]{7fc97b}{\makebox[\mywidth][c]{9.40}} & \colorbox[HTML]{77c578}{\makebox[\mywidth][c]{9.38}}\\
	4chan\_meta\_sep\_val-00000003  & \colorbox[HTML]{81c97c}{\makebox[\mywidth][c]{9.78}} & \colorbox[HTML]{c6e89a}{\makebox[\mywidth][c]{9.93}} & \colorbox[HTML]{ffffe5}{\makebox[\mywidth][c]{11.48}} & \colorbox[HTML]{7cc77a}{\makebox[\mywidth][c]{9.77}} & \colorbox[HTML]{77c578}{\makebox[\mywidth][c]{9.77}}\\
	c4\_100dom\_val\_100\_www.ign.com  & \colorbox[HTML]{fdfede}{\makebox[\mywidth][c]{16.22}} & \colorbox[HTML]{eaf7af}{\makebox[\mywidth][c]{15.75}} & \colorbox[HTML]{ffffe5}{\makebox[\mywidth][c]{17.10}} & \colorbox[HTML]{ddf1a6}{\makebox[\mywidth][c]{15.67}} & \colorbox[HTML]{77c578}{\makebox[\mywidth][c]{15.43}}\\
\end{longtable}
}

%##########################################################
%##########################################################
%##########################################################

% \clearpage

% \section*{Acknowledgements}
% We thank NXAI for generously providing the computational resources for this project.
% \\
% We acknowledge EuroHPC Joint Undertaking for awarding us access to Karolina at IT4Innovations, Czech Republic, MeluXina at LuxProvide, Luxembourg, LUMI at CSC, Finland and Leonardo at CINECA, Italy (proposals EHPC-DEV-2023D09-028, EHPC-DEV-2023D10-005, EHPC-DEV-2023D10-022, EHPC-DEV-2023D10-024, EHPC-DEV-2023D10-004, EHPC-DEV-2023D09-027).

% We also thank Sebastian Lehner, Daniel Klotz, Thomas Adler, Niklas Schmidinger, Lukas Hautzenberger, Fabian Paischer, Kajetan Schweighofer, Anna Zimmel, Lukas Aichberger, Gerald Gutenbrunner and Matthias Dellago for helpful discussions regarding the project. 

\end{appendix}
\end{document}